\documentclass[IJRR]{RLIpaper}

\usepackage{geometric_algebra}
\usepackage{subfiles}
\usepackage{todonotes}

\usetikzlibrary{shapes.geometric,arrows,intersections}
\tikzstyle{arrow} = [thick,->,>=stealth]    

\usepackage{array}
\newcolumntype{b}{>{\bfseries} l}                            

\title{\textcolor{black}{Cooperative Task Spaces for Multi-Arm Manipulation Control based on Similarity Transformations}}
\author{Tobias L\"ow \affilnum{1}\,\affilnum{2},
Cem Bilaloglu \affilnum{1}\,\affilnum{2},
Sylvain Calinon \affilnum{1}\,\affilnum{2}
} 
\runninghead{L\"ow et al.}
    
\affiliation{\affilnum{1}Idiap Research Institute, Martigny, Switzerland\\
  \affilnum{2}EPFL, Lausanne, Switzerland\\
}

\corrauth{Tobias L\"ow,
Idiap Research Institute,
EPFL,
Switzerland.}

\email{tobias.loew@epfl.ch}

\begin{document}

\begin{abstract}
    Many tasks in human environments require collaborative behavior between multiple kinematic chains, either to provide additional support for carrying big and bulky objects or to enable the dexterity that is required for in-hand manipulation. Since these complex systems often have a very high number of degrees of freedom coordinating their movements is notoriously difficult to model. In this article, we present the derivation of the theoretical foundations for cooperative task spaces of multi-arm robotic systems based on geometric primitives defined using conformal geometric algebra. Based on the similarity transformations of these cooperative geometric primitives, we derive an abstraction of complex robotic systems that enables representing these systems in a way that directly corresponds to single-arm systems. By deriving the associated analytic and geometric Jacobian matrices, we then show the straightforward integration of our approach into classical control techniques rooted in operational space control. We demonstrate this using bimanual manipulators, humanoids and multi-fingered hands in optimal control experiments for reaching desired geometric primitives and in teleoperation experiments using differential kinematics control. We then discuss how the geometric primitives naturally embed nullspace structures into the controllers that can be exploited for introducing secondary control objectives. This work, represents the theoretical foundations of this cooperative manipulation control framework, and thus the experiments are presented in an abstract way, while giving pointers towards potential future applications.  
\end{abstract}

\keywords{Conformal Geometric Algebra}

\maketitle


\section{Introduction}
\label{sec:introduction}
    
    Humans excel at dexterous manipulation using the redundancy of their arms and hands to achieve complex tasks where cooperative behaviors between the different kinematic chains is required. Many environments are built around these advanced manipulation capabilities, and robots that are increasingly operating within them. Yet, it still remains challenging for robots to achieve dexterous manipulation that is on par with humans \parencite{kadalageresampathReviewHumanlikeRobot2023}. Nevertheless, robots are now expected to handle various tasks that require them to interact with objects in a way that goes beyond traditional parallel jaw grippers or suction cups that are often featured on single arm manipulators. These types of end-effectors are limiting the robot manipulation capabilities, since many objects are often more complex and require different strategies. For instance, big and bulky objects may call for dual-arm manipulators for picking and placing. This kind of coordination between two arms is vital when grasping, lifting, and transferring objects of varying shape, weight, and fragility. This is similarly true for intricate tasks such as in-hand manipulating and multi-fingered grasping using robotics hands. The common ground here is coordinating the movements of two or more parallel kinematic chains to interact with the environment during manipulation tasks.

    However, many manipulation tasks do not require exploiting the full redundancy of the system. This is because the kinematic chains are not acting independently but instead are coordinated to achieve a shared objective. As a result, the effective behavior of the system often lies on a lower-dimensional manifold embedded within the full configuration space. We refer to these manifolds exhibiting specific geometric structures as cooperative task spaces. Recognizing and leveraging this structure not only reduces the dimensionality of the control or planning problems but also improves interpretability. Moreover, these geometric structures naturally give rise to geometric nullspaces, which can be exploited for secondary objectives, such as regulating contact forces, without interfering with the primary task objective, as we previously demonstrated in~\parencite{bilalogluTactileErgodicCoverage2025}.
    
\begin{figure}[!ht]
    	\centering
    	\includegraphics[width=0.8\linewidth]{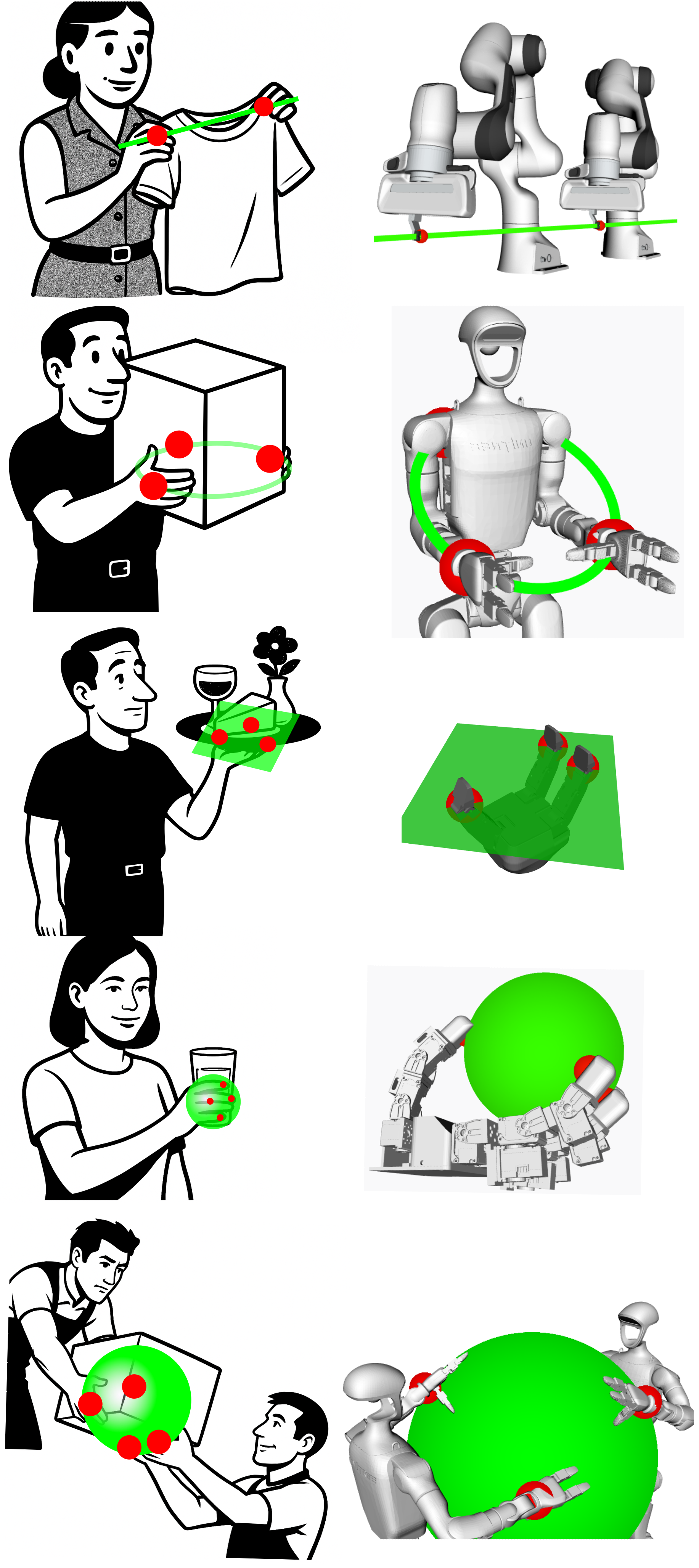}
    	\caption{Overview of various daily-life tasks and how they can be modeled using the cooperative geometric primitives. Note that the geometric primitives do not approximate the shape of the objects, but instead model the cooperative behavior of the kinematic chains. The corresponding reference points are shown in red. \emph{Left:} Human activities with an overlay of the proposed geometric primitive to capture the task constraints. \emph{Right:} Corresponding robotic system with the cooperative geometric primitive achieving the same behavior.}
    	\label{fig:overview}
    \end{figure}

    Existing work using dual quaternions~\parencite{adornoDualPositionControl2010} and geometric algebra~\parencite{lowExtendingCooperativeDualTask2024} has begun to explore cooperative task spaces, particularly in the context of dual-arm manipulation. However, the notion of cooperation extends well beyond dual-arm settings. Many real-world scenarios involving multiple kinematic chains, such as torso-arm-hand coordination or multi-fingered in-hand manipulation, can benefit from a unified geometric framework that generalizes cooperative control. To illustrate this, we provide a non-exhaustive collection of robotic systems and manipulation challenges that can be abstracted through cooperative subspaces corresponding to geometric primitives, as shown in Figure~\ref{fig:overview}. The geometric primitives only depend on the number of kinematic chains that are involved in the task. Hence, they model the cooperative behavior, and are not chosen based on an approximation of the object shape. Our overarching goal is to generalize operational space control~\parencite{khatibUnifiedApproachMotion1987}, which was originally formulated for single manipulators, to systems with multiple kinematic chains by representing cooperative task spaces as geometric primitives within the control architecture.

    In this article, we address the challenge of cooperative manipulation control for robotic systems involving multiple parallel kinematic chains. We do so by combining the various ideas around cooperative control, grasping nullspaces and geometric primitives, into a single, mathematically coherent framework by introducing cooperative geometric primitives and integrating them into geometrically consistent control schemes. For this, we leverage conformal geometric algebra (CGA) to find algebraic objects that represent geometric primitives that go beyond keypoints. In prior work, we have shown how these primitives define geometric nullspaces and how they can be used in optimal control \parencite{lowGeometricAlgebraOptimal2023}. In the same optimal control framework, we have then extended the cooperative dual-task space using CGA for bimanual robots \parencite{lowExtendingCooperativeDualTask2024}. Here, we further extend this idea, by exploring how cooperative geometric primitives enable decentralized control strategies that harmonize task execution, disturbance rejection, and inter-agent coordination. We show that by exploiting CGA, different systems can be modelled in a uniform manner that allows informed collaboration strategies based on the geometric primitives and conformal transformations that can be expressed in this algebra. Our contributions therefore are
    \begin{itemize}
        \item A geometric representation for cooperative behaviors of multiple parallel kinematic chains using geometric primitives, enabling intuitive modeling of coordinated manipulation tasks;
        \item Introduction of similarity transformations of geometric primitives, extending classical rigid body transformations from single end-effector control to multi-chain cooperative manipulation;
        \item A unified mathematical framework for cooperative control based on similarity transformations, that can be used with optimal control or differential kinematics formulations;
        \item Exploitation of geometric nullspaces inherent in the cooperative task space for achieving secondary control objectives, such as contact force regulation or redundancy resolution, without affecting primary task execution;
        \item Demonstration of the versatility and scalability of the proposed approach on complex robotic systems, including humanoid platforms and multi-fingered hands, through tasks involving cooperative reaching and teleoperation.
    \end{itemize}

    
    \subsection{Related Work}
    \label{sub:related_work}
        
        A common example for highlighting the challenges of cooperative manipulation is the collective transport of big and bulky objects that cannot be handled by a single robot. This task is very challenging, since multiple robots interact with the same object via local control actions \parencite{erhartImpedancebasedControlArchitecture2013}, which requires them to cooperate in order to achieve a common, global goal. Traditional approaches often use a leader-follower strategy, where the different roles are predefined, and the agents communicate explicitly. However, these strategies often fail due to uncertainties stemming from the task parameters, environmental dynamics, or the team composition \parencite{farivarnejadMultirobotControlStrategies2022}. Solving these issues requires strategies that do not rely on centralized coordination, but instead use the manipulated object as a physical channel for implicit coordination. By sensing the exerted forces on the shared payload, the robots can decompose the motion into an allowed task space motion and a nullspace motion that is used rejecting perturbations \parencite{careyForceMediatedControllerCooperative2024}. This strategy minimizes force conflicts, and enables cooperative manipulation even under strict communication constraints. Recent advances in cooperative control favor frameworks that further emphasize distributed optimization and adaptability. They optimize the individual performance of the robots in the system that suffers reduced operation capabilities due to the physical connection through grasping \parencite{heDistributedOptimalControl2022}. Furthermore, the optimization can include objectives for enhancing the manipulability, avoiding obstacles, and reducing internal forces caused by robot motion errors due to coupling \parencite{heOptimisationBasedDistributedCooperative2020}. Hence, there is an inherent trade-off between physical coupling (e.g. rigid grasps) and operational flexibility, which makes it important to deal with disturbances to avoid unexpected behaviors \parencite{aladeleAdaptiveCooperativeManipulation2022}. Learning-based approaches try to simplify the problem of coordinating multiple agents by introducing actions through shared latent spaces, which reduces the sample complexity of high-dimensional tasks like dual-arm manipulation \parencite{aljalboutCLASCoordinatingMultiRobot2023}. Alternatively, bi-manual end-effector poses are learned from demonstrations using task-parameterized dynamical systems \parencite{silverioLearningBimanualEndeffector2015}.

        {\color{black} 
        Dynamic Movement Primitives (DMPs) have emerged as a powerful way to encode complex trajectories while preserving stability guarantees. A comprehensive tutorial survey frames DMPs as a bridge between model‑based control and data‑driven learning, highlighting their suitability for multi‑robot settings where each agent can generate locally consistent motions that nevertheless respect a globally defined attractor \parencite{saverianoDynamicMovementPrimitives2023}. Building on this foundation, it has been demonstrated that DMPs can be combined with velocity scaling to personalise human‑robot collaborative transport. By modulating the DMP speed according to the measured interaction forces, the system achieves smooth, safe hand‑overs even when the human operator varies the payload weight or trajectory on‑the‑fly \parencite{franceschiHumanrobotCollaborativeTransport2025}. Other work has leveraged human motion prediction based on DMPs to enable seamless collaborative object transfer, showing that anticipating the human partner’s trajectory improves coordination and reduces interaction forces \parencite{sidiropoulosHumanrobotCollaborativeObject2019}. The sensed forces on the shared object become the scaling signal for the DMPs, thereby turning the payload itself into a communication medium that implicitly synchronises the robots’ motion. While many cooperative manipulation studies assume rigid payloads, real‑world logistics increasingly involve deformable or soft objects (e.g., bags of grain, flexible containers). Movement‑Primitive Diffusion introduced a generative framework that learns gentle manipulation policies for deformable items directly from demonstrations \parencite{scheiklMovementPrimitiveDiffusion2024}. The diffusion model produces smooth DMP parameters that respect material compliance, effectively reducing internal stresses that would otherwise arise from hard‑grasping strategies. This approach complements the earlier discussion on internal‑force minimisation \parencite{heOptimisationBasedDistributedCooperative2020} by providing a data‑driven means to shape the primitive itself, rather than relying solely on post‑hoc optimisation. Moreover, the diffusion‑generated primitives can be shared across robots, enabling distributed execution without explicit communication—each robot samples the same stochastic primitive conditioned on its local perception of the object’s deformation.
        }

        Some approaches are inspired by the formation control that is often used in the area of autonomous ground vehicles \parencite{spletzerCooperativeLocalizationControl2001}, and are extended to manipulation, where the agents are treated as part of a cohesive geometric formation \parencite{sieberFormationbasedApproachMultirobot2013}. In this setting, one of the agents can be a human that is guiding the robots, which in turn try to maintain desired geometry w.r.t. each other, while reducing internal stresses \parencite{sieberHumanGuidedMultirobotCooperative2019}. It can then be shown that, the internal interactions forces in cooperative manipulation that arise from a formation of robots, lie in the null space of the grasp matrix, i.e. the dynamics of the object are not affected by them \parencite{depascaliDecouplingSchemeForce2022}. This insight also bridges the connection to literature on grasping and in-hand manipulation, where the problem setting is similar, in the sense that multiple parallel kinematic chains are tasks to cooperatively manipulate a shared object. Both scenarios have a high kinematic redundancy can be exploited \parencite{yaoExploitingKinematicRedundancy2023}. We can therefore also mirror control strategies for robotic grip adjustment, to enable adaptive force modulation for cooperative robotic manipulation in response to environmental feedback. Here, compliant control behaviors are achieved through impedance control schemes on the object level \parencite{pfanneObjectLevelImpedanceControl2020}. Dexterity in contact-rich manipulation is often limited, because in contrast to humans, ty{}pically grasping models for robotic hands only use single contact points at the fingertips, instead of the whole surface~\parencite{bircherComplexManipulationSimple2021}. Smart gripper design can alleviate this issue to a certain extent, where the geometry is optimized to achieve more dexterous manipulation capabilities, such as fixing the movement of the object's pose along sphere \parencite{patelRobotHandBased2021}. This highlights the importance of geometric considerations for encoding spatial relationships and constraints and the role of geometric primitives. 

        Geometric primitives also play an important role in recent advances on general manipulation control, currently mainly in the form of object-centric keypoints. The goal here is to reduce complex tasks across a category of objects, where shape variations are allowed, to geometric constraints on keypoints \parencite{gaoKPAM20Feedback2021}. When used in closed-loop manipulation control for achieving contact-rich tasks, this keypoint-based representation leads to automatic generalization to a category of objects by incorporating an object-centric action representation \parencite{manuelliKPAMKeyPointAffordances2022}. The simplicity of the representation via keypoints has also been proven to have advantages for visual imitation learning. Here, task representations are extracted from demonstrations and decomposed into keypoint-based geometric constraints \parencite{gaoKVILKeypointsBasedVisual2023}. This approach has been extended to bimanual manipulation and was shown to generalize to novel scenes in-category objects, due to the object-centric, embodiment-independent, and viewpoint-invariant representation \parencite{gaoBiKVILKeypointsbasedVisual2024}. Expressing task constraints via geometric constraints on keypoints also facilitates the connection to vision-language models, such that tasks can be defined using language instructions \parencite{huangReKepSpatioTemporalReasoning2025}.







    \section{Mathematical Background} 
    
        
        \subsection{Geometric Algebra}
        \label{sub:geometric_algebra}
            
            In this article, we use the specific variant known as conformal geometric algebra (CGA) $\cga$. CGA extends the Euclidean space $\mathbb{R}^3$, characterized by the three basis vectors $\gae{1}, \gae{2}, \gae{3}$, by two additional basis vectors $\gae{4}$ and $\gae{5}$, where $\gae{4}^2=1$ and $\gae{5}^2=-1$. The conformal model is then found by a change of basis that introduces the basis vectors $\gae{0} = \frac{1}{2}(\gae{5} - \gae{4})$ and $\gae{\infty} = \gae{4} + \gae{5}$, which can be understood as a point at the origin and one at infinity. CGA contains both geometric primitives and conformal transformations in the same algebra. We use these properties in our method to define control schemes based on cooperative geometric primitives and their similarity transformations. In this section, we introduce the necessary mathematical background, and we will use the following notation throughout the paper: $x$ to denote scalars, $\bm{x}$ for vectors, $\bm{X}$ for matrices, $X$ for multivectors and $\bm{\mathcal{X}}$ for matrices of multivectors.

            
            \subsubsection{Geometric Primitives}
            \label{ssub:geometric_primitives}
                Euclidean points $\bm{x}$ are embedded in CGA by using the conformal embedding 
                \begin{equation}\label{eq:conformal_embedding}
                    P(\bm{x}) = \gae{0} + \bm{x} + \frac{1}{2}\bm{x}^2\gae{\infty}.
                \end{equation}
                These conformal points form the basic building blocks for geometric primitives that can be represented in the algebra. Note that, this nonlinear embedding turns flat Euclidean space into a parabolic space. Furthermore, this embedding is similar to how we traditionally embed vectors in $\mathbb{R}^3$ into $\mathbb{R}^4$ when using homogeneous coordinates.

                In general, geometric primitives, such as lines, circles and spheres, can be constructed from conformal points using the outer product, i.e. 
                \begin{equation}\label{eq:outer_product_construction}
                    X = \bigwedge_{i=1}^n P_i.
                \end{equation}
                In the above equation, depending on the number of points $n$ and the presence of the point at infinity $\gae{\infty}$, different geometric primitives can be constructed. For example, one can construct a line from two points passing through it and the point at infinity. A circle can be constructed from any three points lying on its orbit. Similar constructions can be used for other geometric primitives, such as planes and spheres. The geometric primitives are subspaces of the algebra, defining a nullspace under the outer product
                \begin{equation}\label{eq:outer_product_nullspace}
                    \outerproductnullspace.
                \end{equation}
                This means that the set of conformalized Euclidean points that result in zero using the outer product forms the geometric primitives \parencite{perwassGeometricAlgebraApplications2009}. We have shown previously how this outer product formulation can be formulated within an optimal control framework to define reaching tasks involving the geometric primitives \parencite{lowGeometricAlgebraOptimal2023}.

                As part of the geometric algebra, the geometric primitives can be used in algebraic expressions that have geometrically meaningful interpretations. For example, the projection of a point $P$ to another geometric primitive $X$ is achieved by the general formula
                \begin{equation}\label{eq:projection}
                    P' = (P\inner X)X^{-1}.
                \end{equation}
                Using what is known as the \emph{meet} operator, it is also possible to calculate intersections between any two geometric primitives
                \begin{equation}\label{eq:meet_operator}
                    Y = (X_1^* \outer X_2^*)^*,
                \end{equation}
                where the $^*$ operator denotes the dual of a multivector, which in CGA amounts to multiplication by the pseudoscalar $I~=~\gae{0123\infty}$. Here, it is not required to additionally consider edge cases. For example, in the case of a line and a circle, it is not necessary to check whether the line is tangential, intersecting the circle twice or not all. Equation \eqref{eq:meet_operator} will always return a meaningful geometric primitive that conveys the information of these different cases. 

            
            \subsubsection{Transformation Groups in CGA}
            \label{ssub:transformation_groups_in_cga}
                For the purpose of robot kinematics and dynamics, we are interested in a representation of the special Euclidean group $\bm{SE}(3)$, i.e. the group of rotations and translations. In CGA it can be identified as the \emph{motor group} $\group{\rigid}$, which is a representation of $Spin(3) \ltimes \mathbb{R}^3$, a double-cover of $\bm{SE}(3)$, meaning that $\versor_{\rigid},-\versor_{\rigid} \in \group{\rigid}$ represent the same transformation. The motor group is a six-dimensional manifold that is also a Lie group with the group constraint $\versor_{\rigid}\inverse{\versor_{\rigid}} = 1$. The corresponding Lie algebra is the bivector algebra $\algebra{\bivector}_\group{\rigid} = \left\{ \gae{23},\gae{13},\gae{12},\gae{1\infty},\gae{2\infty},\gae{3\infty} \right\} $. Elements of the Lie algebra $\bivector\in \algebra{\bivector}_\group{\rigid}$ can be mapped to group elements $\versor_{\rigid}\in\group{\rigid}$ via a surjective map called the exponential map $\exp: \algebra{\bivector} \to \group{\rigid}$. Accordingly, its inverse operation for projecting group elements to the Lie algebra is named the logarithmic map $\log: \group{\rigid} \to \algebra{\bivector}$. Motors can be used to transform any multivector within the algebra, i.e. they can be used to transform geometric primitives. This adjoint operation is formulated as the product 
                \begin{equation}\label{eq:versor_adjoint}
                    X' = \versor_{\rigid}X\reverse{\versor}_{\rigid},
                \end{equation}
                where $\reverse{\versor}_{\rigid}$ denotes the reverse of a multivector.

                In addition to rigid body transformations, CGA also contains elements, called dilators, that cause uniform scaling. These dilators form a group under the geometric product that we denote as $\group{\dilator}$. If we combine rigid body transformations with uniform scaling using the semi-direct product, we obtain the similarity transformation group $\group{\similarity}$, i.e. $\group{\similarity}~=~\group{\rigid}~\rtimes~\group{\dilator}$. Note that, we omit reflections from this group definition, since we want to preserve the handedness. \emph{Similarity transformations} $\versor_\similarity \in \group{\similarity}$ transform geometric primitives using the same Equation~\eqref{eq:versor_adjoint} as motors. In this article, we use similarity transformations to find control laws for systems of multiple parallel kinematic chains. 
                
                For conciseness, we omit here the full definition of the groups. The equations for exponential and logarithmic maps of the different transformation groups in CGA can be found in Appendix \ref{sec:transformations_groups_in_cga}, along with further explanations of these transformations and their relationship to the matrix Lie algebras. 

        
        \subsection{Robot Modeling using CGA}
        \label{sub:robot_modeling_using_cga}
            Using motors, the forward kinematics given the current joint configuration $\posjoint$ of serial kinematic chains can be found via the product of joint motors, i.e. 
            \begin{equation}\label{eq:forward_kinematics}
                \versor_{\rigid}(\posjoint) = \prod_{j=1}^n M_j(q_j) =  \prod_{j=1}^n \exp(q_jB_j),
            \end{equation}
            where $B_j$ is the bivector describing essentially the screw axis of the $j$-th joint. We described the derivation of the analytic $\jacobian^A_{\posjoint}$ and geometric $\jacobian^G_{\posjoint}$ Jacobians in \parencite{lowGeometricAlgebraOptimal2023}. Expressed w.r.t. the end-effector motor, their relationship can be found as 
            \begin{equation}\label{eq:jacobian_relationship}
                \jacobian^{G}_{\posjoint} = -2 \reverse{\versor_{\rigid}}_{\posjoint} \jacobian^A_{\posjoint}. 
            \end{equation}

        \subsection{Geometric Algebra for Optimal Control}
        \label{sub:geometric_algebra_for_optimal_control}

            In \parencite{lowGeometricAlgebraOptimal2023}, we presented an approach for optimal control using CGA that used the outer product for distance computations between geometric primitives. For a conformal point $P(\posjoint)$, this distance can be formulated as the outer product of the target geometric primitive and the point
            \begin{equation}\label{eq:distance_function_vector}                 
                d(\posjoint) = \big\| X \outer P(\posjoint) \big\|_2^2,
            \end{equation}
            where $P(\posjoint) = \versor_{\rigid}(\posjoint) \gae{0} \reverse{\versor}_{\rigid}(\posjoint)$ is the current end-effector point. By definition, the outer product results in zero if the point is contained within the geometric primitive. Hence, the geometric primitives $X$ determine a geometric nullspace for the controller. We showed that, this nullspace automatically leads to compliant behavior when moving tangentially to the geometric primitive, and stiff behavior when moving orthogonally. In contrast, our control formulation in this article is not based on the outer of geometric primitives. Instead, we formulate the multi-arm manipulation control based on similarity transformations of cooperative geometric primitives. While this formulation differs, the insights about the geometric nullspaces remain valid, as we will explain in more detail in this article. 


        
        \subsection{Geometric Nullspace for Impedance Control}
        \label{sub:impedance_control}

            Nullspaces offer an intuitive way to formulate hierarchical control objectives, i.e. to introduce secondary objectives that do not disturb the tracking of the primary one. We have previously exploited the geometric nullspace formed by a line primitive to separate the objectives of tracking a line, while exerting a force along that line in a tactile control application \parencite{bilalogluTactileErgodicCoverage2025}. This separation was achieved purely geometrically without the need to derive an appropriate precision matrix and transforming it to the correct reference frame. Hence, the geometric nullspaces are entirely coordinate-free and thus greatly simplify the definition of hierarchical control objectives. 


        
        \subsection{Cooperative Dual Task Space}
        \label{sub:cooperative_dual_task_space}
            In this section, we briefly introduce the cooperative dual-task space using CGA that we presented in \parencite{lowExtendingCooperativeDualTask2024} as an extension of \parencite{adornoDualPositionControl2010}. The formulation of the CDTS uses two motors to express the collaborative behaviour of the two manipulators. The motors represent a relative and an absolute pose. We first find the relative motor as
            \begin{equation}\label{eq:cdts_relative_motor}
                M_r(\posjoint_1,\posjoint_2) = \reverse{\versor}_{\rigid_2}(\posjoint_2) M_1(\posjoint_1), 
            \end{equation}
            and similarly, the absolute motor can be found as
            \begin{equation}\label{eq:cdts_absolute_motor}
                    M_a(\posjoint_1,\posjoint_2) 
                    = M_2(\posjoint_2)\exp\left(\frac{1}{2}\log\Big(M_r(\posjoint_1,\posjoint_2)\Big)\right),
            \end{equation}
            where $\posjoint_1$ and $\posjoint_2$ denote the joint configurations of the first and second manipulator, respectively. $M_1$ and $M_2$ are the corresponding end-effector motors found using the forward kinematics formula from Equation \eqref{eq:forward_kinematics}.
            For brevity, we omit the derivation of the corresponding analytic Jacobians $\gamatrix{J}^A_r(\posjoint_1,\posjoint_2)$ and $\gamatrix{J}^A_a(\posjoint_1,\posjoint_2)$ and instead refer readers to \parencite{lowExtendingCooperativeDualTask2024} for details.

            Our extension of the CDTS presented a geometric primitive, called the cooperative pointpair, that corresponds to both end-effector positions simultaneously. This cooperative pointpair is defined as the outer product of the two end-effector points, i.e.
            \begin{equation}\label{eq:cooperative pointpair}
                P_{cdts} = M_1(\posjoint_1)\gae{0}\reverse{\versor}_{\rigid_1}(\posjoint_1) \outer M_2(\posjoint_2)\gae{0}\reverse{\versor}_{\rigid_2}(\posjoint_2).
            \end{equation}
            This cooperative pointpair is the cornerstone of the idea to formulate collaborative geometric primitives for robotic systems involving more than two kinematic chains. 


\section{Method}
\label{sec:method}

    Geometric algebra enables the direct representation of geometric primitives within the algebra. Here, we show how to include these geometric primitives in the control objective by deriving control strategies based on similarity transformations. 

    
    {\color{black} 

    \subsection{Cooperative Task Space Modeling}
    \label{sub:cooperative_task_space_modeling}

        Suppose we have a robotic system that consists of $n$ parallel kinematic chains. Each of these kinematic chains has $m_i$ degrees of freedom, such that the cooperative system has $m=m_1+\ldots + m_n$ degrees of freedom. Hence, the joint space configuration of the cooperative system $\posjoint \in \mathbb{R}^m$ is composed as 
        \begin{equation}\label{eq:joint_space_configuration}
            \posjoint = \transpose{\begin{bmatrix} \transpose{\posjoint}_1 & \transpose{\posjoint}_2 & \cdots & \transpose{\posjoint}_n \end{bmatrix}}.
        \end{equation}
        The approach for modeling this cooperative system can then be described as a function $f: \mathbb{R}^m \to \group{\similarity}$ that maps a given joint angle configuration $\posjoint \in \mathbb{R}^m$ to an element of the group of similarity transformations $\versor_{\similarity c}(\posjoint) \in \group{\similarity}$. We call $\versor_{\similarity c}(\posjoint)$ the cooperative similarity transformation. We schematically depict the function $f$ in Figure~\ref{fig:flow_diagram}. The calculation relies on two intermediary representations. The first one are the points $P_i(\posjoint_i)$ that stand for the end-effector points of the individual kinematic chains. We use these points to calculate a cooperative geometric primitive $X_c(\posjoint)$ that we then use to calculate $\versor_{\similarity c}(\posjoint)$. We will give more details on the involved mathematics in the following sections. Instead, in this section, we focus on the high-level concept given by the modeling of cooperative task spaces via the cooperative similarity transformation. 

        \begin{figure}[!ht]
            {\color{black} 
            \centering
            \begin{tikzpicture}
                \node [] at (0,4) {$f :$};
                \node [] at (0.5,4) {$\mathbb{R}^m$};
                \node [] at (7,4) {$\group{\similarity}$};

                \node [] at (0.5,1.5) {$\posjoint$};

                \node [] at (2,0) {$\posjoint_n$};
                \node [] at (2,1) {\vdots};
                \node [] at (2,2) {$\posjoint_2$};
                \node [] at (2,3) {$\posjoint_1$};

                \node [] at (3.25,0) {$P_n(\posjoint_n)$};
                \node [] at (3.25,1) {\vdots};
                \node [] at (3.25,2) {$P_2(\posjoint_2)$};
                \node [] at (3.25,3) {$P_1(\posjoint_1)$};

                \node [] at (5.25,1.5) {$X_c(\posjoint)$};

                \node [] at (7,1.5) {$\versor_{\similarity c}(\posjoint)$};

                \draw [arrow] (1,4) -- (6.25,4);
                
                \draw [arrow] (1,1.5) -- (1.25,1.5) -- (1.25,0) -- (1.75,0);
                \draw [arrow] (1,1.5) -- (1.25,1.5) -- (1.25,1) -- (1.75,1);
                \draw [arrow] (1,1.5) -- (1.25,1.5) -- (1.25,2) -- (1.75,2);
                \draw [arrow] (1,1.5) -- (1.25,1.5) -- (1.25,3) -- (1.75,3);

                \draw [arrow] (2.25,0) -- (2.75,0);
                \draw [arrow] (2.25,1) -- (2.75,1);
                \draw [arrow] (2.25,2) -- (2.75,2);
                \draw [arrow] (2.25,3) -- (2.75,3);

                \draw [arrow] (4,0) -- (4.25,0) -- (4.25,1.5) -- (4.75,1.5);
                \draw [arrow] (4,1) -- (4.25,1) -- (4.25,1.5) -- (4.75,1.5);
                \draw [arrow] (4,2) -- (4.25,2) -- (4.25,1.5) -- (4.75,1.5);
                \draw [arrow] (4,3) -- (4.25,3) -- (4.25,1.5) -- (4.75,1.5);

                \draw [arrow] (5.75,1.5) -- (6.25,1.5);
            \end{tikzpicture}
            \caption{\color{black} Forward kinematics function for cooperative systems using similarity transformations.}
            \label{fig:flow_diagram}
            }
        \end{figure}
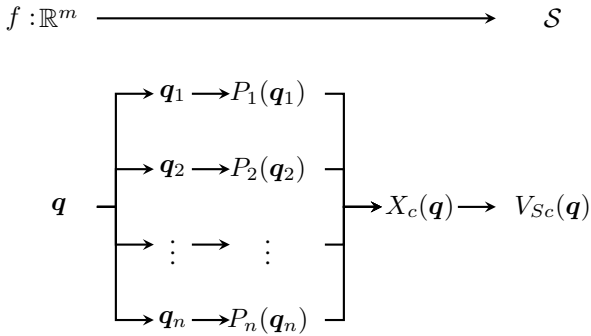

        Usually, the forward kinematics of a single kinematic chain are described by the function $g: \mathbb{R}^m \to SE(3)$, where $SE(3)$ is the Lie group of special Euclidean transformations, i.e. rotations and translations. This group is typically represented by homogeneous transformation matrices $\bm{T} \in \mathbb{R}^{4\times 4}$, making it a matrix Lie group. The matrix obtained by the function $g$ is usually called the end-effector pose of the robot and is used in various control algorithms, such as impedance/admittance control. We claim that the function $f$ that we propose in this article extends the function $g$ in terms of how it is used in control and optimization algorithms to cooperative systems. More concretely, the function $g$ can be considered as special case of the function $f$ in the case of $n=1$, since $SE(3) \subset \group{\similarity}$. This equivalence of functions $f$ and $g$ is ensured by them having the same mathematical properties. Namely, $f$ is continuous, smooth, non-linear, non-injective and non-surjective, but especially $f$ is well-defined. This last property means that for every joint angle configuration $\posjoint$ there exists a unique cooperative similarity transformation $\versor_{\similarity c}(\posjoint)$. We highlight this property, because it means we can use the cooperative similarity transformation, and by extension the cooperative geometric primitive, to obtain a new joint configuration (via e.g. inverse kinematics). As the joint configuration $\posjoint$ contains all kinematic chains without assigning them any specific priority, our modeling approach yields control commands for all kinematic chains simultaneously. It also means that the cooperative geometric primitive is not a constraint that is imposed on the system, but on the contrary, is the \emph{state} of the system.



        }


    
    \subsection{Cooperative Geometric Primitives}
    \label{sub:cooperative_geometric_primitives}
        We assume that the task-space modeling of the robotic system is done via one of the geometric primitives of CGA. We call this primitive the cooperative geometric primitive and denote it as $X_c(\posjoint)$.
        \begin{figure}[!ht]
            \centering
            \begin{subfigure}[t]{0.49\linewidth}
                \centering
                \includegraphics[height=2.75cm]{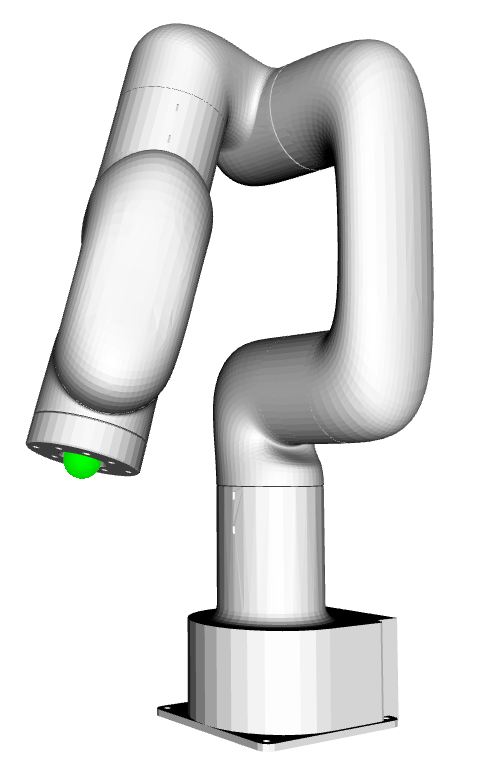}
                \caption{End-effector point of a Ufactory Lite6 robot.}
            \end{subfigure}
            \begin{subfigure}[t]{0.49\linewidth}
                \centering
                \includegraphics[height=2.75cm]{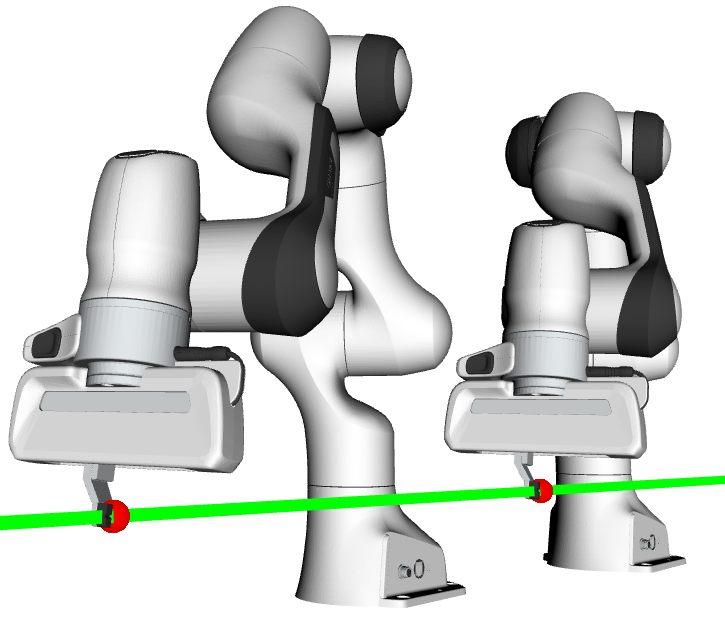}
                \caption{The cooperative dual task space for two Franka robots.}
            \end{subfigure}
            \\
            \begin{subfigure}[t]{0.49\linewidth}
                \centering
                \includegraphics[height=2.75cm]{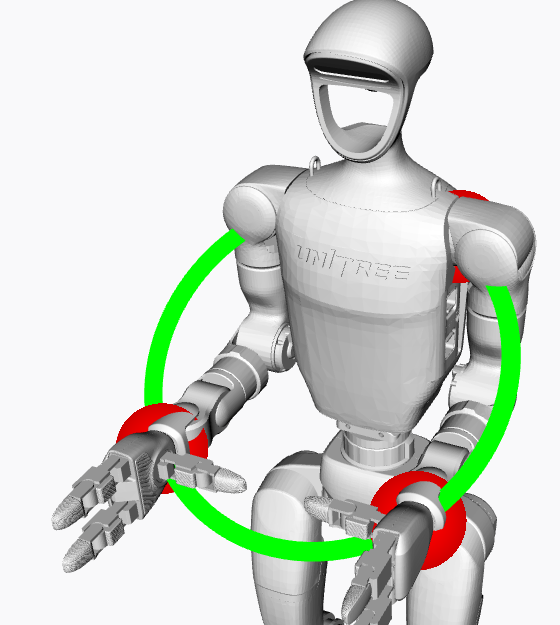}
                \caption{The cooperative circle for a humanoid formed by its two wrists and a point on the torso.}
            \end{subfigure}
            \begin{subfigure}[t]{0.49\linewidth}
                \centering
                \includegraphics[height=2.75cm]{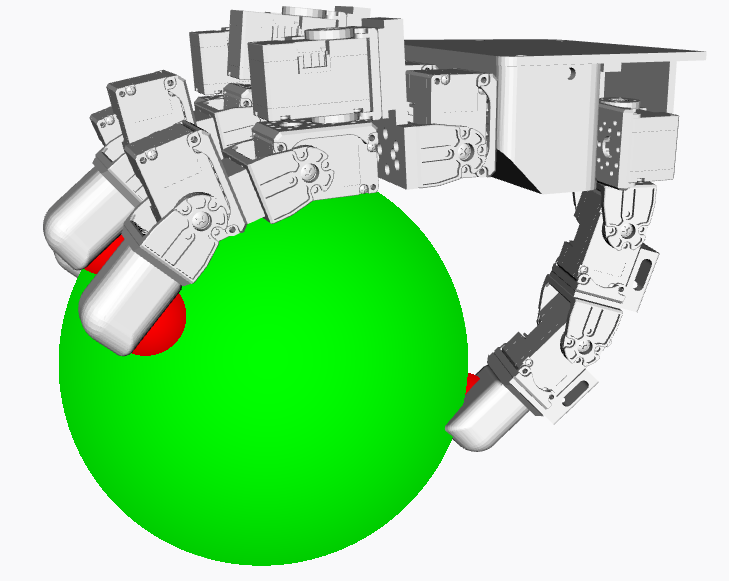}
                \caption{The cooperative sphere of the four fingers of the Leap Hand.}
            \end{subfigure}
            \caption{Cooperative geometric primitives for multiple kinematic chains. The red points indicate the reference points on the robots that are considered for the cooperative geometric primitive, which is then shown in green.}
            \label{fig:multi_robot_setup_using_geometric_primitives_}
        \end{figure}
        %
        $X_c(\posjoint)$ extends the ideas around the cooperative pointpair that was shown in Equation~\eqref{eq:cooperative pointpair} to $n$ parallel kinematic chains. We use the properties of the outer product of points to find the cooperative geometric primitive $X_c(\posjoint)$ formed by the end points of those kinematic chains
        \begin{equation}\label{eq:cooperative_geometric_primitive}
            X_c(\posjoint) = \bigwedge_{i=1}^n P_i(\posjoint) = \bigwedge_{i=1}^n M_i(\posjoint_i)\gae{0}\reverse{M}_i(\posjoint_i),
        \end{equation}
        where the conformal points $P_i$ correspond to the end points of the kinematic chains. The trivial case of one kinematic chain is a point. Two kinematic chains give a pointpair, three a circle, and four a sphere. Examples of the cooperative geometric primitives for varying numbers of kinematic chains can be seen in Figure~\ref{fig:multi_robot_setup_using_geometric_primitives_}. Here, we use three manipulators as a proxy for a three-fingered hand to highlight the mathematical equivalence between multiple robot manipulators and robotic hands with multiple fingers. Note that, instead of a pointpair, we could also use a line and instead of a circle, a plane. Both would only require multiplication with the point at infinity and would lead to a relaxed control law that allowed more degrees of freedom. In general, this construction of cooperative geometric primitives holds for geometric algebras of different dimensions, i.e. for any number of $n$. In our case, using $\cga$, which is a 5-dimensional algebra, the highest dimensional geometric primitive is a sphere. A sphere is constructed from $n=4$ points, i.e. we limit our view in this section on a maximum of four parallel kinematic chains. In theory, however, using higher dimensional geometric algebras, the derived formulas could be extended to $4+n$ parallel kinematic chains. We will go into more detail on this point in the discussion in Section~\ref{sub:extension_to_n_kinematic_chains}.

        The analytic Jacobian for the cooperative geometric primitive $\jacobian^A_{c}(\posjoint)$ can be found as the derivative of $X_c(\posjoint)$ w.r.t. the joint angles, i.e. 
        \begin{equation}\label{eq:cooperative_geometric_primitive_jacobian}
            \jacobian^A_{c}(\posjoint) = \frac{\partial}{\partial \posjoint} X_c(\posjoint) = \mat{\jacobian^{A}_{c,1}(\posjoint) & \ldots & \jacobian^{A}_{c,n}(\posjoint)},
        \end{equation}
        where
        \begin{equation}\label{eq:cooperative_geometric_primitive_jacobian_i}
            \jacobian^A_{c,i}(\posjoint) = P_1 \outer \jacobian_{P,i} \outer \ldots \outer P_n, 
        \end{equation}
        and 
        \begin{equation}
            \jacobian_{P,i} = \jacobian_i^A \gae{0} \reverse{M}_i + M_i \gae{0} \reverse{\jacobian^A_i}.
        \end{equation}
        %

        {\color{black} The cooperative geometric primitives are an intermediate representation that describe the inherent task space state of the system. In some sense, they represent the cooperative task space geometry for multiple parallel kinematic chains. The actual representation that we derive in this work is based on similarity transformations derived from the geometric primitives. As we show in the following section, these similarity transformations can then be seen as a direct extension of the end-effector pose of a single kinematic chain in classical control and optimization.}
       

    
    \subsection{Similarity Transformations}
    \label{sub:similarity_transformations}
\begin{figure}[!ht]
            \centering
            \begin{subfigure}[t]{0.45\linewidth}
                \includegraphics[width=\linewidth]{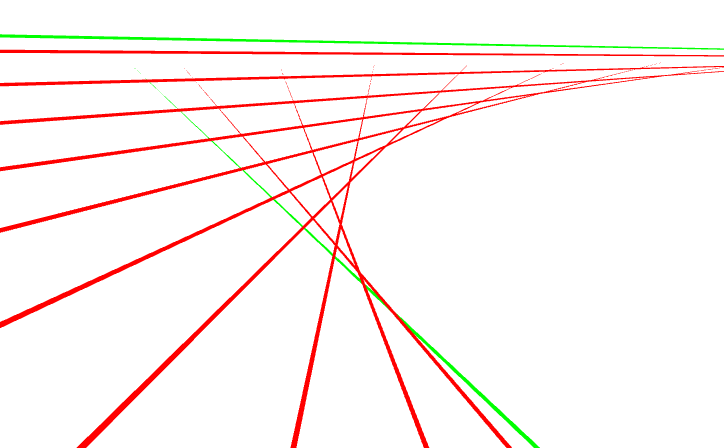}
                \caption{Line.}
            \end{subfigure}
            \begin{subfigure}[t]{0.45\linewidth}
                \includegraphics[width=\linewidth]{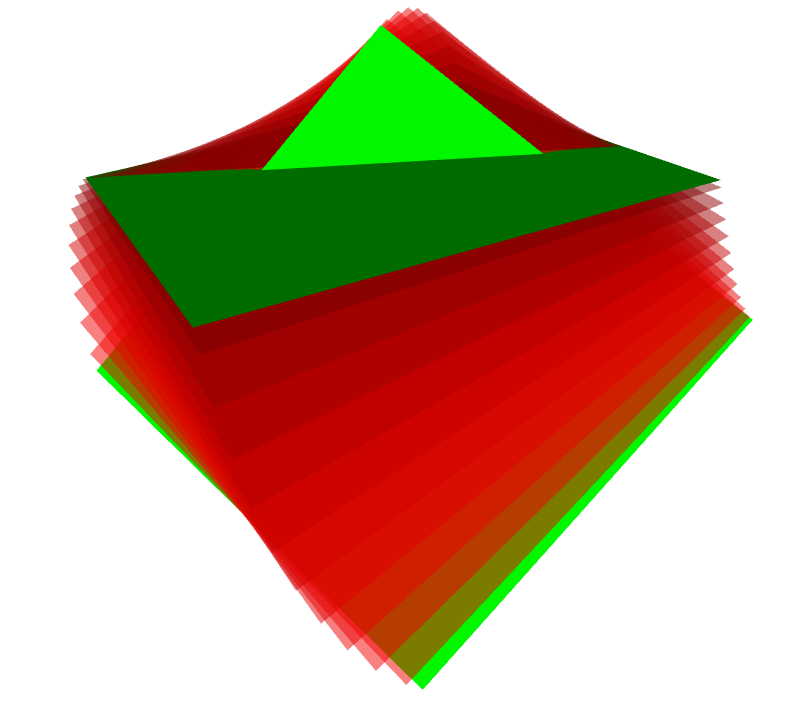}
                \caption{Plane.}
            \end{subfigure}
            \begin{subfigure}[t]{0.25\linewidth}
                \includegraphics[width=\linewidth]{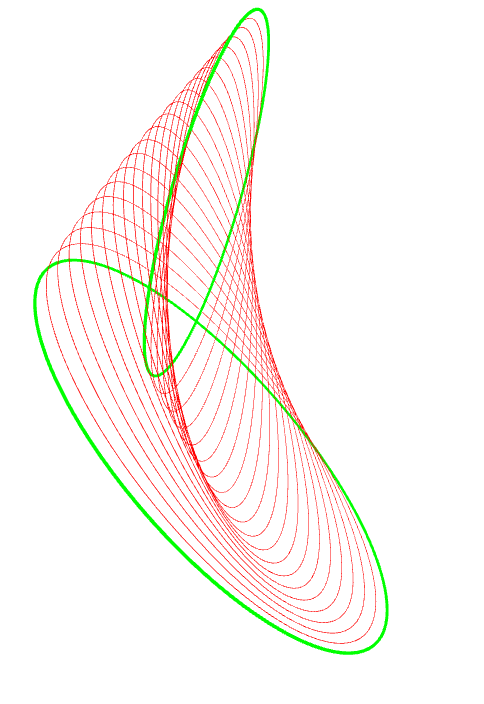}
                \caption{Circle.}
            \end{subfigure}
            \begin{subfigure}[t]{0.65\linewidth}
                \includegraphics[width=\linewidth]{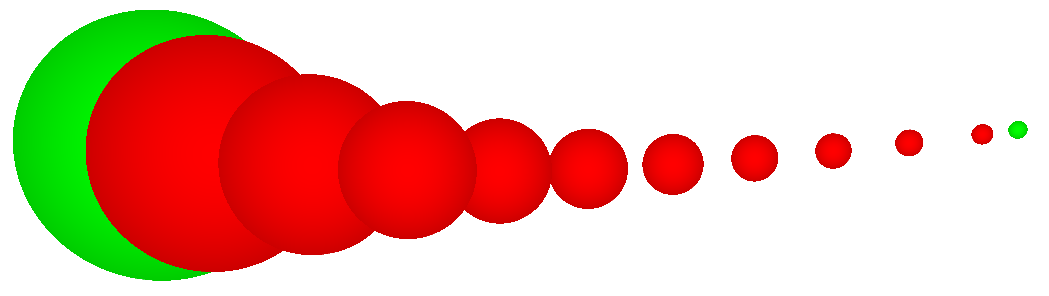}
                \caption{Sphere.}
            \end{subfigure}
            \caption{Interpolation of different geometric primitives using similarity transformations.}
            \label{fig:interpolation_of_different_geometric_primtives_using_similarity_transformations_}
        \end{figure}

        Similarity transformations form a Lie group $\group{\similarity}$ that is a seven dimensional manifold. In the context of this work, they can be used for transforming the cooperative geometric primitives. We define the canonical decomposition of a general similarity transform versor to be
        \begin{equation}\label{eq:similarity_transform}
            \versor_\similarity = \versor_\translator \versor_\rotor \versor_\dilator.
        \end{equation}
        Elements of the associated Lie algebra $\algebra{\bivector}_\similarity$ can be found via the logarithmic map
        \begin{equation}\label{eq:similarity_transform_logarithmic_map}
            \begin{split}
                B_{\similarity} 
                & = \log \left( \versor_{\similarity} \right) 
                \\
                & = \log (\versor_\translator) + \log (\versor_\rotor) + \log (\versor_\dilator).
            \end{split}
        \end{equation}
        More details on the transformation groups and their logarithmic maps can be found in Section~\ref{ssub:transformation_groups_in_cga} and Appendix~\ref{sec:transformations_groups_in_cga}. In the same way the Lie algebra of quaternions has been used for interpolation between group elements in applications such as spherical linear interpolation, we can use elements $\bivector_{\similarity} \in \algebra{\bivector}_{\similarity}$ for interpolating a similarity transformation between two geometric primitives. We show this interpolation in Figure \ref{fig:interpolation_of_different_geometric_primtives_using_similarity_transformations_}. 

        Given the two geometric primitives, $X_1$ and $X_2$, we now derive the similarity transformation $\versor_{\similarity} (X_1, X_2)$ that fulfills
        \begin{equation}\label{eq:similarity_transformation}
            X_2 = \versor_{\similarity} (X_1, X_2) X_1 \reverse{\versor}_{\similarity} (X_1, X_2).
        \end{equation}
        Its bivector $\bivector_{\similarity}(X_1, X_2)$ found from the logarithmic map can then be used in the optimization problem given in Equation~\eqref{eq:optimization_problem_formulation}. A general formula to find a versor $\versor_{\conformal} (X_1, X_2)$ that transforms $X_1$ to $X_2$ was derived in \cite{lasenbyCalculatingRotorConformal2019}
        \begin{equation}\label{eq:versor_between_primitives}
            \versor_{\conformal} (X_1, X_2) = c^{-1}\left( 1 + X_2 X_1 \right),
        \end{equation}
        where $c$ is a normalization constant that depends on the geometric primitives. However, using this general formula, $\versor_{\conformal} (X_1, X_2) \in \group{\conformal}$, i.e. the resulting versor will be an element of the conformal transformation group $\group{\conformal}$, not of the similarity transformation group $\group{\similarity}$. In order to restrict the versor to $\group{\similarity}$, we present the necessary derivations that give a similarity transformation between pairs of geometric primitives according to Equation~\eqref{eq:similarity_transformation}. Note that, this clearly shows that the transformation between primitives is not unique, as was already stated in \cite{lasenbyCalculatingRotorConformal2019}.

        
        \subsubsection{Point to Point}
        \label{ssub:point_to_point}
            The simplest case is transforming one point to another. Since points neither have an orientation nor a size, the rotor $\versor_{\rotor}$ and dilator $\versor_{\dilator}$ simplify to the identity transformation, i.e. $\versor_{\rotor} = \versor_{\dilator} = 1$. Regarding the translation versor, given the two points $\point_1$ and $\point_2$, it is found as 
            \begin{equation}\label{eq:point_to_point_translation_versor}
                \begin{split}
                    \versor_{\similarity} (\point_1, \point_2) 
                    = & \versor_{\translator} (\point_1, \point_2) 
                    \\
                    = & \exp \big( (\point_2 - \point_1) \outer \gae{\infty} \big).
                \end{split}
            \end{equation}

        
        \subsubsection{Line to Line}
        \label{ssub:line_to_line}
            Lines have a flat geometry, i.e. they are part of Euclidean geometry, which means a transformation between two lines can be described purely by rotations and translations. Hence, the versor resulting from Equation \eqref{eq:versor_between_primitives} will only contain a translation and a rotation, i.e. $\versor_{\similarity} (\line_1,\line_2) \in \group{\rigid}$.

        
        \subsubsection{Plane to Plane}
        \label{ssub:plane_to_plane}
            Using the same argumentation as for the case of line to line transformations, the transformation from one plane to another can be found using Equation \eqref{eq:versor_between_primitives}. The resulting versor will only contain a translation and a rotation, i.e. $\versor_{\similarity} (\plane_1,\plane_2) \in \group{\rigid}$.

        
        \subsubsection{Circle to Circle}
        \label{ssub:circle_to_circle}
            Circles require all three versors in the similarity transformation: translations, rotations and dilations. We can directly compute the dilation versor based on the radii $r_1$ and $r_2$ of $X_1$ and $X_2$
            \begin{equation}\label{eq:circle_dilator}
                \versor_{\dilator}(X_1,X_2) = \exp\big(r_2r_1^{-1}\gae{0\infty}\big), 
            \end{equation}
            where that radius of a circle can be computed as
            \begin{equation}\label{eq:circle_radius}
                r = \Big\|  \left( X \inner \gae{\infty}\right) \inverse{X} \Big\|.
            \end{equation}
            Note that the inner expression takes the form of the general subspace projection that we showed in Equation \eqref{eq:projection}. It can thus be interpreted as projecting infinity to the subspace of the circle and then evaluating the resulting norm. 

            Next, we compute the rotation versor between the circles by using the normals $N_1$ and $N_2$ of the planes that the circles lie. These planes can be found via $\plane = \circle \outer \gae{\infty}$ and the normal of plane $E$ is computed as 
            \begin{equation}\label{eq:plane_normal}
                \normal = \dual{E} - \frac{1}{2}(\dual{E} \inner \gae{0}) \gae{\infty}. 
            \end{equation}
            We then insert there normals in Equation \eqref{eq:versor_between_primitives}, which yields
            \begin{equation}\label{eq:circle_plane_rotor}
                \begin{split}
                    \versor_{\rotor}(X_1,X_2) 
                    &= \inverse{c} (1 + N_1N_2) 
                    \\
                    &= \inverse{c} \Big(1 + (N_2 \inner N_1) - (N_2 \outer N_2)\Big).
                \end{split}
            \end{equation}
            Note that this is the geometric algebra variant of the standard expression for computing a unit quaternion between unit vectors.

            Lastly, we compute the translation versor via the center points of the circles. The center point of circles can be found as
            \begin{equation}\label{eq:circle_sphere_center_points}
                \point(X) = X\gae{\infty}X,
            \end{equation}
            which in this case amounts to reflecting infinity in the circle or sphere. Given the center points, the translation versor can then be computed as was indicated for points in Equation~\eqref{eq:point_to_point_translation_versor}. Note that, for computing the center of the cooperative circle, we first apply the transformation given by $\versor_{\rotor}(X_1,X_2)\versor_{\dilator}(X_1,X_2)$, because this translation changes the center point of the circle. 

        
        \subsubsection{Sphere to Sphere}
        \label{ssub:sphere_to_sphere}
            For spheres, the rotation versor is the identity, because spheres do not have an intrinsic orientation. The dilation and translation versors can then be calculated using the same strategy as for the circles. The computation of the dilation versor follows Equation \eqref{eq:circle_dilator}, where the radius of a sphere is also computed in the same was as for the circle, i.e. using Equation \eqref{eq:circle_radius}. We then use Equation \eqref{eq:circle_sphere_center_points} to compute their center points and afterwards Equation \eqref{eq:point_to_point_translation_versor} to obtain the translation versor.

    
    \subsection{Cooperative Task Space Kinematics}
    \label{sub:cooperative_similarity_control}

        First, we define the cooperative similarity transformation $\versor_{\similarity c}(\posjoint)$. Similarly to how an end-effector pose represents the transformation of the coordinate system at the origin to the end-effector, it is expressed as the similarity transformation w.r.t. a unit geometric primitive at the origin. We denote this primitive as $X_u$. In section \ref{sub:similarity_transformations}, we have already presented how to find similarity transformations between two geometric primitives of the same kind. Accordingly, the versor $\versor_{\similarity c}(\posjoint)$ can be found following the explanations around Equation~\eqref{eq:similarity_transform} as
        \begin{equation}\label{eq:cooperative_similarity_versor}
            \versor_{\similarity c}(\posjoint) = \versor_\similarity \left( X_u, X_c(\posjoint) \right). 
        \end{equation}
        This versor then fulfills $X_c(\posjoint) = \versor_{\similarity c}(\posjoint) X_u \reverse{\versor}_{\similarity c}(\posjoint)$. It is a direct equivalent to the end-effector motor found by the forward kinematics in Equation~\eqref{eq:forward_kinematics}, albeit it represents a cooperation between multiple kinematic chains and not just a single one. {\color{black} Hence, the control laws that we propose, based on this cooperative similarity transformation, yield joint space commands for all kinematic chains simultaneously, without adopting principles such as leader/follower. As mentioned before, the cooperative geometric primitives are an intermediate representation, whereas the cooperative similarity transformations are the actual representation that the rest of our formulations are based on. Therefore, in the following sections, we will only mention the similarity transformations and only assume the cooperative geometric primitives implicitly as the task space geometries. Since these task space geometries are more of an abstract concept rather than a physical representation, we collect the different task space geometries in Table~\ref{tab:cooperative_similarity_task_spaces_}. This table also shows the corresponding controllable bivector space, depending on the number of kinematic chains, and formulated w.r.t. the cooperative similarity transformation, not the origin. Note that the corresponding controllable bivector spaces all have a dimension that is strictly smaller than 7, which means that all systems have a geometric nullspace.}

        \begin{table*}[!ht]
        {\color{black}
            \centering
            \caption{\color{black} Description of the cooperative similarity task spaces for 1, 2, 3, and 4 parallel kinematic chains. In the case of 2 and 3 kinematic chains, two different cooperative geometric primitives are possible by including the basis vector $\gae{\infty}$ via outer product. This changes the controllable bivector space to be smaller, which essentially means that the system is less constrained and gain additional degrees of freedom.}
            \small
            \begin{tabularx}{\linewidth}{lllX}
                \toprule
                    \textbf{Joint Space}
                    & \multicolumn{2}{l}{\textbf{Cooperative Geometric Primitive}} 
                    & \textbf{Controllable Bivector Space}
                    \\
                \toprule
                    $\posjoint = \transpose{\begin{bmatrix} \transpose{\posjoint}_1 \end{bmatrix}}$
                    & Point
                    & $X_c(\posjoint) = P(\posjoint)$
                    & $\gae{1\infty}, \gae{2\infty}, \gae{3\infty}$
                    \\
                \midrule
                    $\posjoint = \transpose{\begin{bmatrix} \transpose{\posjoint}_1 & \transpose{\posjoint}_2 \end{bmatrix}}$
                    & Point Pair
                    & $X_c(\posjoint) = P(\posjoint_1) \outer P(\posjoint_2)$
                    & $\gae{12}, \gae{23}, \gae{0\infty}, \gae{1\infty}, \gae{2\infty}, \gae{3\infty}$
                    \\
                    & Line
                    & $X_c(\posjoint) = P(\posjoint_1) \outer P(\posjoint_2) \outer \gae{\infty}$
                    & $\gae{12}, \gae{23}, \gae{1\infty}, \gae{3\infty}$
                    \\
                \midrule
                    $\posjoint = \transpose{\begin{bmatrix} \transpose{\posjoint}_1 & \transpose{\posjoint}_2 & \transpose{\posjoint}_3 \end{bmatrix}}$
                    & Circle
                    & $X_c(\posjoint) = P(\posjoint_1) \outer P(\posjoint_2) \outer P(\posjoint_3)$
                    & $\gae{13}, \gae{23}, \gae{0\infty}, \gae{1\infty}, \gae{2\infty}, \gae{3\infty}$
                    \\
                    & Plane
                    & $X_c(\posjoint) = P(\posjoint_1) \outer P(\posjoint_2) \outer P(\posjoint_3) \outer \gae{\infty}$
                    & $\gae{13}, \gae{23}, \gae{3\infty}$
                    \\
                \midrule
                    $\posjoint = \transpose{\begin{bmatrix} \transpose{\posjoint}_1 & \transpose{\posjoint}_2 & \transpose{\posjoint}_3 & \transpose{\posjoint}_4 \end{bmatrix}}$
                    & Sphere
                    & $X_c(\posjoint) = P(\posjoint_1) \outer P(\posjoint_2) \outer P(\posjoint_3) \outer P(\posjoint_4)$
                    & $\gae{0\infty}, \gae{1\infty}, \gae{2\infty}, \gae{3\infty}$
                    \\
                \bottomrule
            \end{tabularx}
            \label{tab:cooperative_similarity_task_spaces_}
        }
        \end{table*}

        Most commonly, optimization solvers and control algorithms make use of one of various Jacobians that can be found for the cooperative similarity versor $\versor_{\similarity c}(\posjoint)$ and its bivector $\bivector_{\similarity c}(\posjoint) = \log (\versor_{\similarity c}(\posjoint))$. The usual kinematic modeling of robots makes a distinction between the analytic and the geometric Jacobian. Hence, we derive the analytic similarity Jacobian $\jacobian^A_\similarity(\posjoint)$, and the geometric similarity Jacobian $\jacobian^G_\similarity(\posjoint)$. We also show the bivector similarity Jacobian $\jacobian^\bivector_\similarity(\posjoint)$ that relates changes in the joint positions $\posjoint$ to changes in the similarity bivector $\bivector_{\similarity c}(\posjoint)$. These Jacobians can then be used to find the gradient, approximate the Hessian, or compute a control signal, depending on the chosen control approach. 

        As per the usual definition, the analytic similarity Jacobian can be found from the partial derivatives of the cooperative similarity versor w.r.t. $\posjoint$, i.e. 
        \begin{equation}\label{eq:analytic_similarity_jacobian}
            \begin{split}
                \jacobian^A_{\similarity c} (\posjoint)
                = & \frac{\partial }{\partial \posjoint} \versor_{\similarity c} (\posjoint)
                \\
                = & \jacobian^A_{\translator c}(\posjoint) \versor_{\rotor c}(\posjoint) \versor_{\dilator c}(\posjoint) 
                \\
                & + \versor_{\translator c}(\posjoint) \jacobian^A_{\rotor c}(\posjoint) \versor_{\dilator c}(\posjoint) 
                \\
                & + \versor_{\translator c}(\posjoint) \versor_{\rotor c}(\posjoint) \jacobian^A_{\dilator c}(\posjoint), 
            \end{split}
        \end{equation}
        where $\jacobian^A_{\translator c}(\posjoint), \jacobian^A_{\rotor c}(\posjoint),$ and $\jacobian^A_{\dilator c}(\posjoint)$ are the Jacobians of the translation, rotation and dilation versors, respectively. We omit the full derivation here for conciseness, which can be found in Appendix~\ref{sec:jacobians}.

        The geometric similarity Jacobian can be used to formulate control laws such as differential kinematics or impedance/admittance control. Via the group constraint and from the relationship between the end-effector Jacobians shown in Equation~\eqref{eq:jacobian_relationship}, it is easy to see that the following relationship holds:
        \begin{equation}\label{eq:geometric_similarity_jacobian}
            \jacobian^G_{\similarity c}(\posjoint) = -2\reverse{\versor}_{\similarity c}(\posjoint) \jacobian^A_{\similarity c}(\posjoint).
        \end{equation}
        {\color{black} Note that, similar to the traditional geometric Jacobian, the geometric similarity Jacobian is also depending on a frame of reference. The above relationship represents the geometric similarity Jacobian w.r.t. the cooperative similarity transformation, as opposed to, for example, the origin. The equivalent end-effector Jacobian from classical single kinematic chain modeling is sometimes referred to as the \emph{body Jacobian}.}

        Lastly, the bivector similarity Jacobian $\jacobian^\bivector_\similarity(\posjoint)$ is found as the partial derivates of a cooperative similarity bivector $\bivector_{\similarity c}(\posjoint)$ w.r.t. $\posjoint$ {\color{black} by applying the chain rule}
        \begin{equation}\label{eq:cooperative_bivector_jacobian}
            \begin{split}
                \jacobian_{\bivector c}(\posjoint) 
                & {\color{black} = \frac{\partial }{\partial \posjoint} \bivector_{\similarity c}(\posjoint) = \frac{\partial }{\partial \posjoint} \log \big( \versor_{\similarity c}(\posjoint) \big)}
                \\
                & {\color{black} = \frac{\partial }{\partial \versor} \log \big(\versor_{\similarity c}(\posjoint)\big) \frac{\partial }{\partial \posjoint} \versor_{\similarity c}(\posjoint) }
                \\
                & = \jacobian_{\group{\similarity} \to \algebra{\bivector}_{\similarity}}(\posjoint) \jacobian^A_{\similarity c}(\posjoint),
            \end{split}
        \end{equation}
        where $\jacobian_{\group{\similarity} \to \algebra{\bivector}_{\similarity}}(\posjoint)$ is the Jacobian of the logarithmic map of the similarity transformation group. The bivector similarity Jacobian can then be used in applications such as optimization-based inverse kinematics. 


    {\color{black}

    
    \subsection{Cooperative Task Space Dynamics}
    \label{sub:similarity_dynamics}
        
        In this section, we present the derivation of the task space dynamics defined by the cooperative similarity transformations. We start from the well-known joint space dynamics
        \begin{equation}\label{eq:joint_space_dynamics}
            \bm{M}(\posjoint)\accjoint + \bm{C}(\posjoint,\veljoint)\veljoint + \bm{g}(\posjoint) = \bm{\tau} - \bm{\tau}_{\text{ext}},
        \end{equation}
        where $\bm{M}(\posjoint)$ is known as the inertia or generalized mass matrix, $\bm{C}(\posjoint,\veljoint)$ is representing Coriolis/centrifugal forces, $\bm{g}(\posjoint)$ stands for the gravitational forces, $\bm{\tau}$ is the vector of joint torques and $\bm{\tau}_{ext}$ are the external torques. Analogously to the classical formulation of the task space dynamics using the geometric Jacobian $\jacobian^G(\posjoint)$ of a single kinematic chain, we can use the geometric similarity Jacobian $\jacobian^G_{\similarity c}(\posjoint)$ to derive the dynamics of the cooperative task space. Recall that the geometric similarity Jacobian defines the relationship between the joint and task space velocities, i.e. 
        \begin{equation}\label{eq:similarity_twist}
           \twist_{\similarity} = \jacobian^G_{\similarity c}(\posjoint)\veljoint,
        \end{equation}
        where $\twist_{\similarity} \in \algebra{B}_{\similarity}$ is a twist-like quantity. Taking the derivative w.r.t. time yields
        \begin{equation}\label{eq:similarity_task_space_acceleration}
            \dot{\twist}_{\similarity} = \jacobian^G_{\similarity c}(\posjoint)\veljoint + \bm{\mathcal{\dot{J}}}^G_{\similarity c}(\posjoint,\veljoint)\posjoint.
        \end{equation}
        Using this result and following the derivations from the classical operational space formulation, we find the similarity task space dynamics as 
        \begin{equation}\label{eq:similarity_task_space_dynamics}
            \bm{M}_{\similarity c}(\posjoint) \dot{\twist}_{\similarity}
            + \bm{C}_{\similarity c}(\posjoint,\veljoint)
            + \bm{g}_{\similarity c}(\posjoint)
            = \wrench_{\similarity}, 
        \end{equation}
        where the similarity inertia matrix is found as
        \begin{equation}\label{eq:similarity_inertia_matrix}
            \bm{M}_{\similarity c}(\posjoint) = \Big(\big( \jacobian^G_{\similarity c} \big) \inverse{\bm{M}} \transpose{\big( \jacobian^G_{\similarity c} \big)}\Big)^{-1}, 
        \end{equation}
        the Coriolis/centrifugal forces as
        \begin{equation}\label{eq:similarity_coriolis_forces}
            \bm{C}_{\similarity c}(\posjoint,\veljoint) = \bm{M}_{\similarity c} \Big( \jacobian^G_{\similarity c} \inverse{\bm{M}} \bm{C} - \bm{\mathcal{\dot{J}}}^G_{\similarity c} \veljoint \Big),
        \end{equation}
        and the gravitational forces as
        \begin{equation}\label{eq:similarity_gravitational_force}
            \bm{g}_{\similarity c}(\posjoint) = \bm{M}_{\similarity c} \jacobian^G_{\similarity c} \inverse{\bm{M}} \bm{g}.
        \end{equation}
        %

    
    \subsection{Cooperative Similarity Control}
    \label{sub:similarity_objective}

        We base the formulation of the control objective for cooperative manipulation control on similarity transformations. We express the control objective as an optimization problem
        \begin{equation}\label{eq:optimization_problem_formulation}
            \min \Big\| \log \Big(\reverse{\versor}_{\similarity c}(\posjoint) \versor_{\similarity d}\Big)\Big\|, 
        \end{equation}
        where we use the logarithmic map of the versor representation, as explained in Section~\ref{ssub:transformation_groups_in_cga}. The similarity transformation $\versor_{\similarity d}$ is a desired similarity transformation that can be found using the unit geometric primitive
        \begin{equation}\label{eq:desired_similarity_versor}
            \versor_{\similarity d} = \versor_\similarity \left( X_u, X_d \right), 
        \end{equation}
        where $X_d$ is a desired geometric primitive. Note that the desired similarity transformation can also be set directly without the desired geometric primitive, but latter facilitates the process, since it can be visualized. For any pair of geometric primitives $X_c(\posjoint)$ and $X_d$, there is at least one similarity transformation $\versor_\similarity$ that transforms $X_c(\posjoint)$ into $X_d$. Solving the optimization problem in Equation~\ref{eq:optimization_problem_formulation} essentially means making $\versor_\similarity$ the identity transformation. Here, we assume that $X_c(\posjoint)$ and $X_d$ are of the same type. In general, this does not need to be the case, but is outside the scope of this work. We add more insights on mixing geometric primitives in the discussion section.

        The versor from Equation~\eqref{eq:optimization_problem_formulation} can be written as 
        \begin{equation}\label{eq:similarity_identity}
            \versor_\similarity(X_c(\posjoint), X_d) = \versor_{\similarity cd}(\posjoint) = \reverse{\versor}_{\similarity c}(\posjoint) \versor_{\similarity d},
        \end{equation}
        which is equivalent to the formulation of inverse kinematics for a single kinematic chain based on the end-effector motor. Note that, instead of using Equation~\eqref{eq:similarity_identity}, the versor $\versor_\similarity(X_c(\posjoint), X_d)$ can also directly be found following the explanations in Section \ref{sub:similarity_transformations}.
        From Equation~\eqref{eq:similarity_identity}, we can then find the bivector $\bivector_{\similarity cd}(\posjoint)$ via the logarithmic map that represents the task-space control signal, similar to a twist, 
        \begin{equation}\label{eq:cooperative_similarity_bivector}
            \bivector_{\similarity cd}(\posjoint) = \log \left( \versor_{\similarity cd}(\posjoint) \right).
        \end{equation}
        Note that, in general, this bivector can be seven-dimensional, since the similarity transformation group is a seven-dimensional manifold. 


        Here, we do not propose any particular solver for this optimization problem, but focus on the general modeling. In practice, this optimization problem can be solved using state-of-the-art methods from the literature around optimal control, impedance/admittance control or simply using differential kinematics. To account for this generality of the problem formulation, we give all the mathematical details that are necessary for these different control paradigms. In the following, we explain how this control problem can be embedded into existing control and optimization frameworks, in particular impedance control and optimal control.

    \subsection{Inverse Kinematics}
    \label{sub:inverse_kinematics}
        We formulate the optimization problem from Equation~\eqref{eq:optimization_problem_formulation} such that it becomes an inverse kinematics problem
        \begin{equation}\label{eq:similarity_inverse_kinematics}
            \posjoint^* = \arg\min_{\posjoint} \Big\| \bm{e}(\posjoint) \Big\|_2^2,
        \end{equation}
        where the error vector $\bm{e}(\posjoint)$ is 
        \begin{equation}\label{eq:similarity_error_vector}
            \bm{e}(\posjoint) = \log\Big( \reverse{\versor}_{\similarity c}(\posjoint) \versor_{\similarity d} \Big).
        \end{equation}
        To illustrate how this inverse kinematics problem can be solved using optimization-based approaches, we present a simple Gauss-Newton step
        \begin{equation}\label{eq:gauss_newton}
            \posjoint_{k+1} = \posjoint_{k} - \alpha \bm{H}(\posjoint_k) \bm{e}(\posjoint_k),
        \end{equation}
        that starts from an initial guess $\posjoint_0$ and iterates until convergence. The step size $\alpha$ and can be found using line search. The matrix $\bm{H}(\posjoint)$ is the Hessian matrix of Equation~\eqref{eq:similarity_inverse_kinematics}. It can be approximated as
        \begin{equation}\label{eq:hessian_approximation}
            \bm{H}(\posjoint) = \Big( \jacobian^\trsp_{\bivector e}(\posjoint_k) \jacobian_{\bivector e}(\posjoint_k) \Big)^{-1} \jacobian^\trsp_{\bivector e}(\posjoint_k),
        \end{equation}
        where the Jacobian $\jacobian_{\bivector e}(\posjoint_k)$ is found as
        \begin{equation}\label{eq:jacobian_similarity_error_vector}
            \jacobian_{\bivector e}(\posjoint_k) = \jacobian_{\group{\similarity} \to \algebra{\bivector}_{\similarity}}(\posjoint) \reverse{\jacobian}^A_{\similarity c}(\posjoint) \versor_{\similarity d},
        \end{equation}
        and follows the same derivation as the cooperative bivector Jacobian in Equation~\eqref{eq:cooperative_bivector_jacobian}.

        Instead of using the optimization-based inverse kinematics, we can also use a task space velocity based on the differential kinematics. For this we recall Equation~\eqref{eq:similarity_twist} that relates the joint velocity $\veljoint$ to the similarity twist $\twist_{\similarity}$ via the geometric similarity Jacobian. Hence, in order to obtain a joint-space command by using differential kinematics, we invert the geometric similarity Jacobian
        \begin{equation}\label{eq:differential_similarity_kinematics}
            \veljoint = \inverse{\Big(\jacobian^G_{\similarity c}(\posjoint)\Big)} \bivector_{\similarity,i}.
        \end{equation}
        The control scheme is formulated w.r.t. to the cooperative similarity transformation, i.e. the local frame. Note that this control formulation is identical to the usual way of doing differential kinematics. The only difference is that a cooperative task space of multiple kinematic chains is modeled, as opposed to a single one. Therefore, given a similarity twist, this control law will find joint velocities for all kinematic chains that are part of the cooperative system.

    
    \subsection{Impedance Control}
    \label{sub:derivation_impedance_control}
        The traditional control methods require a twist as a task-space control command. Here, the cooperative similarity bivector $\bivector_{\similarity cd}(\posjoint)$ from Equation~\eqref{eq:cooperative_similarity_bivector} can be used to define a corresponding similarity twist $\twist_\similarity$.

        We derive an impedance control law to regulate the system to a desired similarity transformation $\versor_{\similarity d}$. Assuming $\dot{\versor}_{\similarity d} = 0$, we find the error dynamics as 
        \begin{equation}\label{eq:similarity_error_dynamics}
            \begin{split}
                \bivector_{\similarity,e} & = \log \Big( \reverse{\versor}_{\similarity c}(\posjoint) \versor_{\similarity d} \Big),
                \\
                \dot{\bivector}_{\similarity,e} & = - \frac{1}{2} \twist_{\similarity c},
            \end{split}
        \end{equation}
        where $\twist_{\similarity c}$ is the similarity twist from Equation~\eqref{eq:similarity_twist}. 
        These error dynamics can then be used to define a simple control law for the desired similarity task space acceleration $\dot{\twist}_{\similarity d}(\posjoint)$ 
        \begin{equation}\label{eq:desired_similarity_acceleration}
            \dot{\twist}_{\similarity d}(\posjoint) = \bm{K} \bivector_{\similarity,e} + \bm{D} \dot{\bivector}_{\similarity,e}, 
        \end{equation}
        where the matrices $\bm{K}, \bm{D} \in \mathbb{R}^{7\times 7}$ are the stiffness and damping gains, respectively. Using the similarity task space dynamics from Equation~\eqref{eq:similarity_identity}, we then compute the desired similarity wrench $\wrench_{\similarity d}(\posjoint)$
        \begin{equation}\label{eq:desired_similarity_wrench}
            \wrench_{\similarity d}(\posjoint) = \bm{M}_{\similarity c}(\posjoint) \dot{\twist}_{\similarity d}(\posjoint) + \bm{C}_{\similarity c}(\posjoint,\veljoint) + \bm{g}_{\similarity c}(\posjoint), 
        \end{equation}
        which is then used to compute a joint torque control command $\bm{\tau}(\posjoint)$
        \begin{equation}\label{eq:operational_space_control}
            \bm{\tau}(\posjoint) = \Big(\jacobian_{\similarity c}^G(\posjoint)\Big)^{\trsp} \wrench_{\similarity d}(\posjoint).
        \end{equation}
        The derivation of closely related control schemes, such as admittance control, is very similar and could be used to find desired joint positions or velocities for robotic systems lacking torque control.

    
    \subsection{Optimal Control}
    \label{sub:optimal_control}

        A general optimal control problem can be written as
        \begin{equation}\label{eq:OCP}
            \begin{split}
                \{\mathbf{u}^*_k\}_{k=0}^{n-1} = & \argmin_{\{\mathbf{u}_k\}_{k=0}^{n-1}} \sum_{k=0}^{n-1} l \big(\bm{x}_k,\bm{u}_k\big) + l_f \big(\bm{x}_n\big),
                \\
                \text{s.t.}\quad 
                &\bm{x}_{k+1} = f\big(\bm{x}_k,\bm{u}_k\big), 
                \\    
                &\bm{x}_k \in \mathcal{X}, 
                \\
                &\bm{u}_k \in \mathcal{U}, 
            \end{split}
        \end{equation}
        where $\bm{x}_k$ are the states, $\bm{u}_k$ the control commands, $l \big(\bm{x}_k,\bm{u}_k\big)$ the running costs, $l_f \big(\bm{x}_n\big)$ the terminal cost, $f\big(\bm{x}_k,\bm{u}_k\big)$ the system dynamics, and $\mathcal{X}$ and $\mathcal{U}$ are the admissible sets for the state and control, respectively. 

        For simplicity, we assume linear system dynamics 
        \begin{equation}\label{eq:double_integrator_system}
            \bm{x}_{k+1} = f\big(\bm{x}_k,\bm{u}_k\big) = \bm{A}\bm{x}_k + \bm{B}\bm{u}_k,
        \end{equation}
        and choose second-order dynamics in the joint space, such that our state and control become
        \begin{equation}\label{eq:second_order_system_dynamics}
            \bm{x} = \begin{bmatrix}
                \posjoint \\ \veljoint
            \end{bmatrix},
            \hspace{10mm}
            \bm{u} = \accjoint.
        \end{equation}

        Since the main purpose of this section is to demonstrate the usage of the cooperative similarity transforms in optimal control conceptually, we limit the scope to a simple reaching problem. Hence, the running cost is independent of the state $\bm{x}_k$ and only penalizes the magnitude of the control commands $\bm{u}_k$, i.e. 
        \begin{equation}\label{eq:optimal_control_running_cost}
            l(\bm{x}_k, \bm{u}_k) = \transpose{\bm{u}}_k \bm{R} \bm{u}_k, 
        \end{equation}
        where $\bm{R}$ is a regularization term. The terminal cost then adopts Equation~\eqref{eq:optimization_problem_formulation} to minimize the distance to the desired similarity transformation $\versor_{\similarity d}$
        \begin{equation}\label{eq:optimal_control_terminal_cost}
            l_f(\bm{x}_n) = l_f(\posjoint_n,\veljoint_n) = \Big\| \log \big(\reverse{\versor}_{\similarity c}(\posjoint_n) \versor_{\similarity d} \big) \Big\|_{\bm{Q}}^2,
        \end{equation}
        where the $\bm{Q}$ is the precision matrix.        

        While in this formulation the system dynamics are linear, the terminal cost term contains non-linearities. Hence, this problem cannot be solved in closed-form and instead needs to be solved iteratively. Once the optimal trajectory is obtained, the resulting sequence of control commands can be used either open-loop, or in a receding horizon fashion as in model predictive control. Since we opted for second-order dynamics in the joint space, the control commands are accelerations. These can either be integrated to obtain position or velocity commands, or transformed to torque commands via inverse dynamics.

    
    \subsection{Geometric Nullspace}
    \label{sub:geometric_nullspace}

        One of the features of a redundant systems is the existence of a nullspace. This comes from the fact that the linear map, given by the Jacobian of the system, does not have full column rank. The nullspace is the set of vectors that are mapped to zero, i.e. the set of joint velocities that result in no end-effector movement. This is a desirable property for redundancy resolution and hierarchical control with secondary objectives. The classic approach for projecting any velocity vector to the nullspace is using a nullspace projection operator like 

        \begin{equation}\label{eq:nullspace_projector}
            \bm{N}(\posjoint) = \bm{I} - \inverse{\Big(\jacobian_{\similarity}^G(\posjoint)\Big)} \jacobian_{\similarity}^G(\posjoint), 
        \end{equation}
        where in our case $\jacobian_{\similarity}^G(\posjoint)$ is the geometric similarity Jacobian from Equation~\eqref{eq:geometric_similarity_jacobian}. 

        In our work, we use the cooperative similarity transformations to extend the notion of end-effector poses to systems of multiple kinematic chains. The geometric nullspace is induced by the cooperative geometric primitive, since by definition of the geometric primitives, they form a nullspace under the outer product as seen from Equation~\eqref{eq:outer_product_nullspace}. In our formulation of a cooperative geometric primitive, we use the end-effector points of the involved kinematic chains as its defining points. Consequently, those end‑effector points can move arbitrarily along directions tangent to the cooperative primitive without altering the primitive itself. Hence, the cooperative geometric primitive directly encodes the system's geometric nullspace. Due to geometric algebra, the primitives are algebraic objects and not simply parameterizations. But the explicit parameterization, such as radius and center of a sphere, can easily be obtained from these algebraic objects. Furthermore, we can easily relate the derivatives of those parameters to the joint angles by using the analytic Jacobian of the cooperative geometric primitive $\jacobian_c^A(\posjoint)$ that we showed in Equation~\eqref{eq:cooperative_geometric_primitive_jacobian}. This enables using recent redundancy resolution methods such as \cite{haugRedundantSerialManipulator2023} or \cite{ferrentinoDynamicProgrammingFramework2024}.


    }

    
    \subsection{Manipulability Analysis}
    \label{sub:manipulability_analysis}

        For a more fundamental understanding of the proposed cooperative similarity control of these complex robotic systems, we define the manipulability of the system. The cooperative similarity transformation fulfills the same role as the rigid body transformation at the end-effector, represented as a motor, for robotic arms. For traditional manipulability analysis, the geometric Jacobian related to this end-effector motor is used. Accordingly, we define the similarity manipulability ellipsoid as
        \begin{equation}\label{eq:similarity_manipulability}
            \bm{\mathcal{M}}_\similarity(\posjoint) = \jacobian^G_\similarity(\posjoint) \transpose{\big(\jacobian^G_\similarity(\posjoint)\big)}, 
        \end{equation}
        which further highlights the parallels of the system modeling between single kinematic chains using motors and multiple kinematic chains using the cooperative similarity transformation. 

        The similarity manipulability ellipsoid $\bm{\mathcal{M}}_\similarity(\posjoint) \in \mathbb{R}^{7\times 7}$ can be understood as a direct equivalent to the traditional velocity  manipulability ellipsoid. Hence, it informs about the system's capability to move the cooperative geometric primitive, which is the same as the traditional manipulability's information about the robot's ability to move its end-effector. In addition, the cooperative similarity manipulability also contains a dimension that expresses the ability to dilate the cooperative geometric primitive. This dimension becomes of particular interest, when we look at the inverse manipulability $\inverse{\bm{\mathcal{M}}}_\similarity(\posjoint)$. This inverse represents the force manipulability and thus expresses the system's ability to enact forces in different directions. For the dilation, this expresses the ability to apply force that increase or decrease the size the of the cooperative geometric primitive. For example, in the case of three manipulators, this would give a measure of the radial force for changing the radius of the cooperative circle. This is important information for tasks such as carrying big and bulky objects, where the three manipulators are required to apply force to an object while carrying it. The cooperative force manipulability can then be used to optimize for the ideal configuration to enable the maximum force on the dilation. Similarly, this manipulability ellipsoid could then also be used to implement standard techniques for avoiding the singularities discussed in Section~\ref{sub:singularities}. 



\section{Results}
	\label{sec:results}
    We have presented a purely geometric framework for the cooperative control of multiple kinematic chains. Accordingly, the examples are designed to illustrate the underlying mathematical concepts developed in this work through kinematic simulations. We show the cooperative geometric primitives both in optimal control and in teleoperation scenarios. To maintain generality, the examples are intentionally presented in an abstract form, without specific applications. Instead, we only refer to potential applications. The presented results are implemented using our open-source software framework \textit{gafro}\footnote{\url{https://gitlab.com/gafro}}. Additional material and videos of the teleoperation examples can be found on our website\footnote{\url{https://geometric-algebra.tobiloew.ch/cooperative_geometric_primitives/}}.

    

	\subsection{Impedance Control}
	\label{sub:impedance_control_experiments}

\begin{figure}[!ht]
			\centering
			\includegraphics[width=0.6\linewidth]{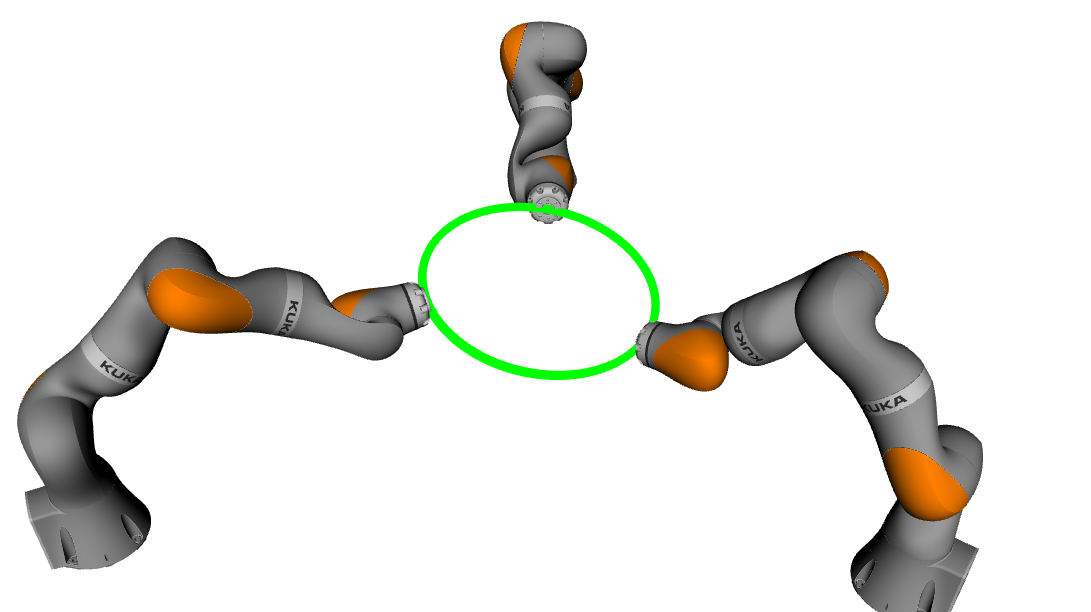}
			\caption{Three Kuka IIWA14 robots arranged in a circle. The green circle shows the cooperative geometric primitive for the current joint configuration.}
			\label{fig:three_franka_robots_}
		\end{figure}

		In this section, we demonstrate the operational space control for the similarity task space that we derived in Section~\ref{sub:impedance_control}. As the setup, we choose three Kuka IIWA14 robots, as shown in Figure~\ref{fig:three_franka_robots_}, giving the full system 21 degrees of freedom. By considering the three end-effector points, the cooperative geometric primitive takes the form of a circle. The control law follows the impedance control in the similarity task space according to Equations~\eqref{eq:desired_similarity_acceleration}, \eqref{eq:desired_similarity_wrench}, and \eqref{eq:operational_space_control}. The final control command will therefore be a torque command. To account for this, we are using a simulation environment that is capable of dynamics simulation using the forward dynamics. For the stiffness and damping parameters we set the gain matrices as $\bm{K} = \text{diag} \big(\begin{bmatrix} 1.0 & 1.0 & 1.0 & 7.5 & 7.5 & 7.5 & 7.5 \end{bmatrix}\big)$ and $\bm{D} = \text{diag}\big(\begin{bmatrix} 5.0 & 5.0 & 5.0 & 5.0 & 5.0 & 5.0 & 5.0 \end{bmatrix}\big)$, respectively.

\begin{figure}[!ht]
			\centering
			\includegraphics[width=\linewidth]{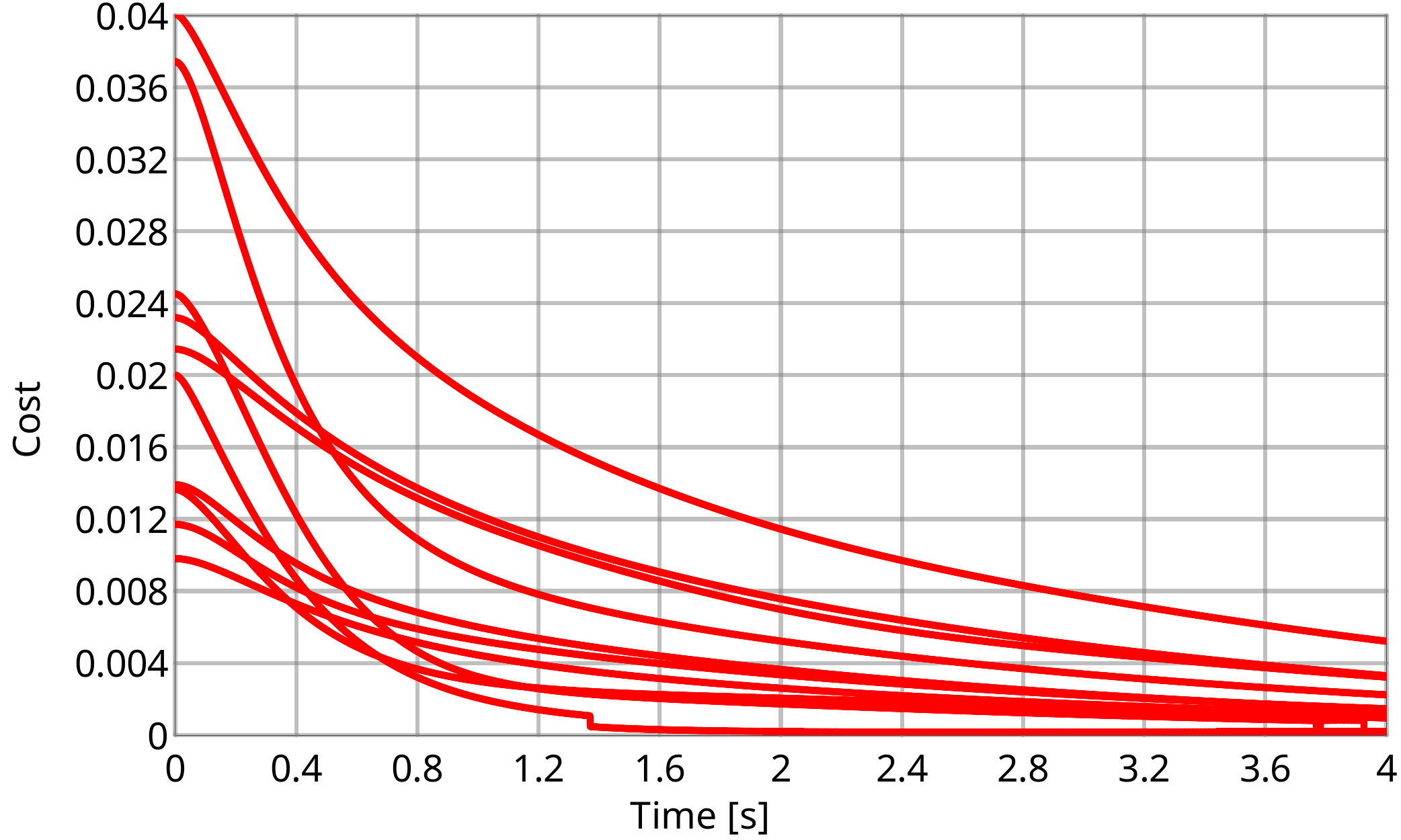}
			\caption{Impedance control response.}
			\label{fig:impedance_control_response_}
		\end{figure}

		For evaluating the similarity task space impedance controller, we uniformly sample ten different random initial joint configurations around a nominal initial joint configuration $\posjoint_0 = \transpose{\begin{bmatrix} \transpose{\posjoint}_{1,0}, \, \transpose{\posjoint}_{2,0}, \, \transpose{\posjoint}_{3,0} \end{bmatrix}}$, with $ \posjoint_{1,0} = \posjoint_{2,0} = \posjoint_{3,0} = \transpose{\begin{bmatrix} 0 & -0.7854 & 0 & 1.3962 & 0 & 0.6109 & 0 \end{bmatrix}}$. We then simulate the system behavior for the given controller and task parameters during 4s. We show the corresponding cost values over time in Figure~\ref{fig:impedance_control_response_} (i.e., the norm of the bivector from Equation~\eqref{eq:similarity_error_dynamics}). It can clearly be seen that the similarity impedance controller minimizes the error and regulates the system towards the target cooperative similarity transformation. Note that the response of the system could be further improved by tuning the stiffness and damping parameters. 

	
	\subsection{Optimal Control Illustrations}
	\label{sub:optimal_control_examples}
		In this section, we show {\color{black} several simulated examples of the optimal control problem formulated in Equation~\eqref{eq:OCP}. As previously described, the problem is formulated as a reaching problem. Accordingly, these results extend reaching a desired pose with a single manipulator to a cooperative system. We use the terminal cost given by Equation~\eqref{eq:optimal_control_terminal_cost} and choose a desired similarity transformation for the cooperative geometric primitive that corresponds to the number of kinematic chains, which can be seen from Table~\ref{tab:cooperative_similarity_task_spaces_}. We choose to solve the optimization problem with the iterative linear quadratic regulator (iLQR) algorithm. We set the number of timesteps to $n=250$, and use $\Delta t = 0.001$. Hence, the planning horizon corresponds to 2.5s. Since the problem setup is very simple, typically the solver converges in five or less iterations. Convergence means that the error bivector from the terminal cost in Equation~\eqref{eq:optimal_control_terminal_cost} essentially becomes zero, which means that the final state has reached the desired similarity transformation.} 

    	
    	\subsubsection{Line Reaching}
    	\label{ssub:line_reaching}
\begin{figure}[!ht]
				\centering
				\begin{subfigure}[t]{\linewidth}
			        \includegraphics[width=0.8\linewidth]{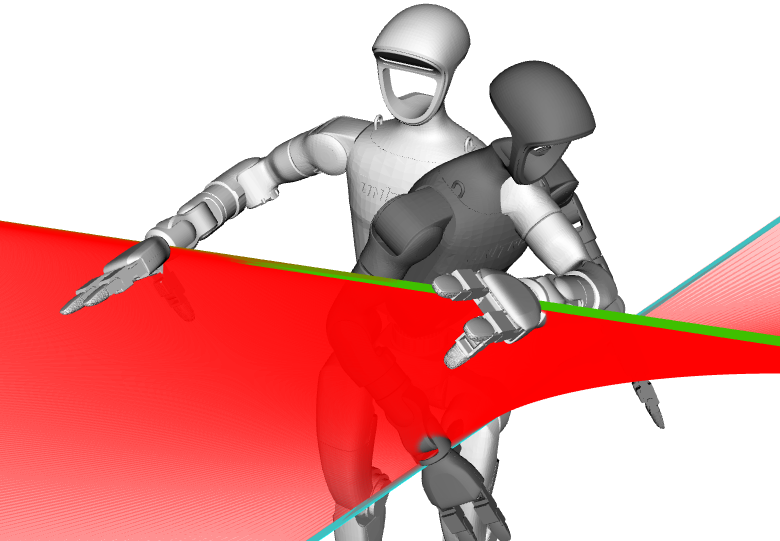}
			        \caption{Task space trajectory of the cooperative line primitive that corresponds to the optimal joint trajectory that was found for reaching the target similarity transformation. The points used for modeling the cooperative line are located at the wrists. The initial configuration is shown in white and the final one in gray. The green line is the initial cooperative line, the turquoise one the final, and in red we show its trajectory in between.}
			        \label{fig:humanoid_reaching_line_}
				\end{subfigure}

				\vspace{5mm}

				\begin{subfigure}[t]{\linewidth}
					\includegraphics[width=\linewidth]{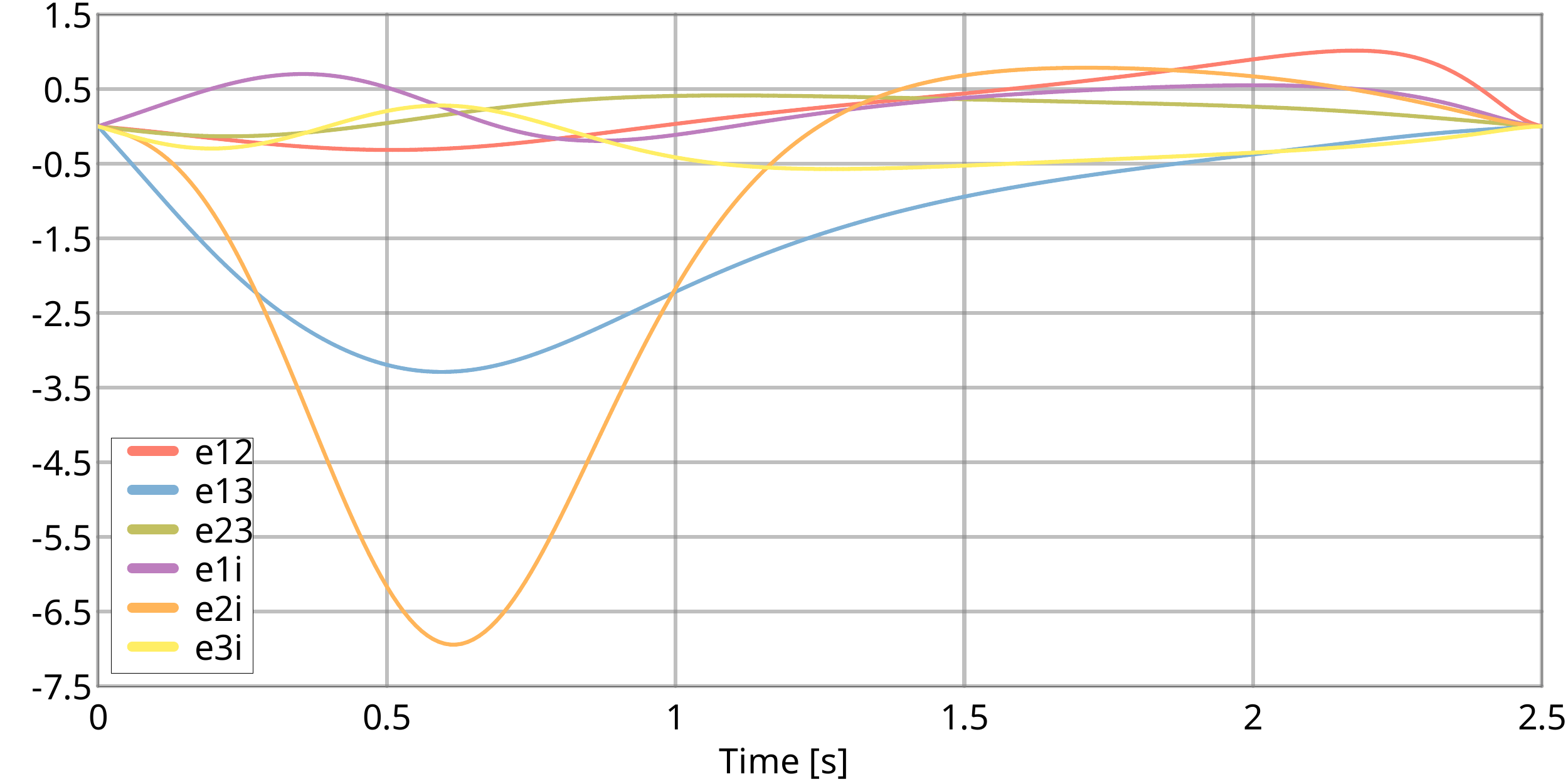}
					\caption{Trajectory of the task space bivector command.}
			        \label{fig:humanoid_reaching_line_bivector}
				\end{subfigure}
				\caption{Humanoid reaching for a desired cooperative line. Including the joints at the waist, this system has 17 degrees of freedom. \ref{fig:humanoid_reaching_line_} shows the task space trajectory and \ref{fig:humanoid_reaching_line_bivector} the corresponding similarity bivector command.}
			\end{figure}

		    In our first example, we examine the simplest case: two points representing the cooperative task space of two independent kinematic chains, such as a bi-manual platform or the arms of a humanoid. In this scenario, we use the Unitree G1 robot model. We neglect the legs in this example and instead focus on the two arms, starting from the waist joints. Since the arms each have seven degrees of freedom and the waist has three, the total system has 17 degrees of freedom. The reference points to construct the line according to Equation~\eqref{eq:cooperative_geometric_primitive} are located at the wrists. By using a line instead of a point pair to model the cooperative behavior of the two arms, we remove constraints on the motion and allow for the arms to move freely along the line, which essentially defines a geometric nullspace. Here, we illustrate a reaching motion from one line primitive to another in Figure~\ref{fig:humanoid_reaching_line_} with a humanoid model. In Figure~\ref{fig:overview}, we have shown the example of cloth folding as a potential application for a dual arm system modeled by a cooperative line. In Figure~\ref{fig:humanoid_reaching_line_bivector} we show the corresponding task space bivector command that was found by solving the optimal control problem.

	    \subsubsection{Circle Reaching}
	    \label{ssub:circle_reaching}
		    In the second example, we consider three points that define the cooperative task space of three independent kinematic chains. In CGA, three non-collinear points uniquely define a circle primitive. Like a line, a circle introduces a null space, allowing the three points to move freely along it. We demonstrate transitions from one circle to another using three different setups: a trio of manipulators, the torso and arms of a humanoid robot, and a collaborative reaching scenario involving both a manipulator and a humanoid.

\begin{figure}[!ht]
				\centering
				\begin{subfigure}[t]{\linewidth}
					\includegraphics[width=0.8\linewidth]{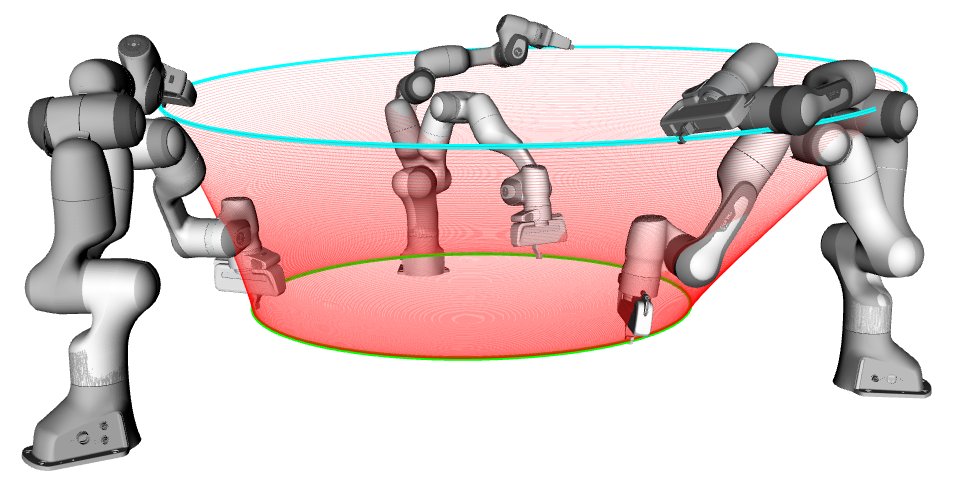}
			        \caption{The initial configuration is shown in white and the final one in gray. The green circle is the initial cooperative circle, the turquoise one the final, and in red we show its trajectory in between.}
					\label{fig:three_franka_robots_reaching_for_a_circle_}
				\end{subfigure}

				\vspace{5mm}
				
				\begin{subfigure}[t]{\linewidth}
					\includegraphics[width=\linewidth]{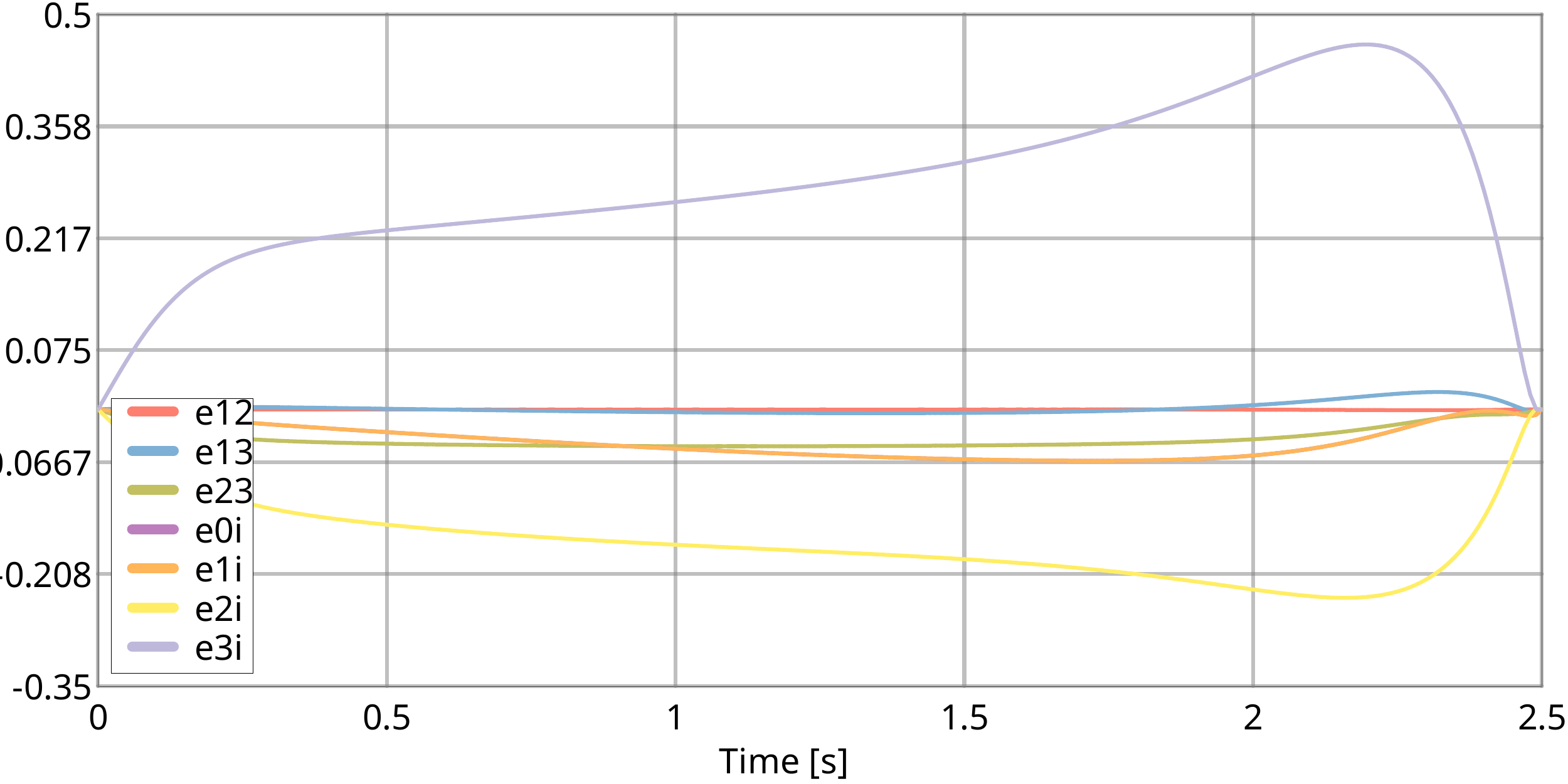}
			        \caption{Trajectory of the task space bivector command.}
					\label{fig:three_franka_robots_reaching_for_a_circle_bivector}
				\end{subfigure}
				\caption{Three Franka robots reaching for a cooperative circle. Since each robot has seven degrees of freedom, the complete system has 21 degrees of freedom.}
			\end{figure}

		    The first scenario of three manipulators, which is shown in Figure~\ref{fig:three_franka_robots_reaching_for_a_circle_}, is the most generic one. It can be seen as a general proxy for tasks involving the lifting and manipulation of big and bulky objects. It is also the same scenario that we used as an example for the impedance controller in Section~\ref{sub:impedance_control_experiments}. We use three Franka robots in this example, which means that the system has 21 degrees of freedom. Notice how the turquoise target circle is larger than the green initial circle. Consequently, the dilation bivector command $\gae{0\infty}$ needs to be non-zero in order for the cooperative system to reach the target. This can be seen in the corresponding Figure~\ref{fig:three_franka_robots_reaching_for_a_circle_bivector}, where we plot the bivector command over the planning horizon.

\begin{figure}[!ht]
				\centering
				\begin{subfigure}[t]{\linewidth}
					\includegraphics[width=0.8\linewidth]{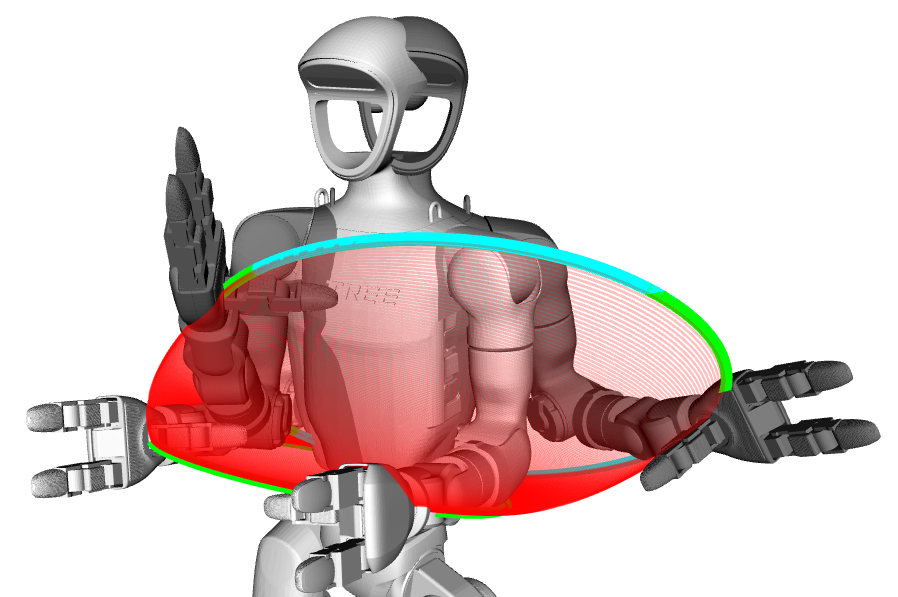}
			        \caption{The initial configuration is shown in white and the final one in gray. The green circle is the initial cooperative circle, the turquoise one the final, and in red we show its trajectory in between.}
					\label{fig:humanoid_torso_reaching_circle}
				\end{subfigure}

				\vspace{5mm}

				\begin{subfigure}[t]{\linewidth}
					\includegraphics[width=\linewidth]{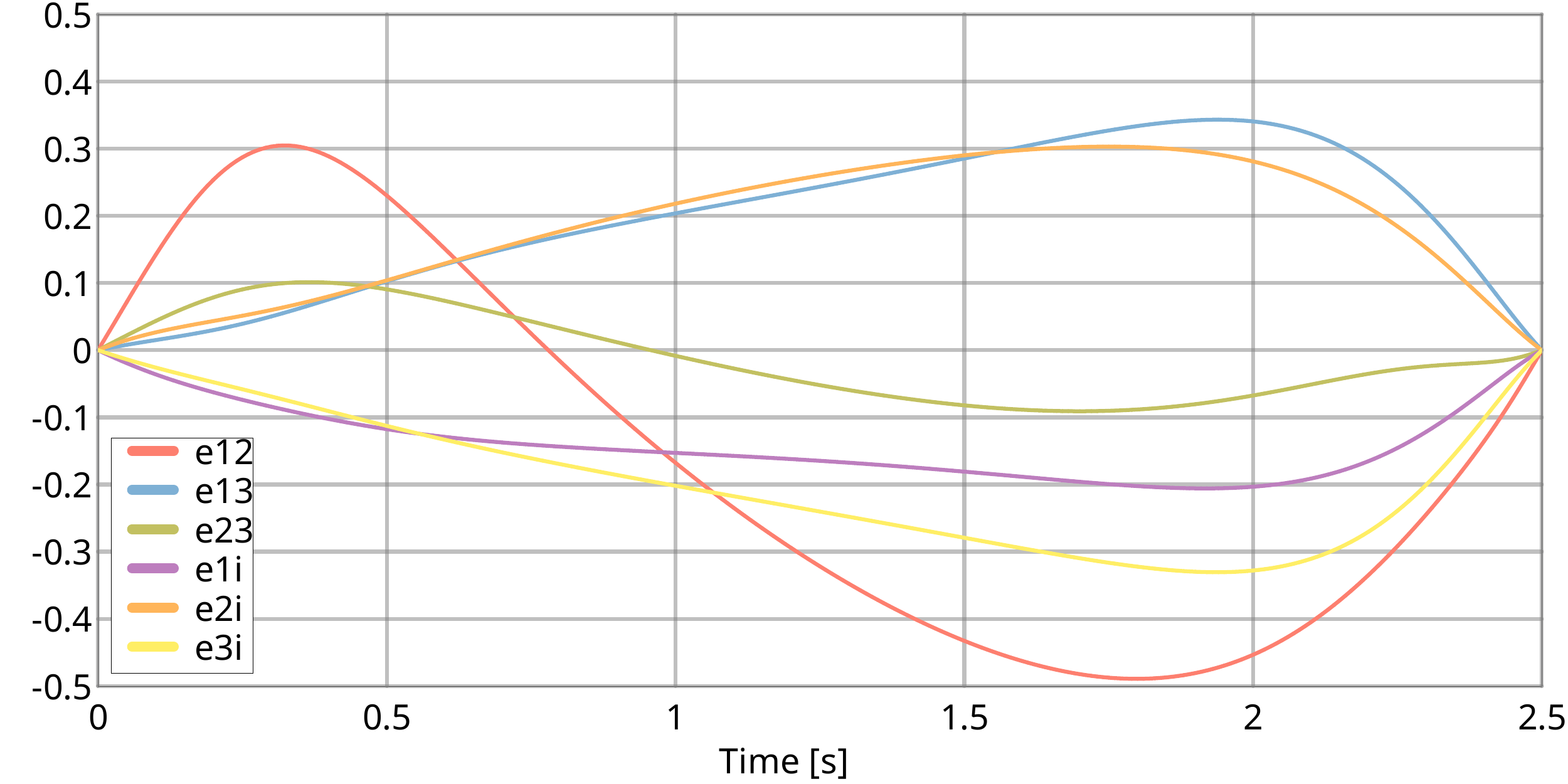}
			        \caption{Trajectory of the task space bivector command.}
					\label{fig:humanoid_torso_reaching_circle_bivector}
				\end{subfigure}
				\caption{Humanoid reaching for a cooperative circle by using a point on its torso.}
			\end{figure}

			We show the second scenario in Figure~\ref{fig:humanoid_torso_reaching_circle} with the bivector command trajectory in Figure~\ref{fig:humanoid_torso_reaching_circle_bivector}. Here, we use a single humanoid and choose a point on the torso and its two arms as the three points that define the circle. Since we use the same robot model, i.e. the Unitree G1, as in Section~\ref{ssub:line_reaching}, the system also has 17 degrees of freedom. This example illustrates how a cooperative circle can be used to model a single humanoid carrying a big and bulky object in arms, while pressing the object against its torso to provide additional support. We show an illustration of this task for a human in Figure~\ref{fig:overview}.

\begin{figure}[!ht]
				\centering
				\begin{subfigure}[t]{\linewidth}
					\includegraphics[width=0.8\linewidth]{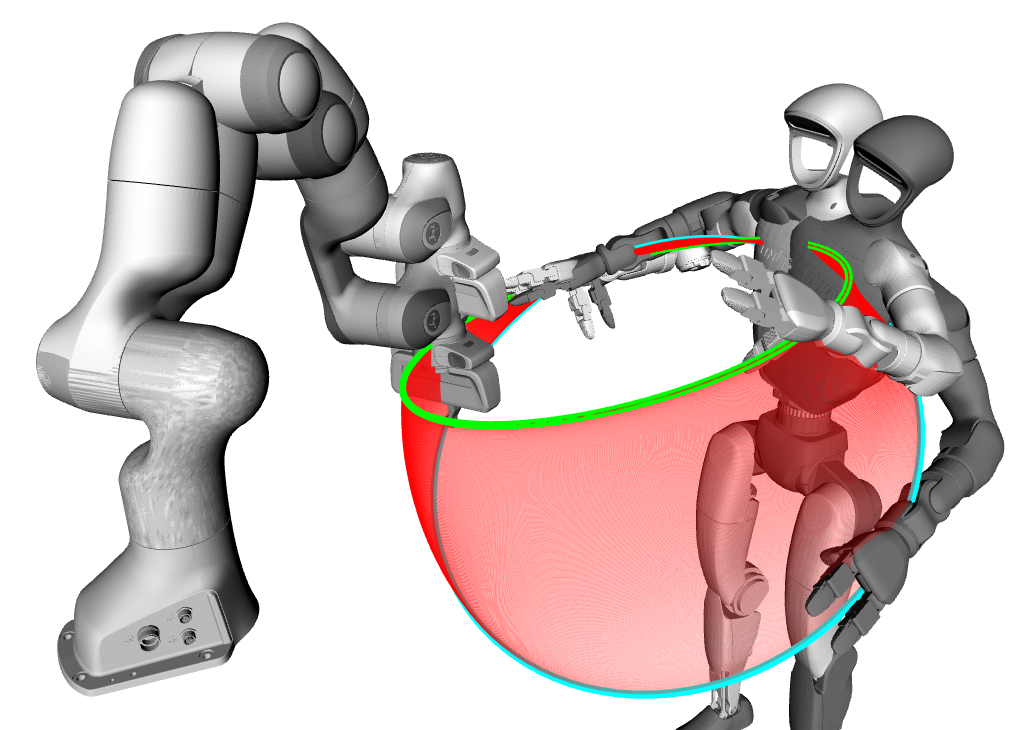}
					\caption{The initial configuration is shown in white and the final one in gray. The green circle is the initial cooperative circle, the turquoise one the final, and in red we show its trajectory in between.}
					\label{fig:collaboration_between_manipulator_and_humanoid_}
				\end{subfigure}
				
				\vspace{5mm}

				\begin{subfigure}[t]{\linewidth}
					\includegraphics[width=\linewidth]{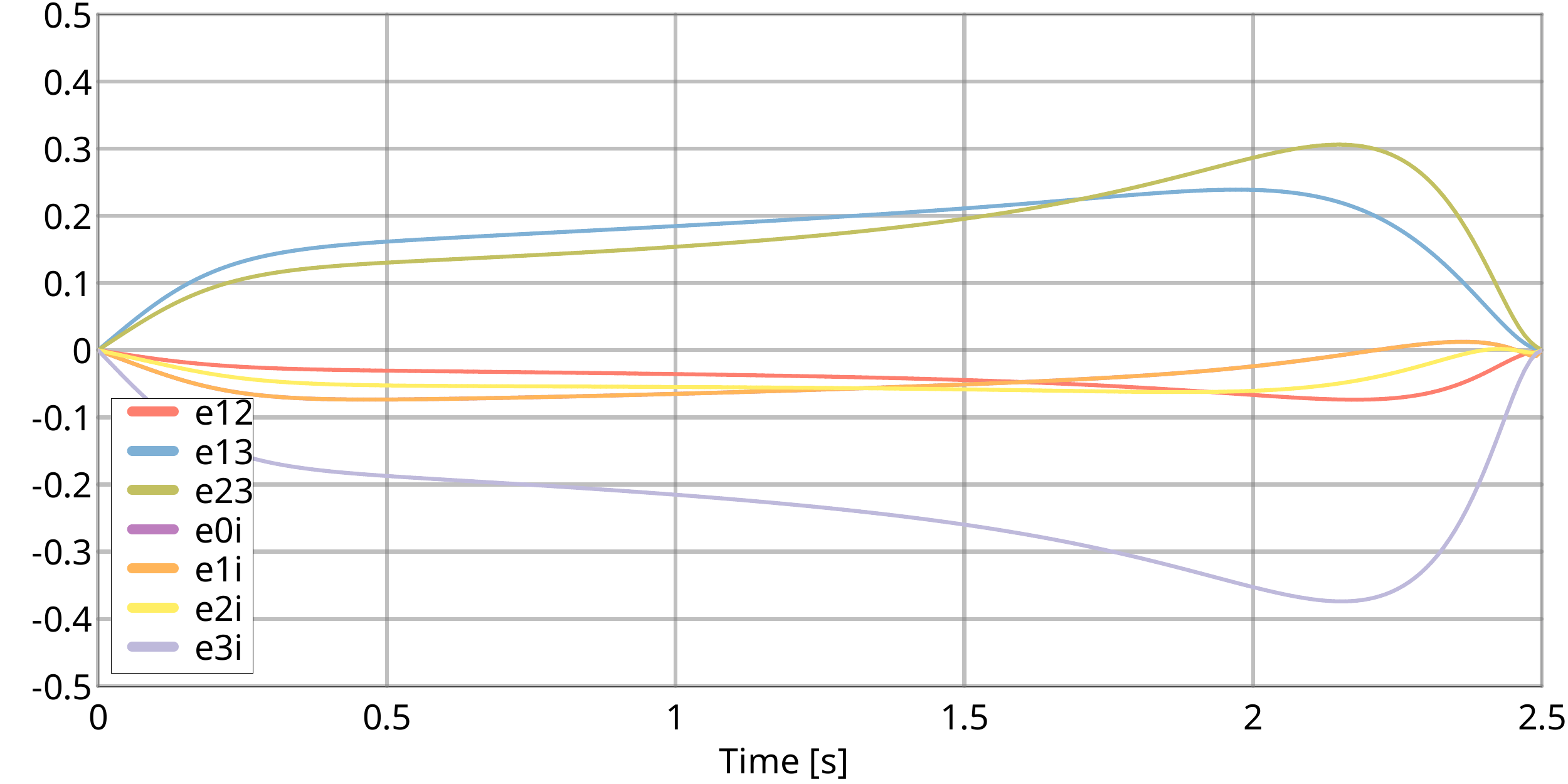}
			        \caption{\color{black} Trajectory of the task space bivector command.}
					\label{fig:collaboration_between_manipulator_and_humanoid_bivector}
				\end{subfigure}
				\caption{Collaboration between a manipulator and a humanoid modeled by a cooperative circle. }
			\end{figure}

			The last scenario for the cooperative circle features a manipulator and a humanoid that are collaborating for the reaching of the desired circle, as shown in Figure~\ref{fig:collaboration_between_manipulator_and_humanoid_}. Although we use a humanoid in this example, it could be easily replaced by a human in order to model a human-robot collaboration scenario. Since this extension provides an interesting opportunity for future work, we discuss it further in Section~\ref{sub:human_robot_collaboration}. {\color{black} The bivector command trajectory can be seen in Figure~\ref{fig:collaboration_between_manipulator_and_humanoid_bivector}.}


	    
	    \subsubsection{Plane Reaching}
	    \label{ssub:plane_reaching}
\begin{figure}[!ht]
		    	\centering
		    	\begin{subfigure}[t]{\linewidth}
			    	\centering
			    	\includegraphics[width=0.8\linewidth]{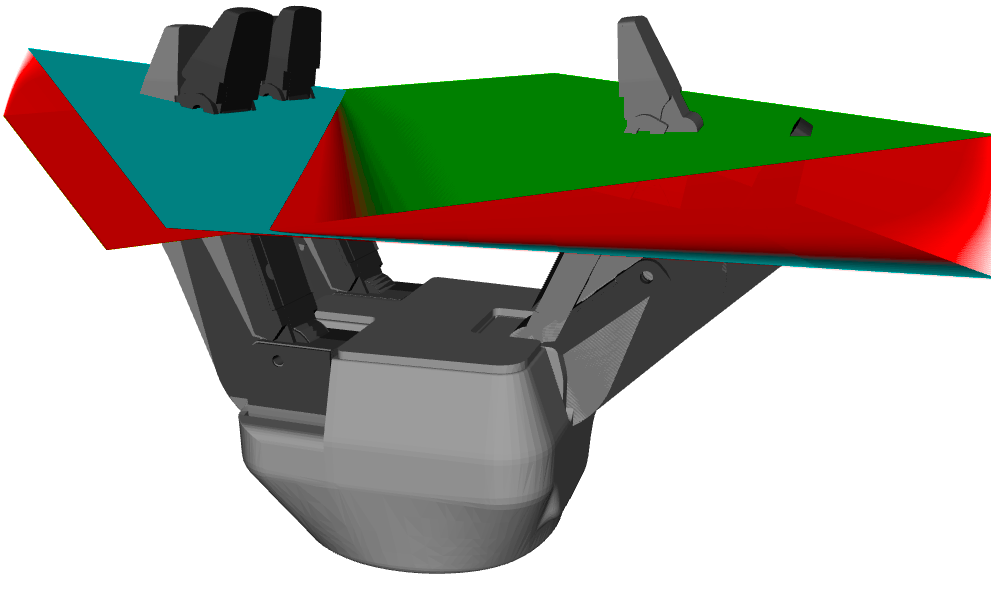}
			    	\caption{The initial configuration is shown in white and the final one in gray. The green plane is the initial cooperative plane, the turquoise one the final, and in red we show its trajectory in between.}
			    	\label{fig:three_finger_hand_plane_reaching_}
		    	\end{subfigure}

				\vspace{5mm}
		    	
		    	\begin{subfigure}[t]{\linewidth}
		    		\includegraphics[width=\linewidth]{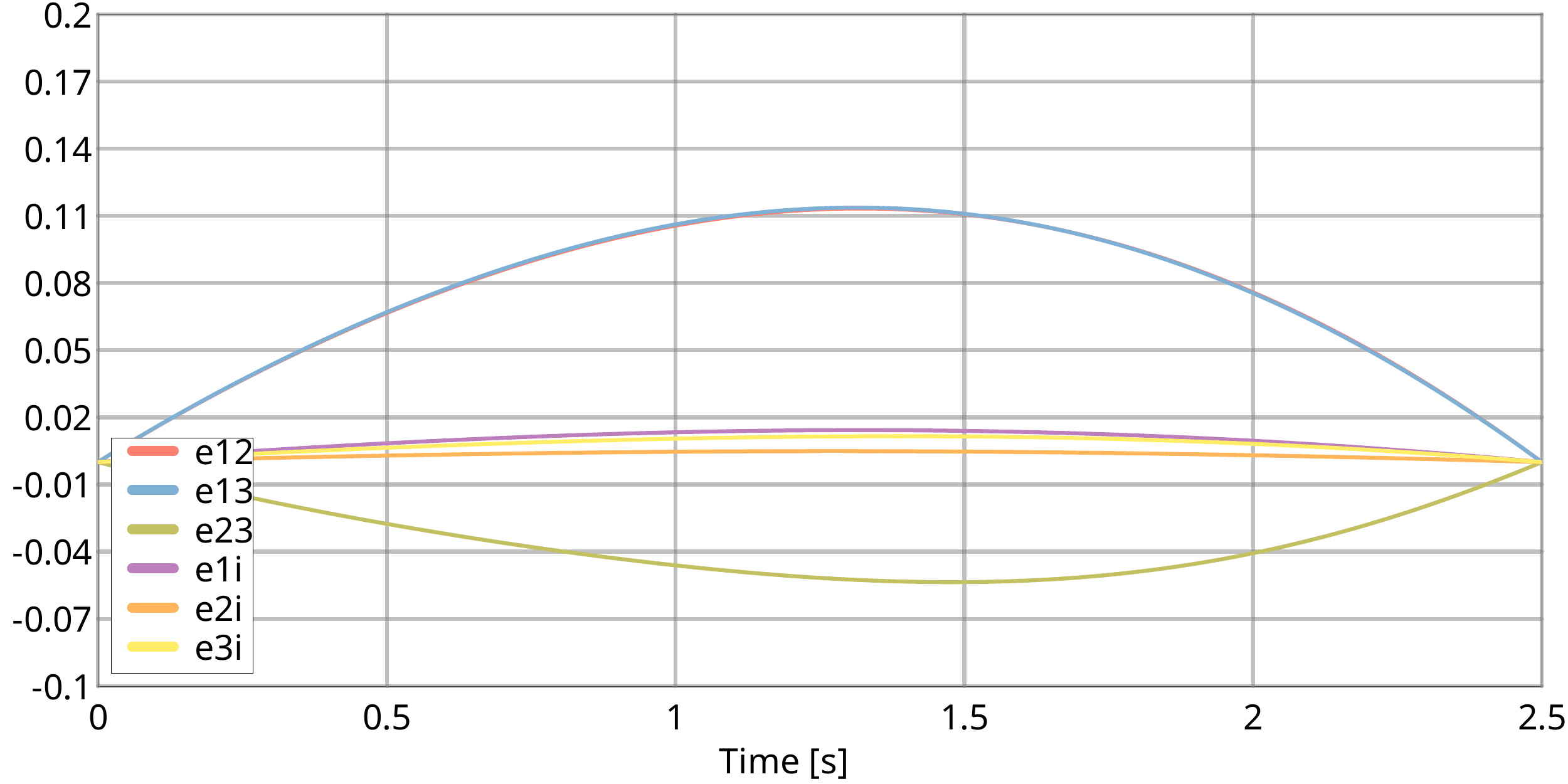}
	    	        \caption{\color{black} Trajectory of the task space bivector command.}
			    	\label{fig:three_finger_hand_plane_reaching_bivector}
		    	\end{subfigure}
		    	\caption{Three-fingered hand reaching for a cooperative plane.}
		    \end{figure}

		    Similar to the line, an infinite plane in CGA can be constructed by combining three points with a point at infinity. As with other primitives in CGA, a plane represents a geometric null space. Therefore, moving the three points within the same plane results in an equivalent control objective. We demonstrate this behavior using a three-fingered robotic hand, as shown in Figure~\ref{fig:three_finger_hand_plane_reaching_}. In Figure~\ref{fig:overview}, we have used the example of a human carrying a plate as a potential application for this scenario. {\color{black} Since a plane is a flat geometric primitive, the dilation component is zero, as can be seen from the bivector command trajectory in Figure~\ref{fig:three_finger_hand_plane_reaching_bivector}.}

	    
	    \subsubsection{Sphere Reaching}
	    \label{ssub:sphere_reaching}
\begin{figure}[!ht]
					\begin{subfigure}{\linewidth}
					\centering
					\includegraphics[width=0.6\linewidth]{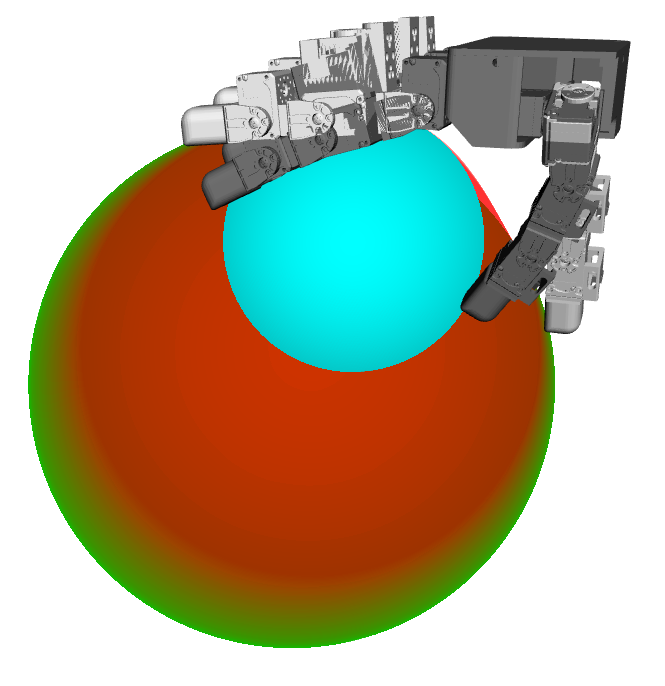}
					\caption{The initial configuration is shown in white and the final one in gray. The green sphere is the initial cooperative sphere, the turquoise one the final, and in red we show its trajectory in between.}
					\label{fig:leap_hand_reaching_for_a_sphere_}
				\end{subfigure}

				\vspace{5mm}

				\begin{subfigure}{\linewidth}
					\includegraphics[width=\linewidth]{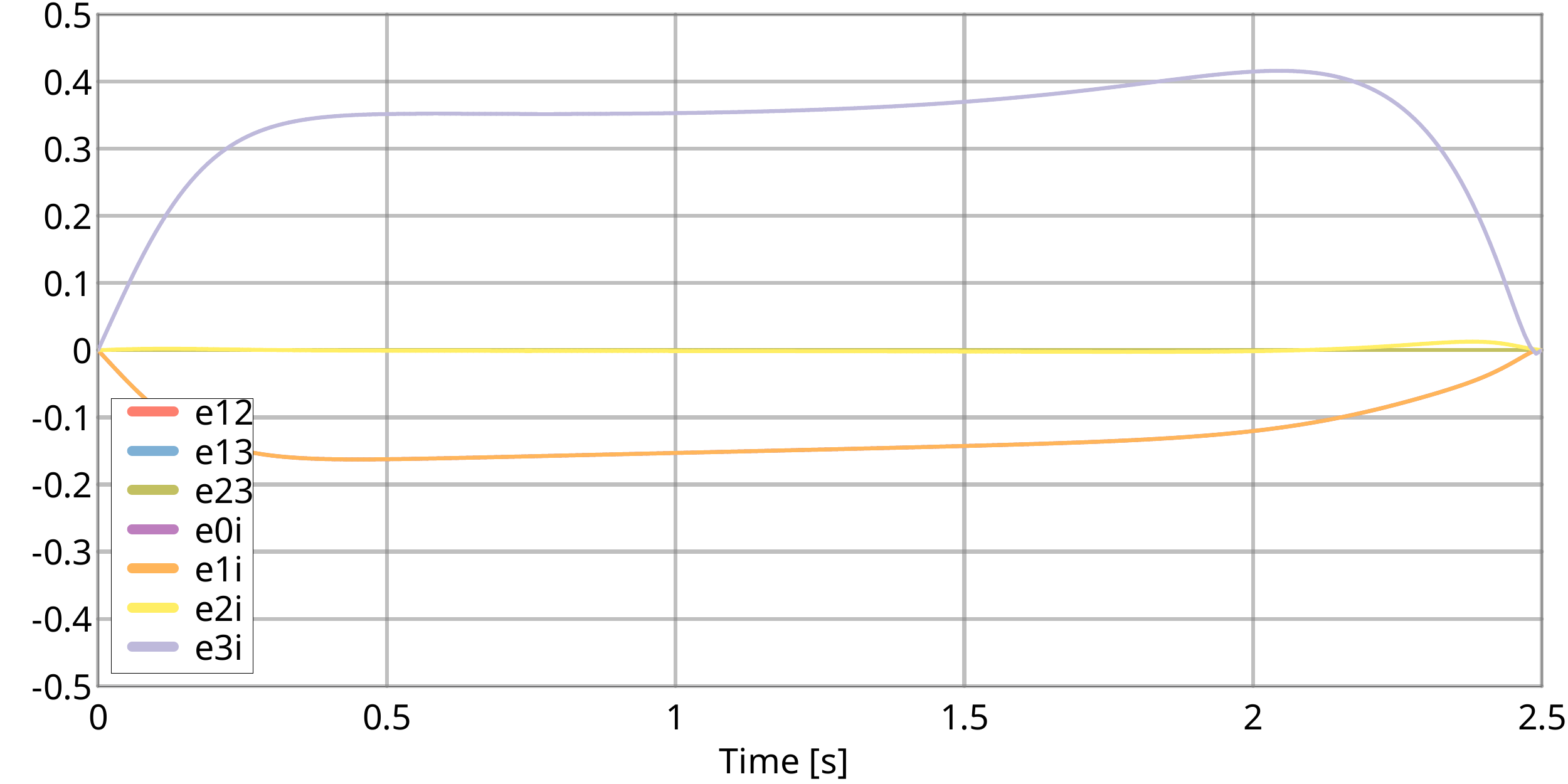}
					\caption{\color{black} Trajectory of the task space bivector command.}
					\label{fig:leap_hand_reaching_for_a_sphere_bivector}
				\end{subfigure}
				\caption{Four-fingered hand reaching for a cooperative sphere. The hand has 16 degrees of freedom. }
			\end{figure}

\begin{figure}[!ht]
				\centering
				\begin{subfigure}{\linewidth}
					\centering
					\includegraphics[width=0.8\linewidth]{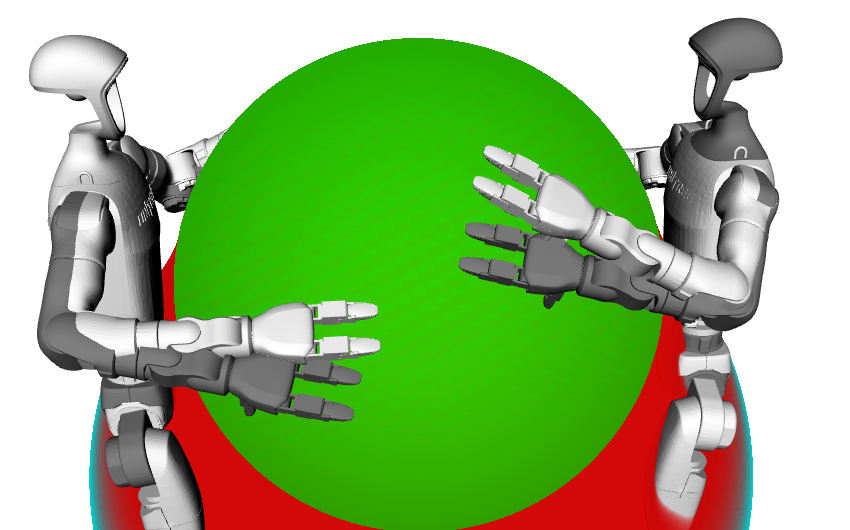}
			        \caption{The initial configuration is shown in white and the final one in gray. The green sphere is the initial cooperative sphere, the turquoise one the final, and in red we show its trajectory in between.}
					\label{fig:two_humanoids_cooperative_sphere_reaching_}
				\end{subfigure}

				\vspace{5mm}
				
				\begin{subfigure}{\linewidth}
			        \includegraphics[width=\linewidth]{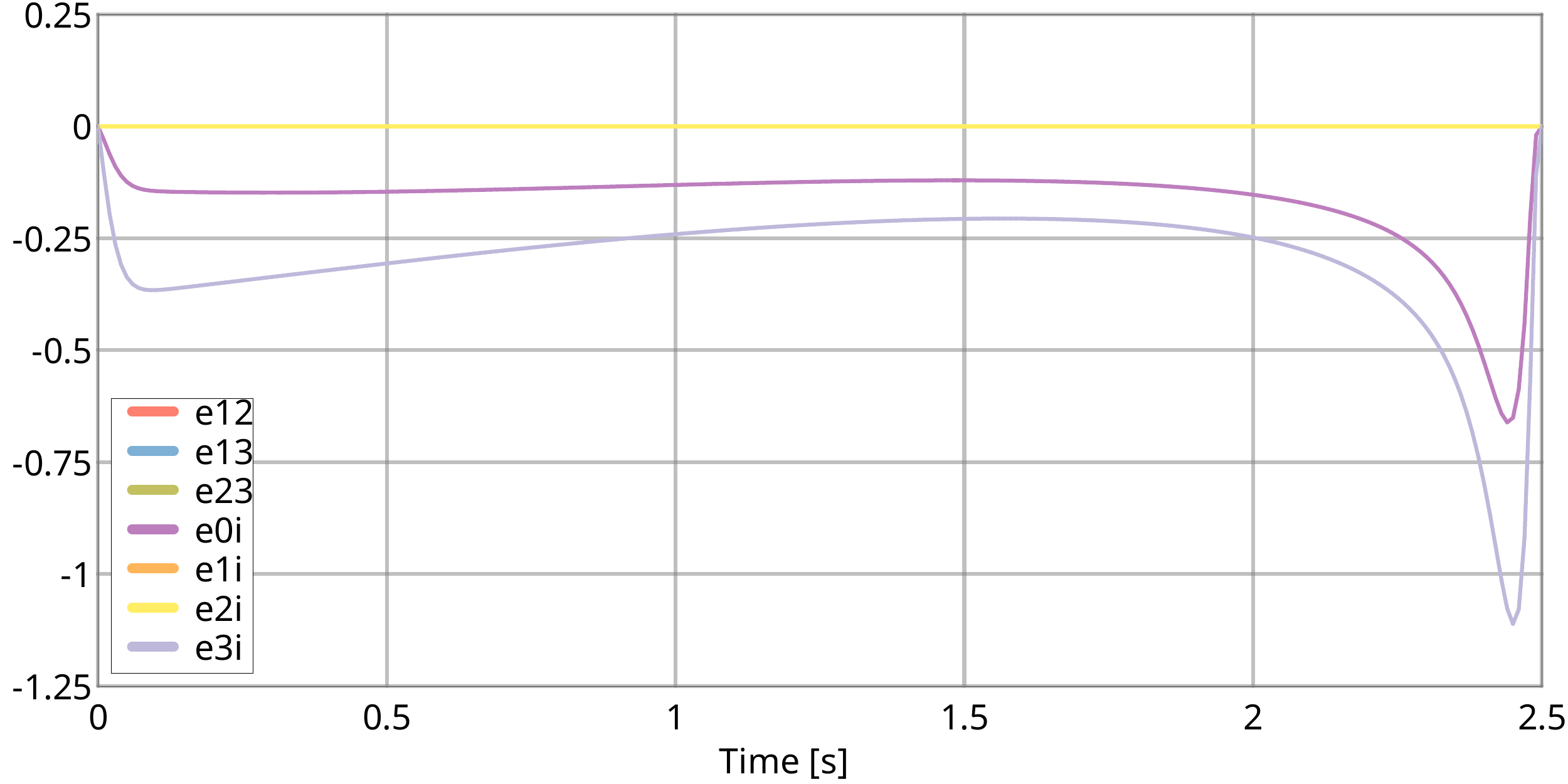}
			        \caption{\color{black} Trajectory of the task space bivector command.}
					\label{fig:two_humanoids_cooperative_sphere_reaching_bivector}
				\end{subfigure}
				\caption{Two humanoids reaching for a cooperative sphere. The combined system has 34 degrees of freedom, including both humanoids' waist joints. }
			\end{figure}

		    As the last example, we considered the sphere primitive constructed using four points, which is the limit that we can reach with CGA.  We demonstrate reaching from one sphere to another using a four-fingered hand in Figure~\ref{fig:leap_hand_reaching_for_a_sphere_}, and a collaborative scenario involving two humanoids in Figure~\ref{fig:two_humanoids_cooperative_sphere_reaching_}. {\color{black} The corresponding bivector command trajectories can be found in Figures~\ref{fig:leap_hand_reaching_for_a_sphere_bivector} and \ref{fig:two_humanoids_cooperative_sphere_reaching_bivector}, respectively.} The cooperative sphere of the four-fingered hand is a general example for grasping using a robotic hand and can thus be applied to a wide range of applications. We also show an example of this in Figure~\ref{fig:overview}. Similarly, in the scenario with the humanoids, one of the humanoids could be replaced by the user in order to model human-robot collaboration. Generally, this example illustrates the collective transport of big and bulky objects using two independent humanoids, as depicted in Figure~\ref{fig:overview}.

	    {\color{black}

	    
	    \subsubsection{Line Constraint}
	    \label{ssub:line_contraint}

	    	The cooperative geometric primitives can not only serve as task objectives but also as constraints for secondary tasks. In the previous examples, we demonstrated reaching problems in which the cooperative primitive transitions from one configuration to another. Here, we illustrate how the cooperative line primitive can be used to impose geometric constraints on the motion of a dual-arm system, while simultaneously executing individual end-effector tasks.

	    	We use the Unitree G1 humanoid robot with its two arms. Together with the waist joints, it provides 17 degrees of freedom. The cooperative line primitive is formed by the two wrist points. In the first scenario, shown in Figure~\ref{fig:cooperative_line_one_hand_fixed_}, we task the right hand to reach a target point, while the left hand remains stationary. This represents an asymmetric reaching task in which only one arm is actively controlled, while the other provides a fixed reference. The cooperative line connecting the two wrists therefore changes freely as the right hand moves towards its target, which can be observed from the trajectory of the line parameters in Figure~\ref{fig:cooperative_line_one_hand_fixed_parameters_}.

	    	In the second scenario, we introduce an additional constraint on the cooperative line, which must now remains horizontal throughout the motion. As shown in Figure~\ref{fig:cooperative_line_horizontal_}, the right hand again reaches a target point, but now the left hand must move in coordination to maintain the horizontal orientation of the line primitive. This coordinated motion ensures that the line parameter corresponding to the vertical component remains constant, which is reflected in Figure~\ref{fig:cooperative_line_horizontal_parameters_}.

	    	This example demonstrates the versatility of the cooperative geometric primitive framework for encoding both task objectives and geometric constraints. By specifying constraints on the cooperative primitive, such as maintaining a horizontal orientation, the framework automatically generates coordinated motion across multiple kinematic chains. This capability is particularly useful for tasks like carrying objects with specific orientation requirements, or maintaining formation constraints during collaborative manipulation.

	    	\begin{figure}[!ht]
	    		\centering
	    		\begin{subfigure}[t]{\linewidth}
	    			\centering
    		        \includegraphics[width=0.8\linewidth]{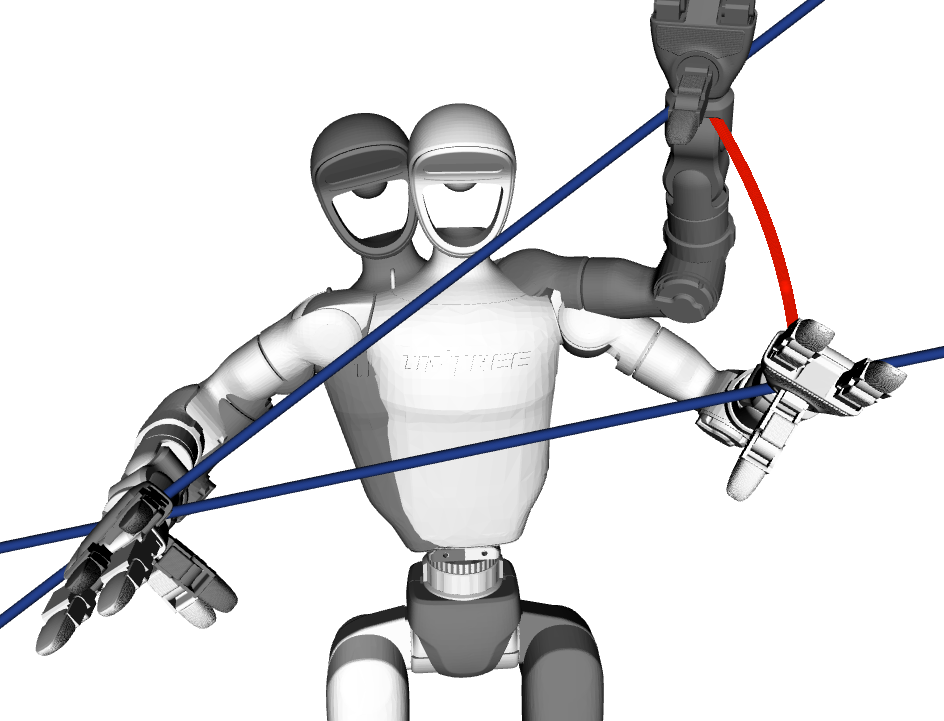}
    		        \caption{\color{black} The initial configuration of the robot is shown in white and the final one in gray. The right hand moves (shown with red trajectory) to reach a target point while the left hand remains stationary. The blue line represents the cooperative line primitive connecting the two wrists.}
    		        \label{fig:cooperative_line_one_hand_fixed_}
	    		\end{subfigure}
	    		\begin{subfigure}[t]{\linewidth}
	    			\centering
    		        \includegraphics[width=\linewidth]{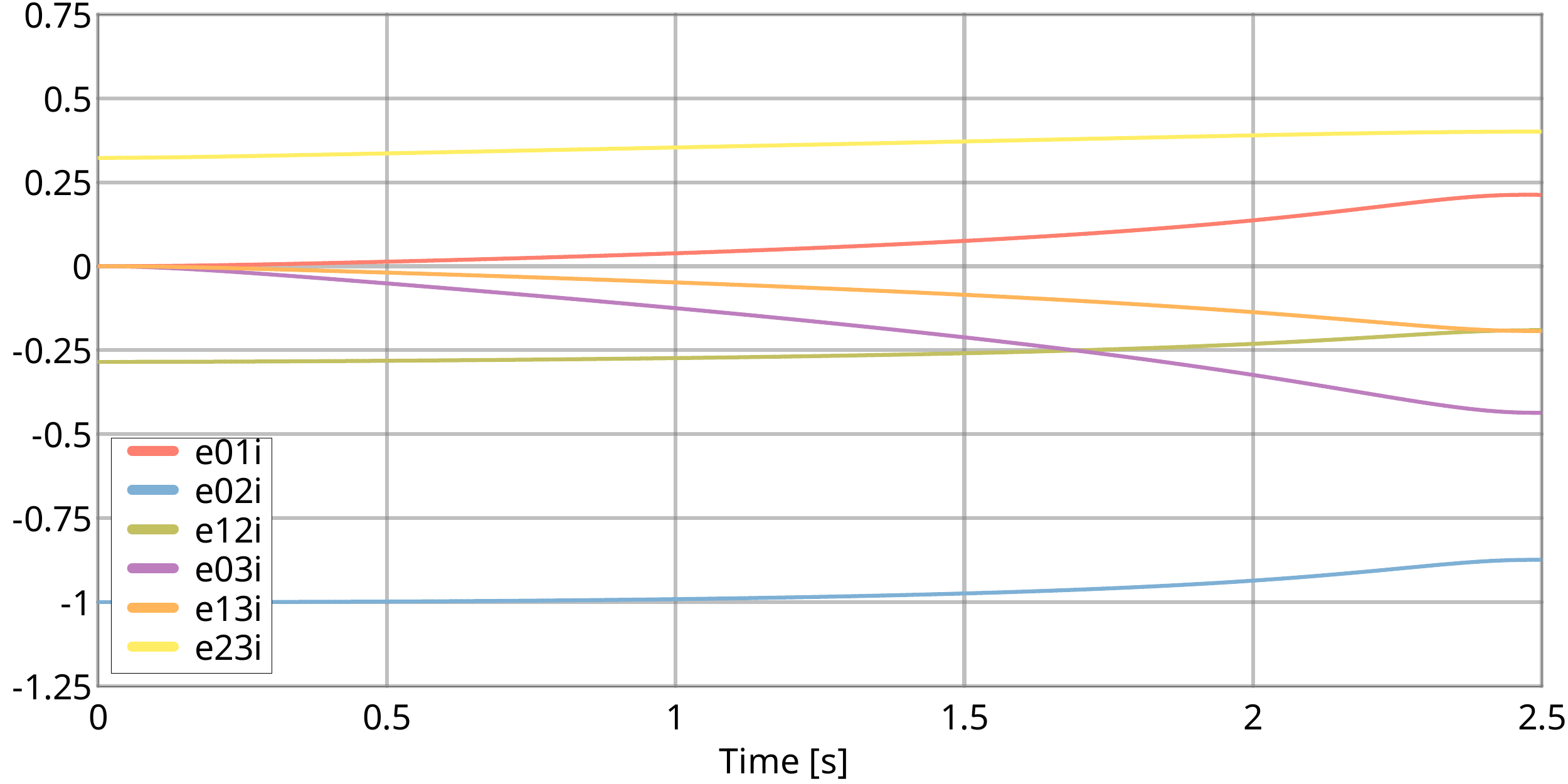}
    		        \caption{\color{black} Trajectory of the line parameters over time. The parameters change as the right hand moves while the left hand remains fixed, showing unconstrained evolution of the cooperative line primitive.}
    		        \label{fig:cooperative_line_one_hand_fixed_parameters_}
	    		\end{subfigure}
	    		\caption{\color{black} Asymmetric reaching task with a cooperative line primitive. The right hand reaches a target point while the left hand stays in place, allowing the cooperative line to change freely.}
	    		\label{fig:cooperative_line_when_reaching_a_point_with_one_hand_while_the_other_hand_stays_in_place_}
	    	\end{figure}

	    	\begin{figure}[!ht]
	    		\centering
	    		\begin{subfigure}[t]{\linewidth}
	    			\centering
    		        \includegraphics[width=0.8\linewidth]{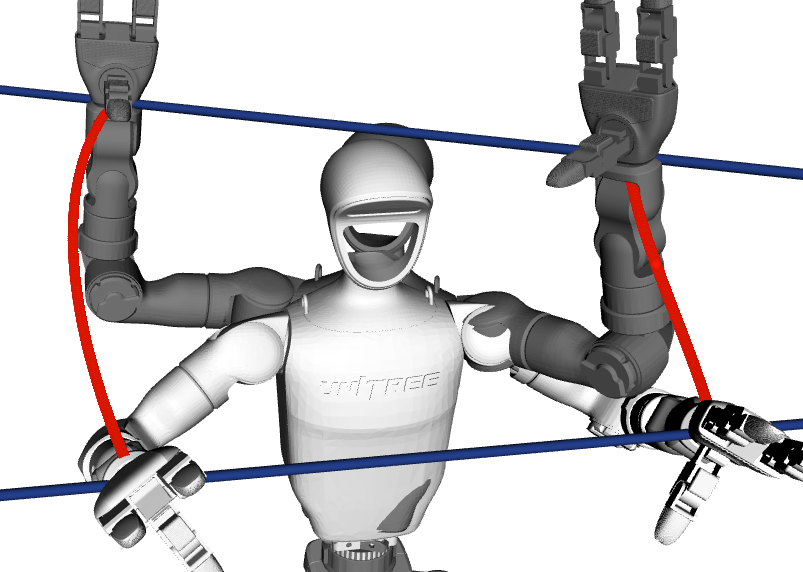}
    		        \caption{\color{black} The initial configuration of the robot is shown in white and the final one in gray. Both hands move (shown with red trajectories) in a coordinated manner to maintain the horizontal orientation of the cooperative line primitive while the right hand reaches the target.}
    		        \label{fig:cooperative_line_horizontal_}
	    		\end{subfigure}
	    		\begin{subfigure}[t]{\linewidth}
	    			\centering
    		        \includegraphics[width=\linewidth]{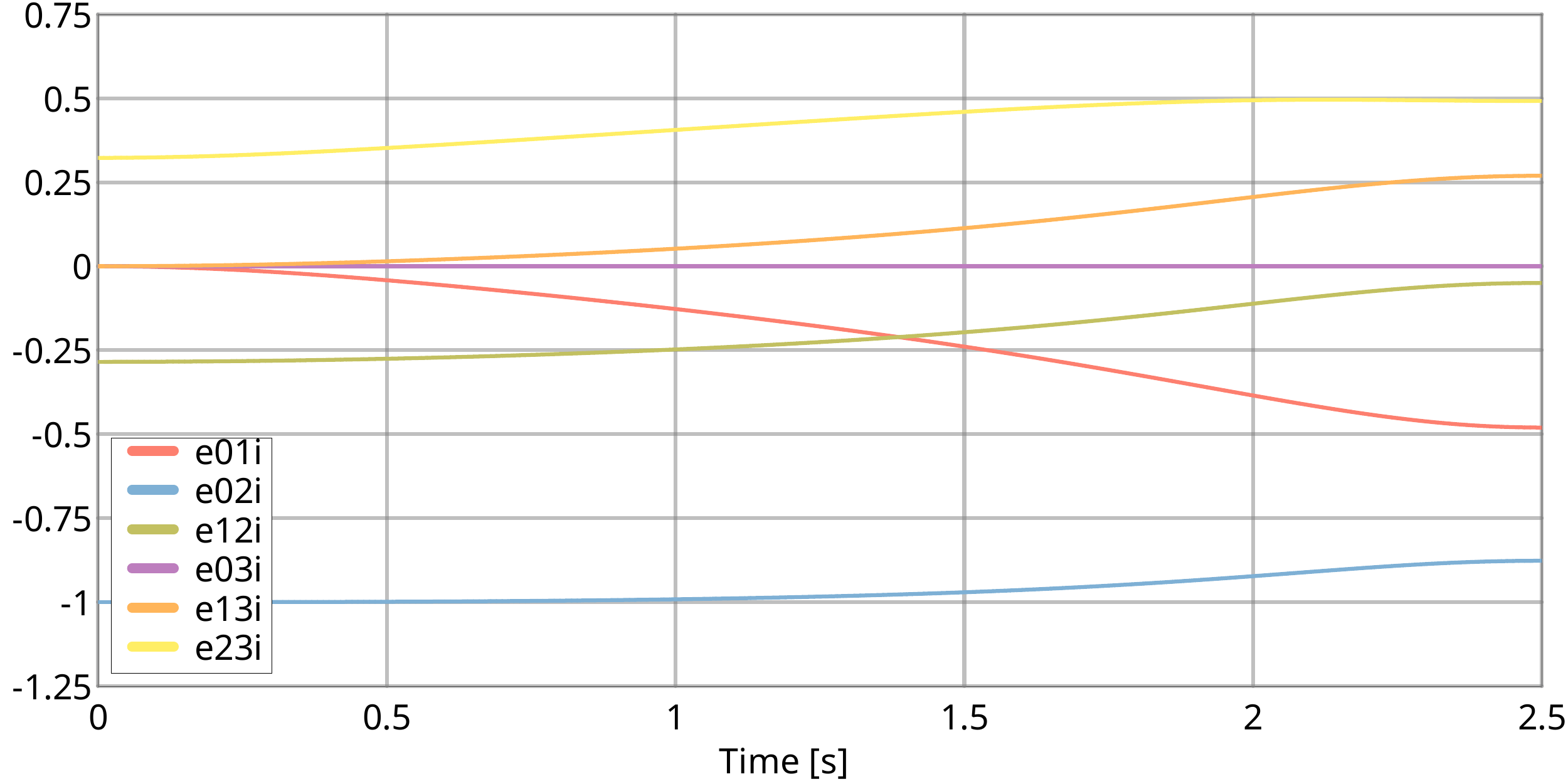}
    		        \caption{\color{black} Trajectory of the line parameters over time. Compared to the unconstrained case, the value of $\gae{03i}$ remains constant, reflecting the horizontal orientation constraint on the cooperative line primitive.}
    		        \label{fig:cooperative_line_horizontal_parameters_}
	    		\end{subfigure}
	    		\caption{\color{black} Constrained reaching task with a cooperative line primitive. The right hand reaches a target point while the cooperative line is constrained to remain horizontal, requiring coordinated motion of both arms.}
	    		\label{fig:cooperative_line_horizontal_constraint_}
	    	\end{figure}


	    }


    
    \subsection{Geometric Nullspace}
    \label{sub:geometric_nullspace_experiments}

\begin{figure}[!ht]
    		\centering
    		\begin{subfigure}[t]{\linewidth}
		        \includegraphics[width=\linewidth]{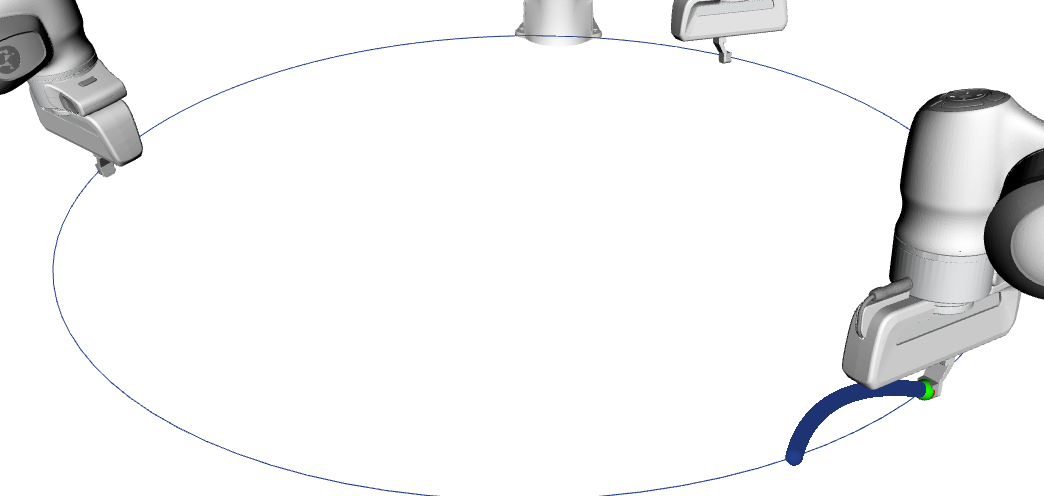}
		        \caption{\color{black} The trajectory of the first robot while being controlled towards the target. Since the geometric nullspace is not considered, the robot deviates from the circle.}
    		\end{subfigure}
    		\begin{subfigure}[t]{\linewidth}
		        \includegraphics[width=\linewidth]{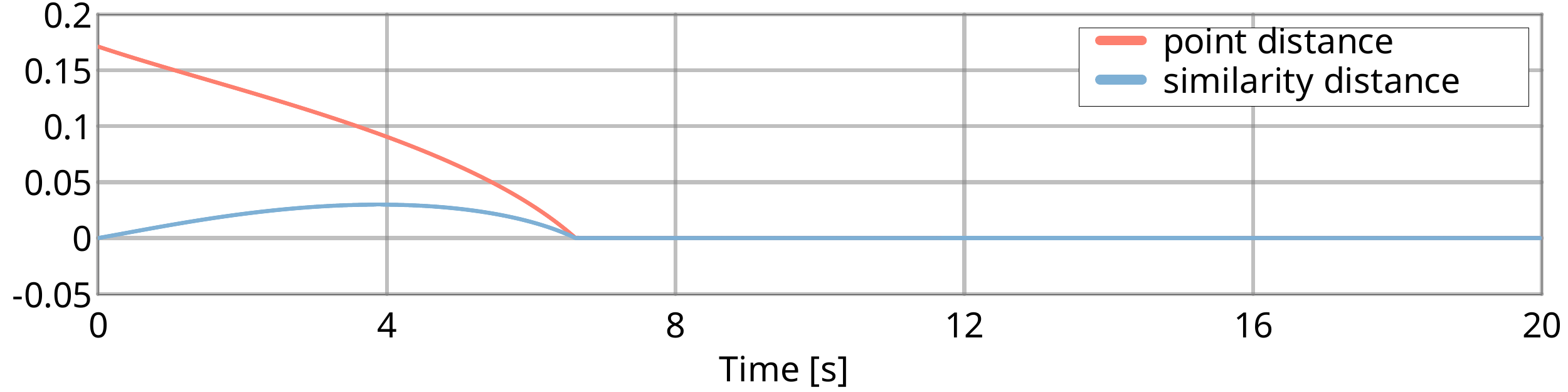}
		        \caption{\color{black} Euclidean distance of the end-effector point to the target point compared to similarity distance between the current cooperative similarity transformation and the initial. Since the geometric nullspace is not considered, the similarity distance is non-zero while the robot is moving towards the target.}
    		\end{subfigure}
    		\caption{\color{black} Cooperative task space of three Franka robots. The first robot is controlled towards the target point (green). The control is not considering the geometric nullspace, i.e. the cooperative circle.}
    		\label{fig:no_nullspace_control_}
    	\end{figure}

    	In this section, we provide a simple example of how the geometric nullspaces can be used in practice to define secondary control objectives. We arrange three Franka robots in circle, yielding a system with 21 degrees of freedom and cooperative geometric primitive that is a circle. The primary objective in this setup is leaving the circle unchanged, i.e. remaining at the same cooperative similarity transformation. The geometric nullspace allows each of robots to move freely on the circle. Hence, we define a secondary objective that tasks one of the manipulators to reach a different point on the circle. Since this secondary task only involves one kinematic chain, we can easily find a desired velocity $\veljoint_{0,d}$ that would move the robot's end-effector towards the goal using classical methods. If we applied $\veljoint_{0,d}$ directly, the end-effector point would not respect the primary objective of staying on the circle. We show this scenario in Figure~\ref{fig:no_nullspace_control_}. It can clearly be seen that the distance between the initial and current similarity transformations is non-zero during the execution of the task.
    	
\begin{figure}[!ht]
    		\centering
    		\begin{subfigure}[t]{\linewidth}
		        \includegraphics[width=\linewidth]{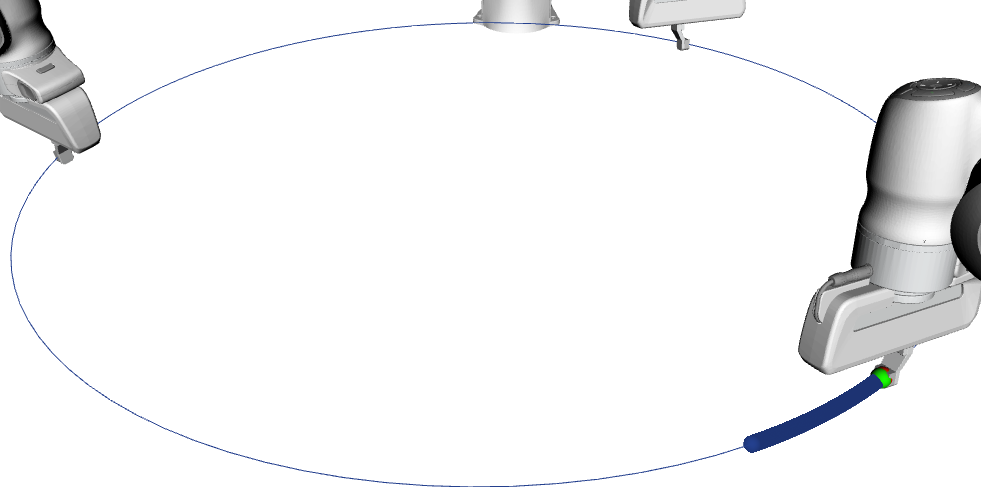}
		        \caption{\color{black} The trajectory of the first robot while being controlled towards the target. Since the task is executed in the geometric nullspace, the robot stays on the circle.}
    		\end{subfigure}
    		\begin{subfigure}[t]{\linewidth}
		        \includegraphics[width=\linewidth]{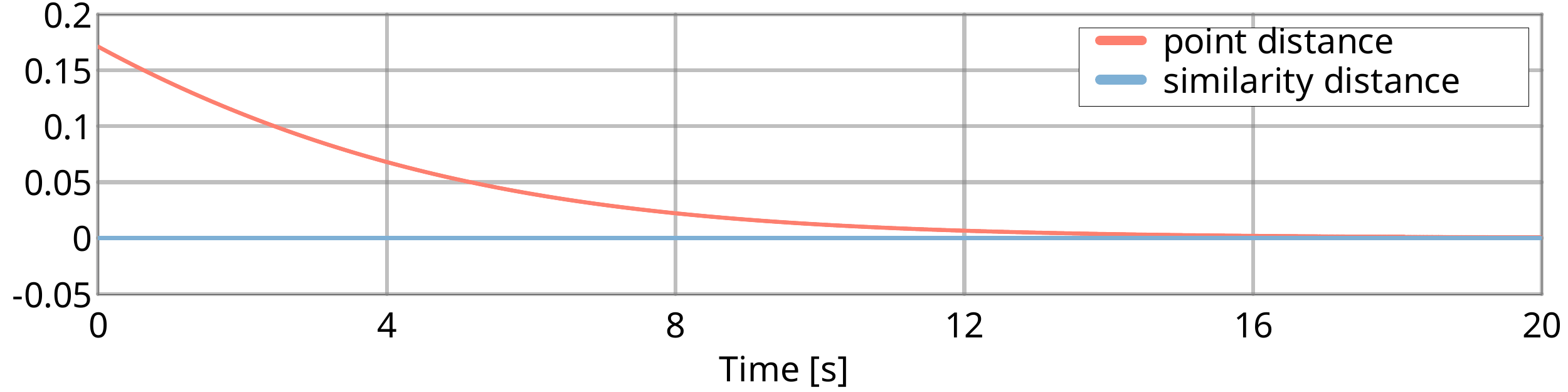}
		        \caption{\color{black} Euclidean distance of the end-effector point to the target point compared to similarity distance between the current cooperative similarity transformation and the initial. Since the geometric nullspace is considered, the similarity distance is zero while the robot is moving towards the target.}
    		\end{subfigure}
    		\caption{\color{black} Cooperative task space of three Franka robots. The first robot is controlled towards the target point (green), while considering the geometric nullspace, i.e. the cooperative circle. The primary objective is keeping the cooperative similarity transformation unchanged, which amounts to staying on the circle.}
    		\label{fig:nullspace_control_}
    	\end{figure}

    	Instead, we project the desired velocity by using the nullspace projector $\bm{N}(\posjoint)$ from Equation~\eqref{eq:nullspace_projector}
    	\begin{equation}\label{eq:projected_velocity}
    		\veljoint_{d,p} = \bm{N}(\posjoint) \begin{bmatrix}
    		    \veljoint_{0,d} \\ \bm{0} \\ \bm{0}
    		\end{bmatrix}.
    	\end{equation}
    	Using $\veljoint_{d,p}$, the first robot still reaches the target point, while staying on the circle. We show this in Figure~\ref{fig:nullspace_control_}. It can be seen that, by executing the motion in the nullspace of the cooperative task space, which corresponds to the cooperative geometric primitive, i.e. the circle, the first robot follows the circle. Therefore, the similarity distance is zero during the task.

    {\color{black}

    
    \subsection{Geometric Singularity}
    \label{sub:geometric_singularity}

    	In addition to the classical kinematic singularities that affect individual manipulators, cooperative systems exhibit geometric singularities that arise from the degeneration of the cooperative geometric primitive itself. These singularities occur when the geometric primitive loses its defining characteristics, such as when three points forming a circle become collinear, or when a circle's radius approaches zero or infinity.

    	To illustrate this phenomenon, we consider the same setup as in Section~\ref{ssub:circle_reaching} (i.e., three Franka robots with a total of 21 degrees of freedom), where the cooperative geometric primitive is a circle defined by the three end-effector points. We task the system with reaching a target configuration that is a geometric singularity, as shown in Figure~\ref{fig:geometric_singularity_configuration_}. In this configuration, the three end-effector points approach a collinear arrangement, causing the circle to degenerate. Note that we are not controlling the cooperative circle here, but instead track its evolution while one end-effector reaches the line formed by the other two.

    	The trajectory of the circle parameters over time is shown in Figure~\ref{fig:geometric_singularity_circle_}. As the system approaches the singular configuration, several of the circle parameters drop rapidly towards zero, indicating the degeneration of the circle primitive. Correspondingly, the bivector trajectory in Figure~\ref{fig:geometric_singularity_bivector_} shows a drastic change in the cooperative similarity transformation. These large components reflect that lines in CGA can essentially be seen as circles with infinite radius  (i.e. the similarity transformation becomes infinite).

    	The effect of the geometric singularity on the system's manipulability is depicted in Figure~\ref{fig:geometric_singularity_manipulability_}. As expected, several components of the task space manipulability decrease significantly when approaching the singularity, with some dropping to zero values. This loss of manipulability indicates that certain directions in the cooperative task space become difficult or impossible to control, which is the property of a singular configuration.

    	This example demonstrates that geometric singularities present an important consideration for the control of cooperative systems modeled by geometric primitives. Unlike classical kinematic singularities that depend solely on individual joint configurations, geometric singularities are intrinsic to the cooperative task space representation and can occur even when all individual manipulators are far from their kinematic singularities. Strategies for singularity avoidance in this context would need to monitor the geometric properties of the cooperative primitive and potentially modify the desired trajectory or task space representation when approaching degenerate configurations.

    	\begin{figure}[!ht]
    		\centering
	        \includegraphics[width=\linewidth]{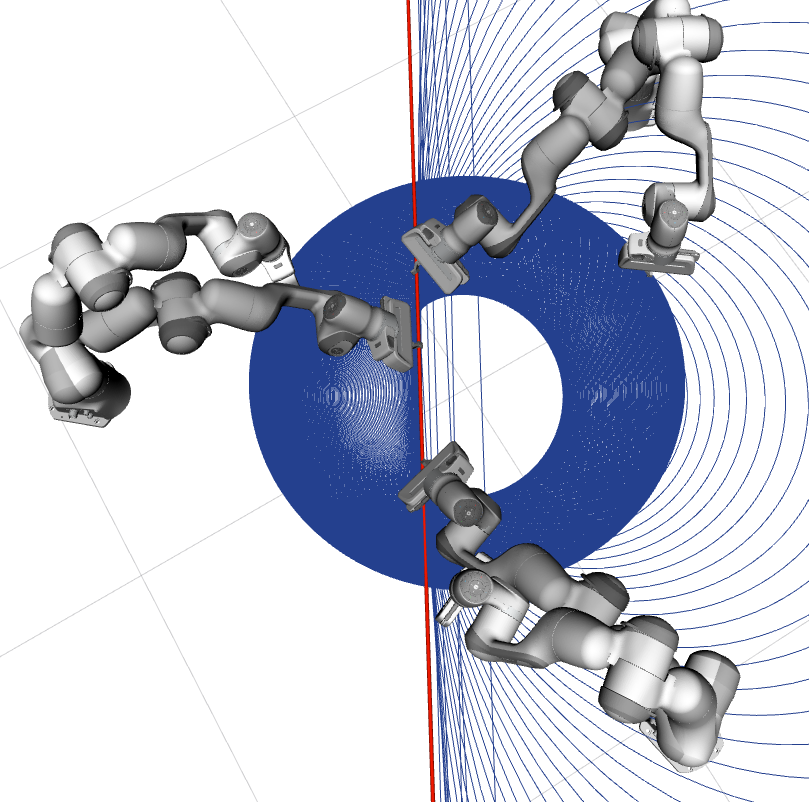}
	        \caption{\color{black} Configuration of the three Franka robots approaching a geometric singularity. The end-effector points approach a collinear arrangement, causing the cooperative circle primitive (shown in blue) to degenerate to a line (shown in red). The initial robot configurations are shown with lighter color.}
	        \label{fig:geometric_singularity_configuration_}
    	\end{figure}
    	\begin{figure}[!ht]
    		\centering
    		\begin{subfigure}[t]{\linewidth}
		        \includegraphics[width=\linewidth]{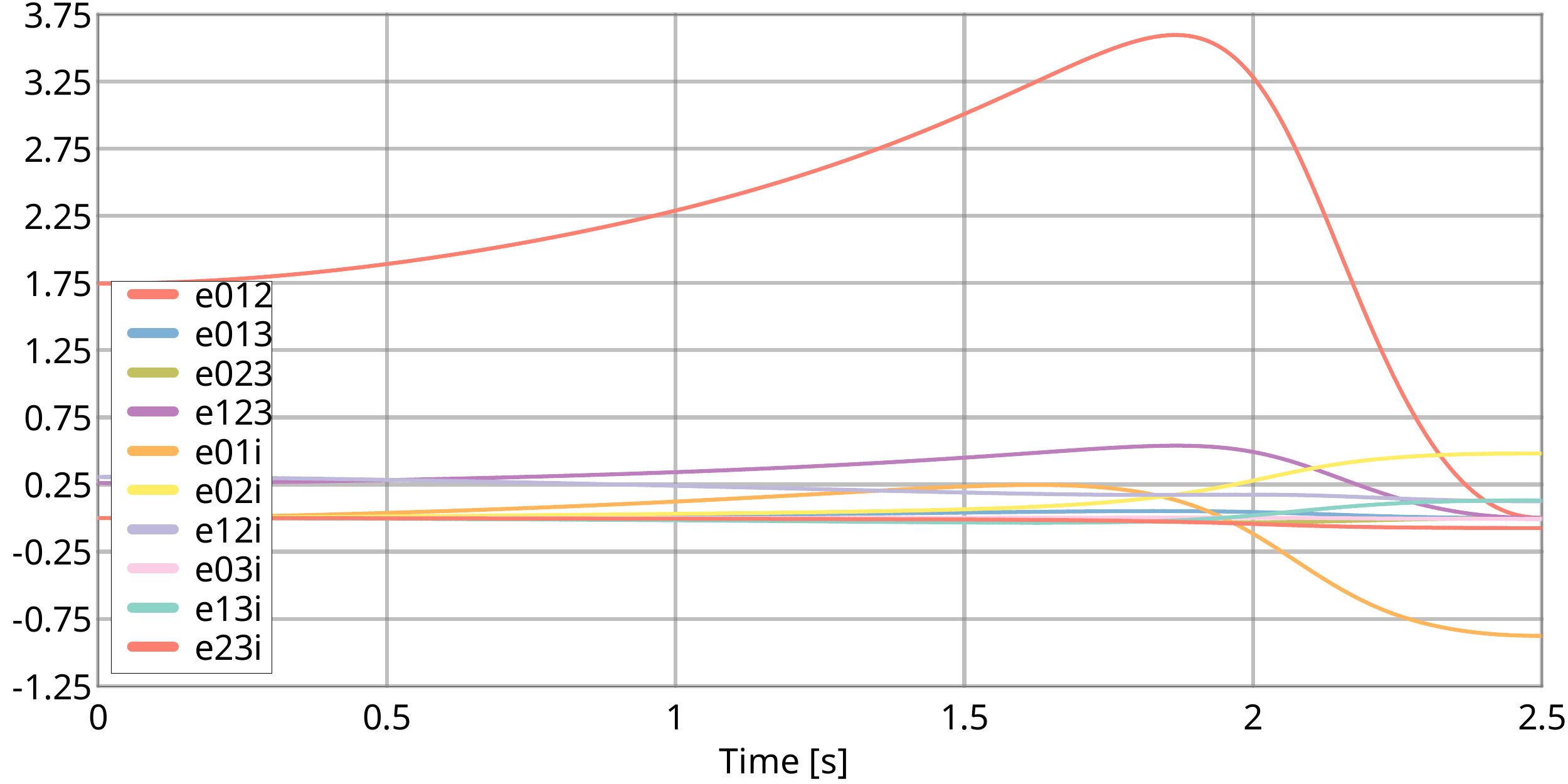}
		        \caption{\color{black} Trajectory of the circle primitive parameters over time. The parameters drop rapidly to zero as the circle degenerates when the three points approach collinearity.}
		        \label{fig:geometric_singularity_circle_}
    		\end{subfigure}
    		\begin{subfigure}[t]{\linewidth}
		        \includegraphics[width=\linewidth]{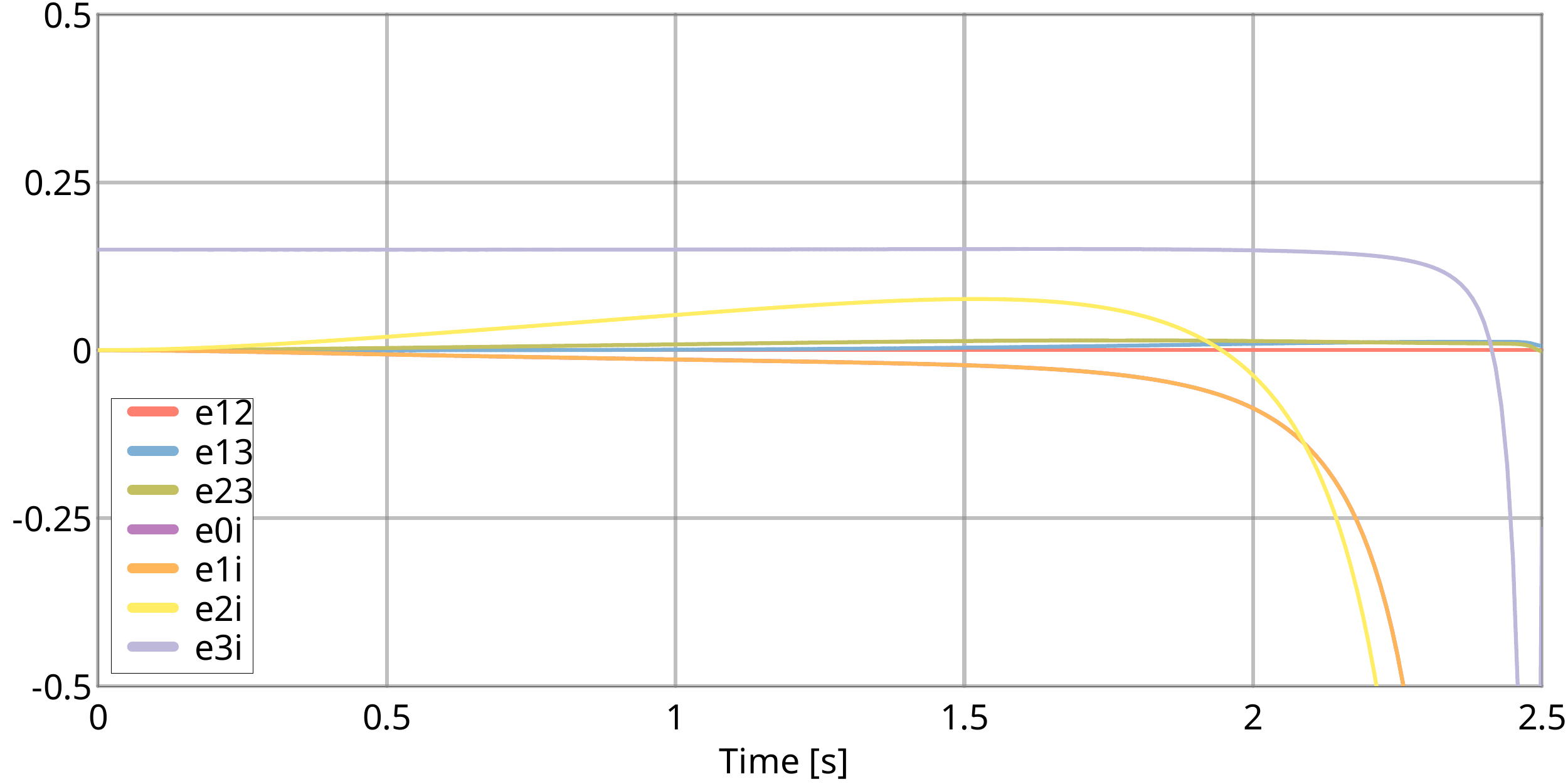}
		        \caption{\color{black} Trajectory of the similarity transformation bivector. Several components diverge, reflecting the large transformation for a line, i.e. a circle with infinite radius.}
		        \label{fig:geometric_singularity_bivector_}
    		\end{subfigure}
    		\begin{subfigure}[t]{\linewidth}
		        \includegraphics[width=\linewidth]{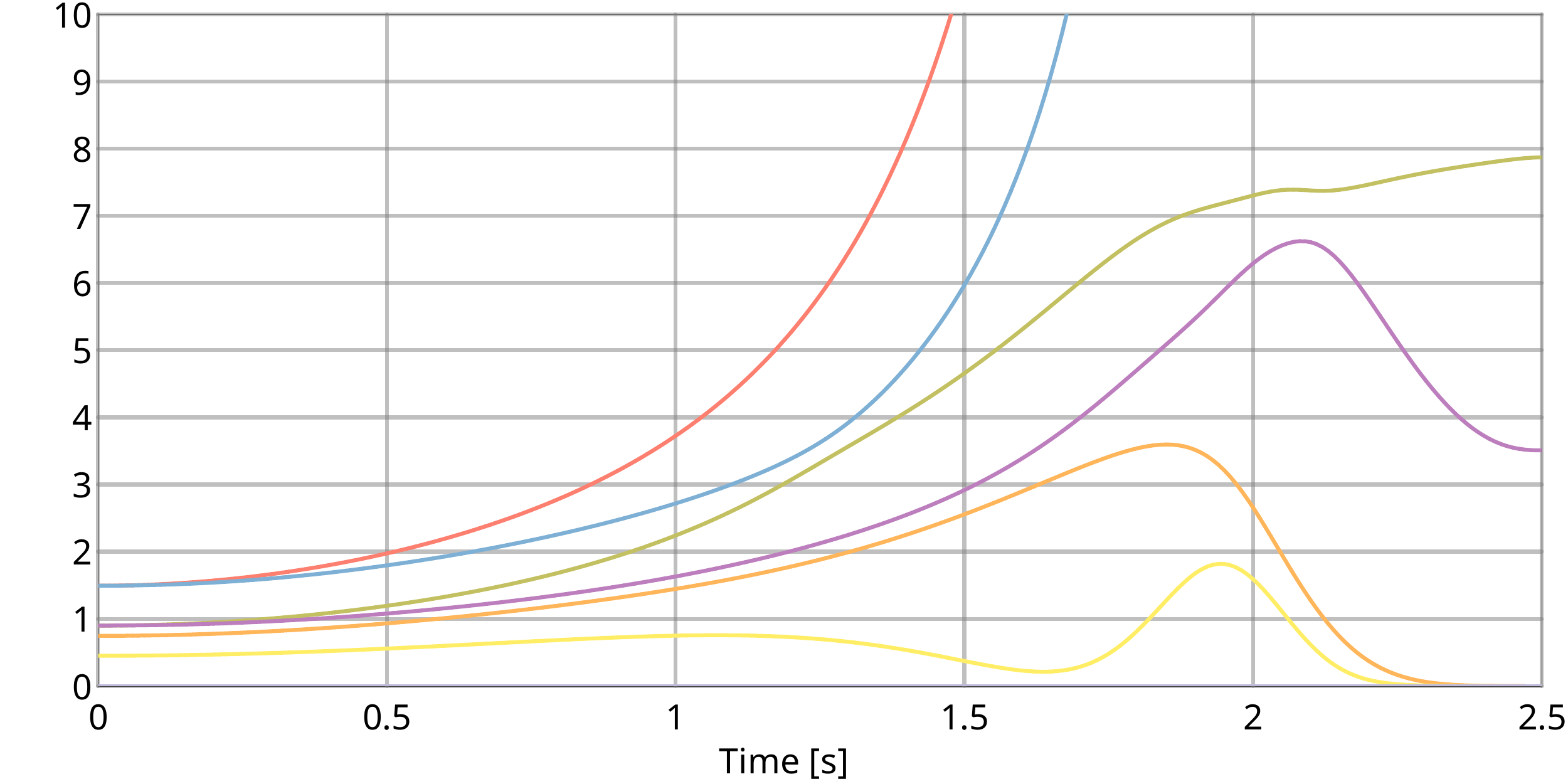}
		        \caption{\color{black} Eigenvalues of the task space manipulability over time. Several components drop to zero values, indicating loss of controllability in certain task space directions due to the geometric singularity.}
		        \label{fig:geometric_singularity_manipulability_}
    		\end{subfigure}
    		\caption{\color{black} Effect of geometric singularity in a cooperative system of three Franka robots. The system passes through a configuration where the circle primitive degenerates as the three end-effector points become nearly collinear.}
    		\label{fig:geometric_singularity_}
    	\end{figure}

    }


	\subsection{Teleoperation Demonstration}
	\label{sub:teleoperation_examples}

		In this section, we present teleoperation for controlling complex robotic systems with a high number of degrees of freedom. The purpose of the example is to demonstrate the simplicity of the cooperative geometric primitives, and how using the cooperative geometric primitives for modeling the cooperative task space of multiple parallel kinematic chains constrains the degrees of freedom and thus greatly facilitates the control of these complex systems. \textcolor{black}{While we simulate the robotic systems, the input device is real hardware.}

\begin{figure}[!ht]
			\centering
			\includegraphics[width=1.0\linewidth]{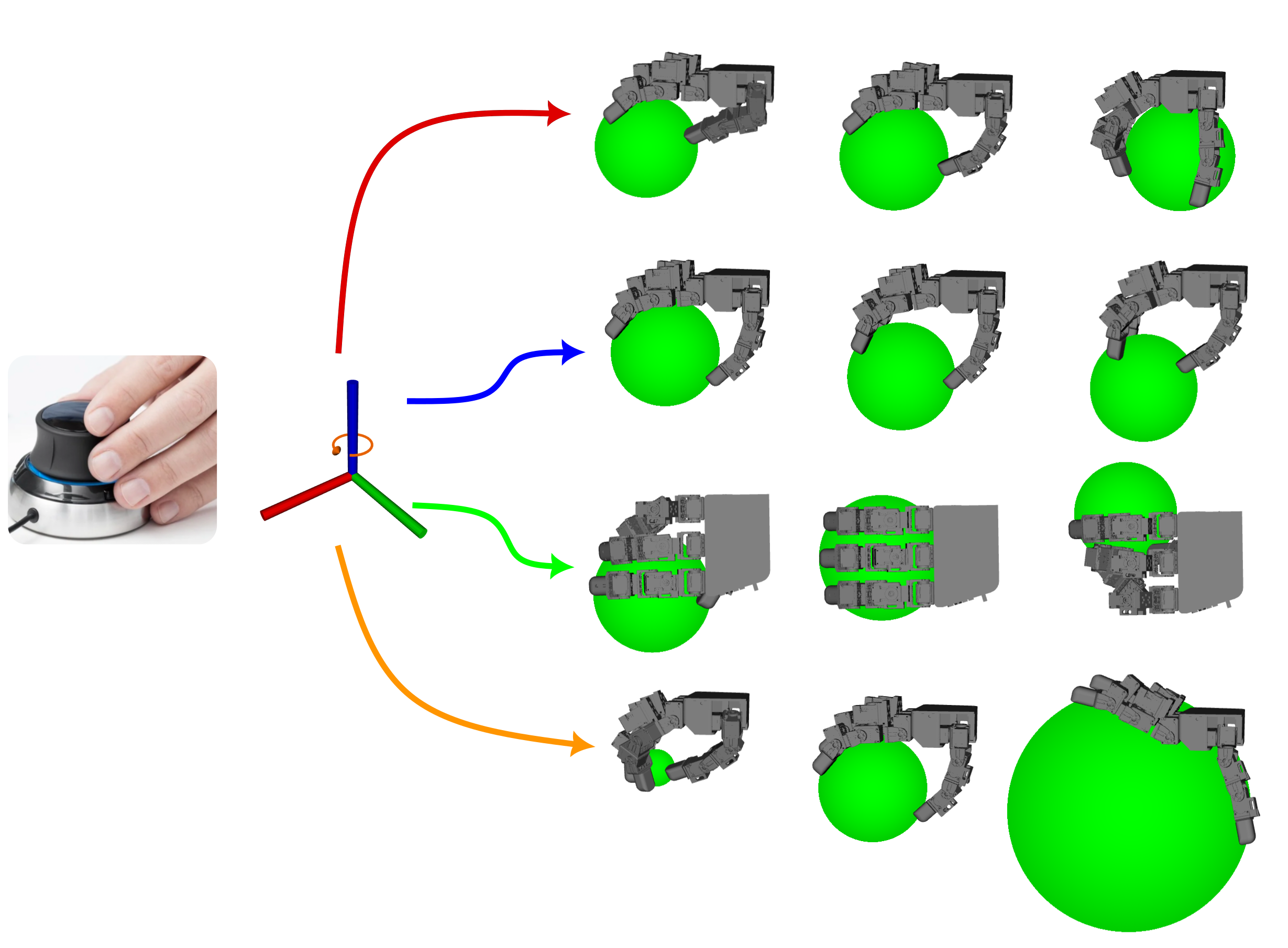}
			\caption{Teleoperation of an anthropomorphic hand. The axes of teleoperation device on the left, are mapped to the different similarity transformations as shown in the center. The three principal axes map to the corresponding translations, while the rotation around the $z$-axis is mapped to the dilation. Rotations are not required, since a sphere is invariant to rotations around its center. The resulting motion of the robotic hand is then depicted on the right.}
			\label{fig:leap_hand_teleoperation_}
		\end{figure}

		We use a readily available six degree of freedom device that is commonly used in CAD applications as a single input device to obtain the commands for the teleoperation. We then map the six axes of the input device to a Lie algebra element of the similarity transformation group, i.e. a bivector, that then represents a task space command. The mapping of the input device's axes to the similarity transformations and the corresponding motion of the robotic hand is shown in Figure~\ref{fig:leap_hand_teleoperation_}.
		{\color{black} We then use the differential kinematics from Equation~\eqref{eq:differential_similarity_kinematics} to obtain a joint velocity command $\veljoint$. This command is used in a first-order kinematics simulation, i.e. we obtain the new joint position as
		\begin{equation}\label{eq:joint_position_kinematics_update}
			\posjoint_{t+1} = \posjoint_t + \Delta t \veljoint, 
		\end{equation}
		where the time step $\Delta t$ is chosen to be 0.01s.
		}

		We want to point out that the sphere in Figure~\ref{fig:leap_hand_teleoperation_} is only visualizing the current cooperative sphere. It is, however, not used for control, in the sense that the sphere is moved and then tracked by the fingers. Instead, the controller from Equation~\eqref{eq:differential_similarity_kinematics} causes a joint movement that then results in the sphere changing accordingly. {\color{black} Since the cooperative similarity transformation and the related geometric similarity Jacobian capture all involved kinematic chains directly, all fingers are controlled simultaneously. We want to further demonstrate this important point by using the pure dilation bivector command as an example. A pure dilation bivector means that the bivector command is $\bivector_\similarity = \log (d) \gae{0\infty}$, where $d \in \mathbb{R}^+$ is the the scaling factor. More information on the uniform scaling group can be found in Appendix~\ref{sub:uniform_scaling_group}. It can be inferred that $d=1$ leaves the hand position unchanged, $d\in (0,1)$ will close the hand, and $d>1$ will open it. We can therefore entirely control the opening and closing of the hand using a single scalar. We show the effects of this pure dilation command in Figure~\ref{fig:example_of_the_effect_of_a_dilation_command_}.
		\begin{figure}[!ht]
			\centering
			\includegraphics[width=0.325\linewidth]{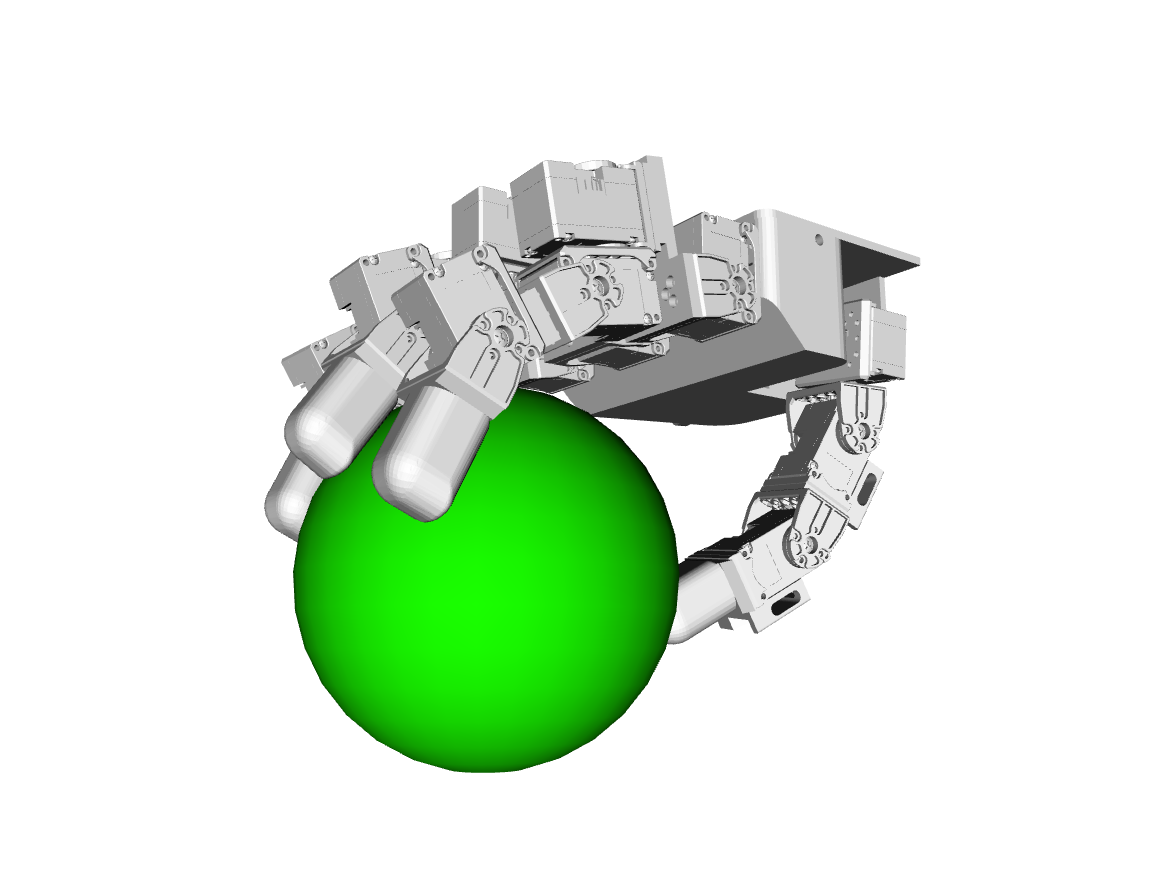}
			\includegraphics[width=0.325\linewidth]{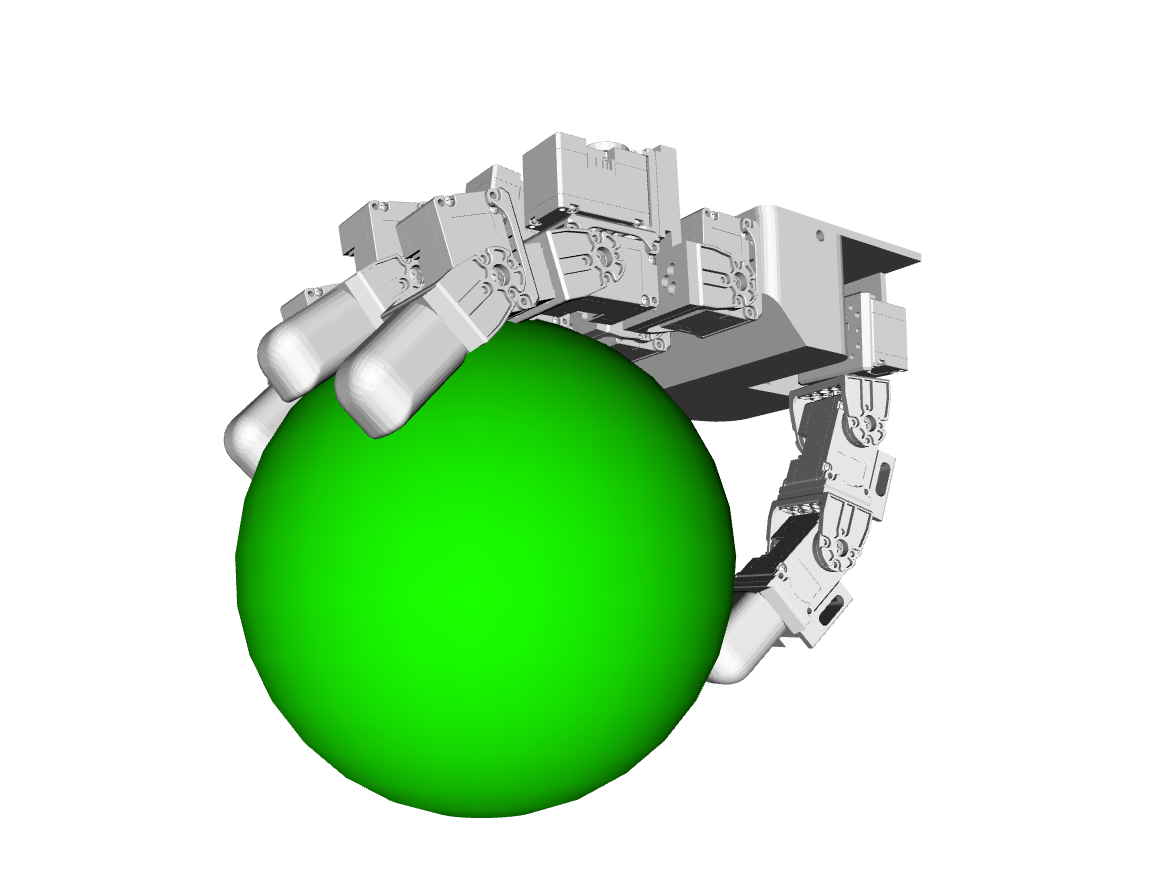}
			\includegraphics[width=0.325\linewidth]{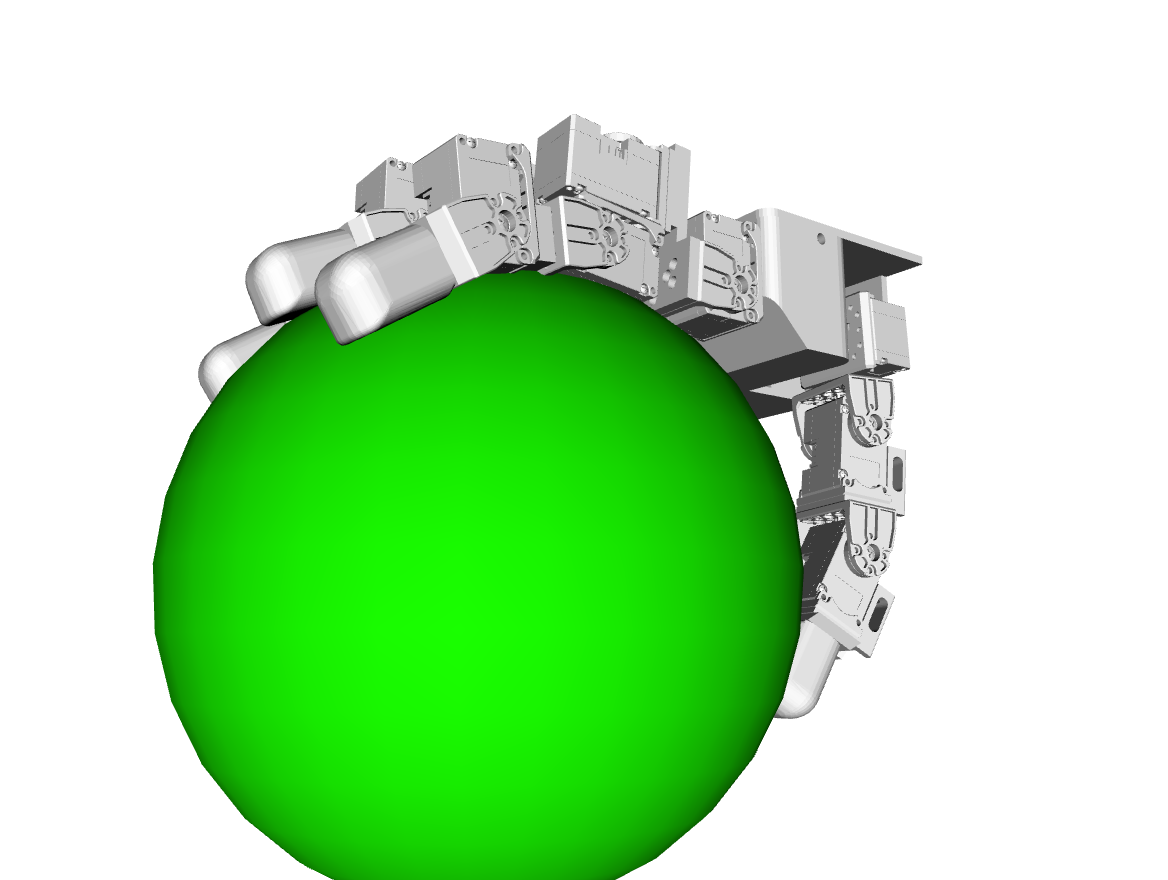}
			\caption{\color{black} Example of the effect of a pure dilation command. We use $\Delta t = 0.25$s, with two dilation commands $\bivector_{\similarity 1} = -\gae{0\infty}$ and $\bivector_{\similarity 2} = \gae{0\infty}$ that correspond to $d=0.3679$ and $d=2.718$. The original hand position is shown in the middle. The left image shows the updated hand position after we apply the dilation command $\bivector_{\similarity 1}$ and the right image  shows the updated hand position after we apply the dilation command $\bivector_{\similarity 2}$.}
			\label{fig:example_of_the_effect_of_a_dilation_command_}
		\end{figure}
		
		Here, we have shown the example for teleoperating the Leap hand, which has 16 degrees of freedom. The same principles also apply to other systems. We show in the accompanying video how we teleoperate a system of three manipulators having 21 degrees, and a system of two humanoids having 34 degrees of freedom. In all cases, we use the simple six-axes device as an input, which shows how the proposed modeling approach greatly simplifies the control of complex robotic systems.
		}




\section{Discussion}
\label{sec:discussion}

    Our unified control framework based on cooperative similarity transformations generalizes the rigid body transformation of a single robot's end-effector to the cooperative control of multiple kinematic chains. The experimental results demonstrated this approach across a variety of robotic systems, ranging from a bi-manual setup to scenarios involving collaborating humanoids. Given that the framework is purely geometric, and considering the variety of systems explored, the kinematic simulations show both the strengths and limitations of our method, which are further discussed in the following subsections.

    
    \subsection{Connection to Traditional Single-Arm Control}
    \label{sub:connection_to_traditional_single_arm_control}
        Throughout this article, we have mentioned several times that the cooperative similarity transformation and its Jacobians fulfill the same role as the rigid body transformation to the robot end-effector that is used in single-arm systems. Since similarity transformations form a seven-dimensional manifold, as opposed to the six-dimensional one of rigid body transformations, the modeling using the cooperative similarity transformations only introduces one additional dimension to achieve cooperative control behaviors of highly complex robotic systems. Although the involved mathematics that we derived in this article might seem unfamiliar, all the findings in the literature for traditional control methods of single-arm systems remain valid and are directly applicable to the scenarios shown in this article. We have shown this via differential kinematics control in the teleoperation experiments in Section~\ref{sub:teleoperation_examples}. This further highlights how the cooperative modeling greatly facilitates the control of complex robotic systems.  
    
    \subsection{Geometric Nullspace}
    \label{sub:geometric_nullspace_discussion}
\begin{figure}[!ht]
            \centering
            \begin{subfigure}[t]{\linewidth}
                \centering
                \includegraphics[width=0.8\linewidth]{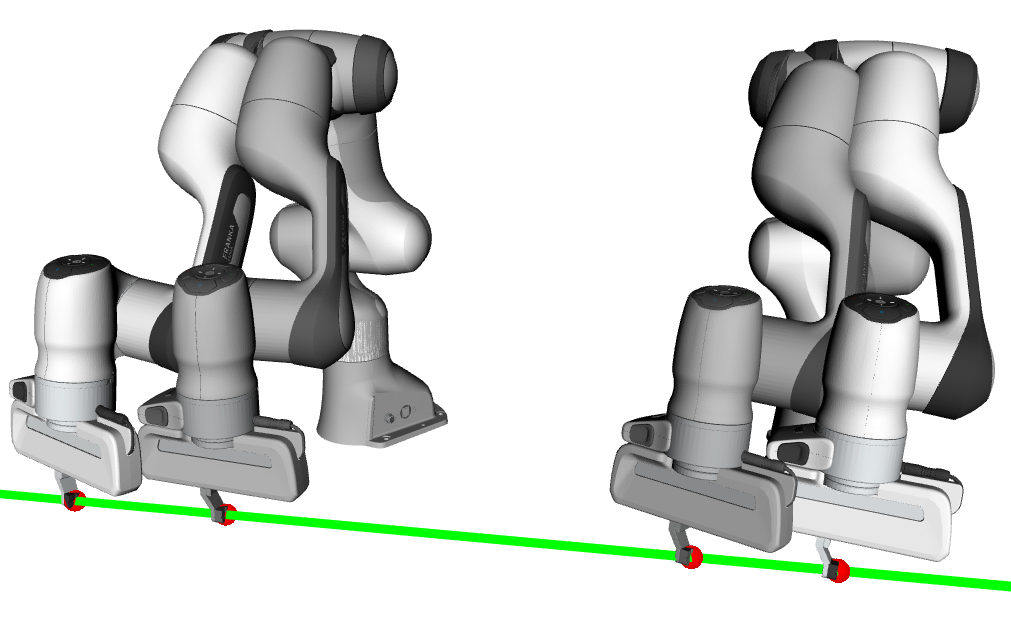}
                \caption{The line nullspace formed by two Franka robots. Each robot's end-effector is free to move along the line unrestricted. Secondary control objectives tangential to the line will not change the line.}
                \label{fig:two_pandas_line_nullspace_}
            \end{subfigure}
            \begin{subfigure}[t]{\linewidth}
                \centering
                \includegraphics[width=0.8\linewidth]{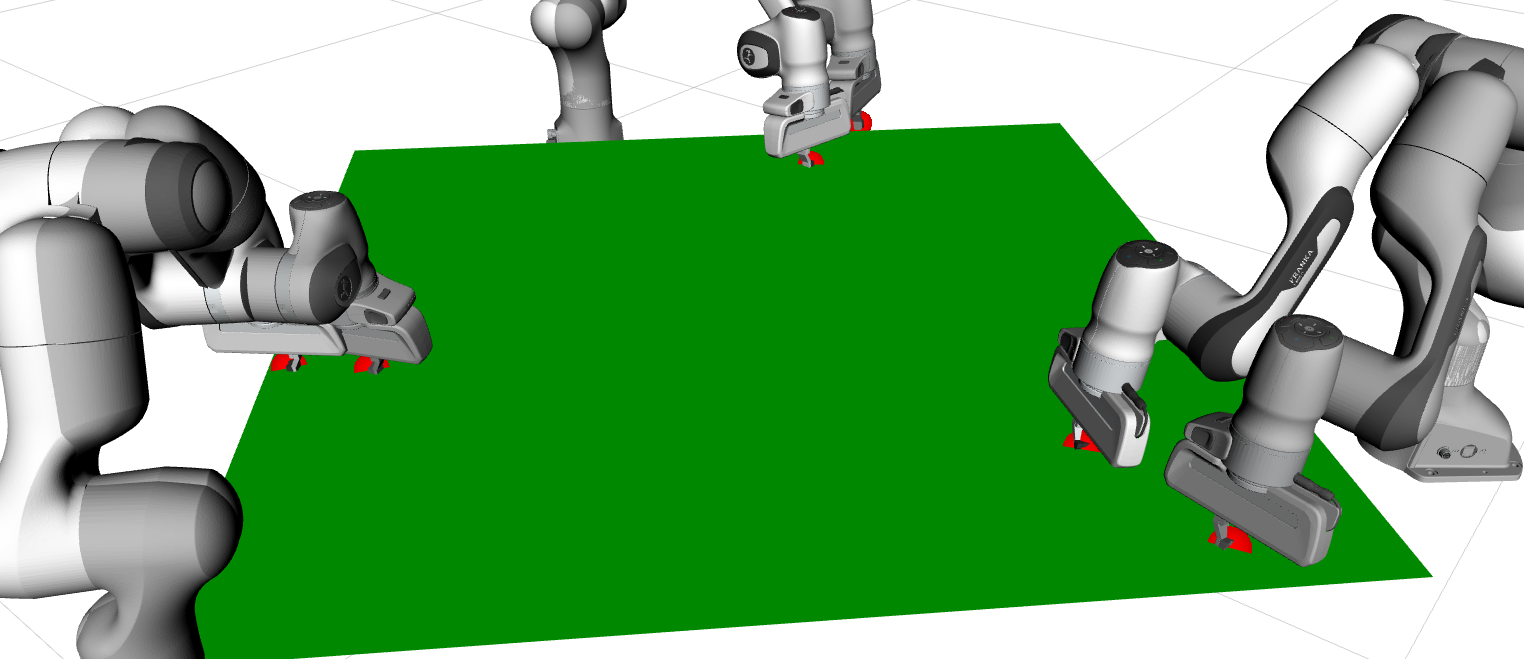}
                \caption{The plane nullspace formed by three Franka robots. Each robot's end-effector is free to move in the plane unrestricted. Secondary control objectives tangential to the plane will not change the plane.}
                \label{fig:three_pandas_plane_nullspace_}
            \end{subfigure}
            \caption{Geometric nullspaces that admit the introduction of secondary control objectives. Perturbations that are orthogonal to the geometric primitives will be rejected. The definition of these geometric nullspaces is coordinate-free.}
            \label{fig:geometric_nullspaces_}
        \end{figure}

        The systems considered in our experiments range from a bimanual setup with 14 degrees of freedom (DoF) (see Figure~\ref{fig:two_pandas_line_nullspace_}) to collaborating humanoids with 34 DoF in total. The resulting systems are highly redundant, and thus are inherently challenging to control. Even when restricted to the task space, the dimensionality remains high, i.e. ranging from 12 to 24 DoF. Notably, encoding the cooperative task space using geometric primitives significantly reduces the effective dimensionality of the problem. The similarity transformations associated with these primitives have at most 7 DoF, accounting for translation, rotation, and uniform scaling. Depending on the symmetry of the primitive, the effective number of DoF can be even lower. For instance, a sphere (see Figure~\ref{fig:leap_hand_reaching_for_a_sphere_}) admits only 4 DoF, as rotation does not alter its configuration. As can be inferred from the teleoperation experiment given in Figure~\ref{fig:leap_hand_teleoperation_}, this reduction in dimensionality is critical: no teleoperation device exists for 28-DoF systems, nor can a human or an existing algorithm feasibly control such a high-dimensional system directly. In contrast, using our proposed method for controlling the translation and dilation of a target sphere is straightforward.

        A second key aspect of the redundancy resolution arises from the use of geometric nullspaces. Geometric primitives naturally define subspaces of the underlying algebra, giving rise to geometric nullspace formulations for control. This structure ensures that the controller is inherently stiff when the system deviates from the primitive, and compliant when moving along it, as we previously showed in the context of optimal control for a single robotic arm~\cite{lowGeometricAlgebraOptimal2023}. This principle extends directly to the cooperative geometric primitives introduced in this work, as illustrated in Figures~\ref{fig:two_pandas_line_nullspace_} and \ref{fig:three_pandas_plane_nullspace_}.

        The geometric nullspaces induced by these primitives enable a decoupling of orthogonal control objectives in a coordinate-free manner, while eliminating the need for basis-specific representations or manual tuning of off-diagonal matrix terms. For example, aligning the end-effector with a target line does not constrain motion along the line, enabling the controlled application of contact forces in that direction. We previously demonstrated this feature with a single robot in an impedance control application on curved surfaces~\cite{bilalogluTactileErgodicCoverage2025}. As shown in Figure~\ref{fig:two_pandas_line_nullspace_}, this property naturally extends to cooperative settings, such as coordinated object lifting.

        Similarly, in the case of cooperative reaching toward a plane, optimization yields the closest configuration due to the presence of a geometric nullspace, as shown in Figure~\ref{fig:three_pandas_plane_nullspace_}. Although not illustrated here for brevity, this principle generalizes to other geometric primitives such as circles and spheres. For example, when four robotic arms cooperatively maintain a spherical constraint, they can move compliantly along the tangent directions of the sphere’s surface without violating the constraint, while the controller remains stiff in directions orthogonal to the sphere.

    \subsection{Singularities of Cooperative Primitives}
    \label{sub:singularities}
        Singularities in kinematic chains typically result in the loss of one or more degrees of freedom. These configurations are characterized by a degenerate Jacobian matrix that loses full rank. Similarly, singularities can arise in cooperative geometric primitives when the primitive is no longer uniquely defined. To illustrate this, consider the example of the circle primitive. If all three end-effector points lie on a straight line, the defining circle degenerates into a line. In the conformal geometric model, lines are interpreted as circles with infinite radius. As a result, the cooperative primitive becomes ill-defined, and its behavior cannot be uniquely determined.

        These degeneracies produce effects akin to conventional singularities: it becomes ambiguous which end-effector should move, and the associated geometric Jacobian grows unbounded as the three points approach co-linearity. While such configurations are rare and typically do not disrupt control in practice, we included manipulability matrices for the geometric primitives in Section~\ref{sub:manipulability_analysis} to provide tools for analyzing and avoiding these cases. Just as manipulability is used to avoid singularities in single-robot systems, these matrices can be employed alongside standard techniques for singularity-avoiding cooperative control.

    {\color{black}

        \subsubsection{Conditions for Well-Definedness}
        The cooperative geometric primitive $X_c(\posjoint)$ defined in Equation~\eqref{eq:cooperative_geometric_primitive} is well-defined if and only if the outer product of the defining points yields a non-degenerate geometric entity. This requirement imposes geometric constraints on the relative positions of the end-effector points. Specifically:

        \begin{itemize}
            \item \textbf{Pointpair/Line:} Two points must be distinct, i.e., $P_1 \neq P_2$. When the points coincide, the pointpair degenerates to a single point.
            \item \textbf{Circle/Plane:} Three points must be non-collinear. When the three points lie on a straight line, the circle degenerates to a line (or equivalently, a circle with infinite radius in CGA).
            \item \textbf{Sphere:} Four points must be non-coplanar. When all four points lie in the same plane, the sphere degenerates to a plane.
        \end{itemize}

        As the system approaches these degenerate configurations, the cooperative geometric primitive becomes increasingly ill-defined. In the limiting case, the primitive cannot uniquely determine the cooperative similarity transformation $\versor_{\similarity c}(\posjoint)$, and the control objective loses its geometric meaning.

        \subsubsection{Rank Deficiency and Jacobian Conditioning}
        The cooperative Jacobian matrices derived in Section~\ref{sub:cooperative_similarity_control} maintain sufficient rank as long as the cooperative primitive remains well-defined and the individual manipulator Jacobians are not singular. However, near degenerate configurations, the cooperative Jacobian exhibits poor conditioning and may lose rank.

        Consider the example of a cooperative circle approaching collinearity as shown in Section \ref{sub:geometric_singularity}. As the three defining points align, the radius of the circle grows unbounded, and certain components of the similarity transformation become ambiguous. Consequently, the geometric similarity Jacobian $\jacobian^G_{\similarity c}(\posjoint)$ grows unbounded in specific directions, similar to the behavior observed near kinematic singularities in single-arm systems.

        The manipulability analysis presented in Section~\ref{sub:manipulability_analysis} provides a quantitative measure for detecting such configurations. The manipulability ellipsoid $\bm{\mathcal{M}}_\similarity(\posjoint)$ becomes increasingly elongated as degeneracy is approached, with one or more eigenvalues tending toward zero. This behavior can be exploited to implement singularity avoidance strategies, analogous to those used in traditional manipulator control.

        \subsubsection{Practical Implications and Mitigation Strategies}
        In practice, the limitations discussed above can be mitigated through singularity avoidance strategies using the manipulability measure from Section~\ref{sub:manipulability_analysis} to detect and avoid degenerate configurations, similar to standard techniques in single-arm control. While these challenges are inherent to the cooperative primitive framework, they are not fundamentally different from the analogous challenges faced in traditional multi-robot coordination or single-arm control near singularities. The key distinction is that the cooperative geometric primitive provides an explicit, coordinate-free characterization of the degenerate configurations, making them easier to detect and avoid.


    }

    \subsection{Extension to Arbitrary Contact Points on the Robot Body}
        \label{sub:whole_body_control}
        In this article, we presented cooperative geometric primitives with respect to end-effector points. While this is a reasonable and fairly standard assumption, since the end-effector typically has the highest manipulability and can be equipped with a tool or sensor, there are many scenarios in which it is desirable to define cooperative geometric primitives with respect to arbitrary points on the robot's body.
        
        For example, when carrying a large and bulky object, humans often use additional contact points such as the torso to help distribute the load and improve stability. We illustrated a preliminary example of this idea in Figure~\ref{fig:humanoid_torso_reaching_circle}, where the torso of a humanoid robot was used to define the third point of a cooperative circle.
        
        In the general case, however, this requires to define a distance function from the robot surface to the geometric primitives and use its Jacobian to guide the optimization. Accordingly, the optimization can select these points dynamically, depending on the task requirements.

 
    \subsection{Extension to Human-Robot Collaboration}
    \label{sub:human_robot_collaboration}

        In Figures~\ref{fig:collaboration_between_manipulator_and_humanoid_} and~\ref{fig:two_humanoids_cooperative_sphere_reaching_}, we presented experiments involving humanoid robots. With a slight change in perspective, these humanoids can be interpreted as abstractions of humans in a human-robot collaboration setting. This observation suggests that cooperative geometric primitives could also be applied to model scenarios involving human-robot collaboration.

        The primary distinction between a human and a humanoid robot lies in controllability: the human's motion cannot be directly controlled. As a result, the system must adopt a leader-follower control paradigm, where the human acts as the leader and the robot as the follower. The robot then adapts its end-effector motion in response to the human's hand movements, in order to maintain a desired cooperative primitive, such as a circle or a sphere, defined by the relative positions of the human hand and the robot end-effector(s). Due to the benefits of using the cooperative geometric primitives, the resulting control scheme would remain geometrically consistent. 

    \subsection{Extension to $n$ Kinematic Chains}
    \label{sub:extension_to_n_kinematic_chains}
        While we presented control strategies for up to four parallel kinematic chains, this practical limitation only stems from our choice of algebra, i.e. conformal geometric algebra $\cga$. Here, CGA is the smallest algebra that allows for the representation of geometric primitives capturing the collaborative behaviour of up to four parallel kinematic chains. This choice is justified, since three and four-fingered hands are common and able to secure a grasp by removing all DoFs of a target object. Hence, we want to point out that this limitation is only of practical nature, but not of theoretical one. More concretely, using a different choice of the underlying quadratic space $\left( \mathbb{R}^{p,q,r}, g(\bm{x},\bm{x}) \right) $, the corresponding geometric algebra $\ga{p,q,r}$ would allow for the representation of more parallel kinematic chains, such as in the case of five-fingered hands. The presented results based on the derivation of the collaborative geometric primitives would remain valid.

    
    \subsection{Different Types of Geometric Primitives}
    \label{sub:different_types_of_geometric_primitives}
        In this work, we described the general cooperative similarity task spaces for multiple parallel kinematic chains. We wanted to emphasize the usage of the cooperative similarity transformation as an equivalent to the end-effector pose. Thus, in our control and optimization examples, we always assumed the target geometric primitives to be of the same type as the task space geometry, as listed in Table~\ref{tab:cooperative_similarity_task_spaces_}. In general, however, it is possible to formulate task objectives and constraints involving different types of geometric primitives. In principle, this could be approached via the outer product minimization that we presented for a single manipulator in \cite{lowGeometricAlgebraOptimal2023}, where cooperative geometric primitive could then be used to reach a desired point. Geometric relationships, such as intersections between primitives, can also be fully described in the algebra, and thus can be used in optimization problems. Note that these relationships are well-defined, i.e. they will always yield a mathematically valid result even if the primitives do not intersect.

    {\color{black}

    
    \subsection{Numerical Analysis}
    \label{sub:numerical_analysis}
        A critical numerical consideration in the implementation of our framework concerns the conditioning of the logarithmic map used to extract similarity bivectors from versors. The logarithmic maps for rotors, translators, dilators, and similarity transformations presented in Appendix~\ref{sec:transformation_groups_in_cga} all involve trigonometric functions in their denominators. The primary source of numerical instability arises when the rotation angle is a whole multiple of $\pi$, which manifests as the scalar part of the rotor $\langle V_R \rangle_0$ approaching $\pm 1$. In these configurations, the denominator $\sin(\cos^{-1}(\langle V_R \rangle_0))$ in Equation~\eqref{eq:log_rotor} becomes zero, leading to division by zero. These singular cases are straightforward to detect by monitoring the scalar component $|\langle V_R \rangle_0|$ of the rotor before computing the logarithm. Note that rotors are essentially identical to quaternions, and therefore the same challenges and solutions to the conditioning of the logarithmic map apply. Similar considerations apply to the other transformation groups, though the rotation component typically presents the most significant numerical challenge. 

    
    \subsection{Computational Efficiency}
    \label{sub:computational_efficiency}
        A similarity bivector $B_{\similarity} \in \mathbb{B}_{\similarity}$ requires storage of 7 scalar components, while the corresponding similarity transformation versor $V_{\similarity} \in \similarity$ requires 12 components (1 scalar, 7 bivector, and 4 quadvector components). The geometric product of two similarity transformations requires 112 floating-point multiplications and 100 additions, according to the geometric product rules where certain blade combinations are known to result in zero. For a single manipulator, a homogeneous transformation matrix in SE(3) maps the end effector pose to the target pose through a single matrix product. Exploiting the zeros in the last row of the homogeneous matrices, this multiplication requires about 39 multiplications and 34 additions. Consequently, for three or more independent kinematic chains, the total number of operations required by similarity transforms or by separate rigid body transformations similar in terms of computational complexity. Considering the fact that the similarity transformations encode up to four kinematic chains, this computational overhead is acceptable, since it in turns enables more efficient constraint evaluation on all kinematic chains simultaneously. An in-depth comparison of dual quaternions (i.e. the rigid transformation part of the similarity transformations) versus transformation matrices can be found in \cite{dantamRobustEfficientForward2021}.

    }


    \subsection{Practical Considerations}
    \label{sub:dynamics_and_real_world_systems}

        In this article, the results that we have shown based on our mathematical formulations are all in simulation. Our emphasis is on the theoretical derivations and implications of this novel modeling approach for cooperative systems of multiple parallel kinematic chains. However, we presented all the necessary tools for using our approach in practice. The three different control approaches shown in Section~\ref{sec:results} all yield valid joint space commands that can be used on real systems. Of course, the gain parameters, such as the stiffness and damping, would need to be tuned for a real system. Apart from that, a potential issue are non-zero joint velocities despite reaching the target. This is an inherent issue of redundant systems that can be solved using various methods that add dissipative terms in the nullspace while preserving passivity. 



\section{Conclusion}
\label{sec:conclusion}
    
    In this work, we introduced a framework for modeling the cooperative task space of multiple parallel kinematic chains through the integration of cooperative geometric primitives and similarity transformations. By deriving the mathematical foundations of this approach, we demonstrated that cooperative geometric primitives offer a powerful abstraction for simplifying the representation and analysis of complex robotic systems, particularly those with a high number of degrees of freedom. This abstraction reduces the inherent modeling complexity by encapsulating intricate kinematic interactions into unified geometric constructs, enabling intuitive and scalable coordination strategies. In this article, we focused on the mathematical derivation of the cooperative task spaces based on the geometric primitives and presented their application in abstract experiments in order to preserve generality. Hence, this work offers various interesting extension opportunities that could be addressed in future work. 

    Our formulation of the cooperative similarity transformation presents a direct link to classical control methods. Based on the cooperative similarity transformation, we derived an analytic and geometric Jacobian that can be used in standard control techniques. We demonstrated this using optimal control and teleoperation with differential kinematics as examples. Even though the controlled systems have a high number of degrees of freedom, the controller based on the similarity transformation is almost identical to single-arm controllers, and only adds a single dimension for controlling the dilation. Combined with the concept of geometric nullspaces, which decouple orthogonal control objectives in a coordinate-free manner, we have derived the theory of a versatile control framework that leverages the algebraic properties of geometric primitives. Our experiments have shown the general applicability of this control formulation. Actual real-world tasks, however, will require more complex modeling approaches, where different geometric primitives are combined in a single objective function in order to represent the task constraints. We will address this in future work by focusing on the applications, while using the findings of this article as the theoretical foundations. These applications could include the modeling of human-robot collaboration tasks, where the cooperative geometric primitives are used to implement geometrically consistent leader-follower paradigms that exploit the inherent nullspaces. 
    

\begin{acks}
\end{acks}

\begin{dci}
    The author(s) declared no potential conflicts of interest with respect to the research, authorship, and/or publication of this article.
\end{dci}

\begin{funding}
    This work was supported by the State Secretariat for Education, Research and Innovation in Switzerland for participation in the European Commission’s Horizon Europe Program through the INTELLIMAN project (https://intelliman-project.eu/, HORIZON-CL4-Digital-Emerging Grant 101070136) and the SESTOSENSO project (http://sestosenso.eu/, HORIZON-CL4-Digital-Emerging Grant 101070310).
\end{funding}

\printbibliography

\appendix

    
    \section{Transformation Groups in CGA}
    \label{sec:transformations_groups_in_cga}
        The geometric product of $n$ vectors $\bm{x}_i \in \cga$, i.e. $V = \prod_{i=1}^n \bm{x}_i$, forms the set of versors that combined with the geometric product form the Clifford group~$\Gamma(4,1)$. The restrictions to $VV^{-1}=\pm 1$ or $VV^{-1}= 1$, yield the groups $Pin(4,1)$  and $Spin(4,1)$, respectively. Note that these groups are double covers of the matrix Lie groups $\bm{O}(4,1)$ and $\bm{SO}(4,1)$, i.e. the orthogonal and special orthogonal groups. In general, this means that CGA allows for the representation of conformal transformations in $\mathbb{R}^3$ as a product of vectors instead of highly non-linear matrix expressions, since $\bm{O}(4,1)$ is isomorphic to the group of conformal transformations $Conf(3)$. 
        
        In the following, we list the different transformations groups of CGA. We focus on the subgroups of $Spin^+(4,1)$, i.e. we include only versors that are comprised of an even number of reflections in hyperplanes. This means that we omit pure reflections, since we are only interested in transformations that preserve the handedness.  

        
        \subsection{Rotation Group}
        \label{sub:rotor_group}
            The group of rotations in three-dimensional Euclidean space is usually represented by the special orthogonal group $\bm{SO}(3)$, i.e. a matrix Lie group. The group $Spin(3)$ is its double-cover and can be represented as unit quaternions. In CGA, it is the rotors that form an isomorphic group to unit quaternions and we denote them here as $\group{\rotor}$. Their Lie algebra is the bivector algebra $\algebra{\bivector}_{\rotor} = \text{span} \left\{ \gae{12}, \gae{13}, \gae{23} \right\}$. Given the elements $\versor_{\rotor}\in\group{\rotor}$ and $\bivector_{\rotor} \in \algebra{\bivector}_{\rotor}$, the exponential map $\exp_{\group{\rotor}}: \algebra{\bivector}_{\rotor} \to \group{\rotor}$ and its inverse the logarithmic map $\log_{\group{\rotor}}: \group{\rotor} \to \algebra{\bivector}_{\rotor}$ are

            \begin{equation}
                \begin{split}
                    \versor_{\rotor} 
                    = & \exp (\bivector_{\rotor}) 
                    \\
                    = & \cos\left( \frac{1}{2} \norm{\bivector_{\rotor}} \right) 
                    \\
                    & - \sin\left( \frac{1}{2} \norm{\bivector_{\rotor}} \right) \norm{\bivector_{\rotor}}^{-1} \bivector_{\rotor},
                \end{split}
            \end{equation}
            and
            \begin{equation}
                B_R = \log(\versor_{\rotor}) = \frac{-2\inverse{\cos} \big( \grade{\versor_{\rotor}}{0} \big) }{\sin \Big( \inverse{\cos} \big( \grade{\versor_{\rotor}}{0} \big)  \Big)}  \grade{\versor_{\rotor}}{2}.
            \end{equation}
        
        
        \subsection{Translation Group}
        \label{sub:translator_group}
            The translation group of $\mathbb{R}^3$ is the Euclidean space itself under the addition operation, i.e. $(\mathbb{R}^3,+)$, which is often shortened to simply $\mathbb{R}^3$. In CGA, this group can be represented in versor form. Here, we denote the translation group containing all translator versor in CGA as $\group{\translator}$. Note that, unlike the rotation group in CGA, the group $\group{\translator}$ is not a double-cover of $(\mathbb{R}^3,+)$, since $(\mathbb{R}^3,+)$ is already a simply-connected group. The Lie algebra of the group $\group{\translator}$ is the bivector algebra $\algebra{\bivector}_{\translator} = \text{span} \left\{ \gae{1\infty},\gae{2\infty},\gae{3\infty} \right\}$. Given the elements $\versor_{\translator}\in\group{\translator}$ and $\bivector_{\translator} \in \algebra{\bivector}_{\translator}$, the exponential map $\exp_{\group{\translator}}: \algebra{\bivector}_{\translator} \to \group{\translator}$ and its inverse the logarithmic map $\log_{\group{\translator}}: \group{\translator} \to \algebra{\bivector}_{\translator}$ are  

            \begin{equation}
                \versor_{\translator} = \exp \left( \bivector_{\translator} \right) = 1 - \frac{1}{2} \bivector_{\translator},
            \end{equation}
            and
            \begin{equation}
                \bivector_{\translator} = \log(\versor_{\translator}) = -2 \grade{\versor_{\translator}}{2}.
            \end{equation}
            In general, the translation bivector $\bivector_{\translator}$ can be found from a Euclidean vector $\bm{t} \in \mathbb{R}^3$ as 
            \begin{equation}
                \bivector_{\translator} = \bm{t} \outer \gae{\infty}.
            \end{equation}
            %
            

        
        \subsection{Uniform Scaling Group}
        \label{sub:uniform_scaling_group}
            Uniform scaling is a transformation that preserves geometric similarity, i.e. the shape, proportions, angles and orientation as well as parallelism and collinearity are preserved while distances are changed by an isotropic scaling factor. Here, we restrict uniform scaling to positive scalars $\mathbb{R}^+$ to preserve the handedness as well. In CGA, the versor achieving this is called a dilator $\versor_{\dilator}$ and consequently the set of all dilators is the dilation group $\group{\dilator}$, with its corresponding bivector Lie algebra $\algebra{\bivector}_{\dilator} = \text{span}\left\{ \gae{0\infty} \right\} $. Given the elements $\versor_{\dilator}\in\group{\dilator}$ and $\bivector_{\dilator} \in \algebra{\bivector}_{\dilator}$, the exponential map $\exp_{\group{\dilator}}: \algebra{\bivector}_{\dilator} \to \group{\dilator}$ and its inverse the logarithmic map $\log_{\group{\dilator}}: \group{\dilator} \to \algebra{\bivector}_{\dilator}$ are  

            \begin{equation}
                \begin{split}
                    \versor_{\dilator} 
                    = & \exp \left( \bivector_{\dilator} \right) 
                    \\
                    = & \cosh \left( \frac{1}{2}\norm{\bivector_{\dilator}} \right) - \sinh \left( \frac{1}{2}\norm{\bivector_{\dilator}} \right) \gae{0\infty}, 
                \end{split}
            \end{equation}
            and
            \begin{equation}
                \bivector_{\dilator} = \log(\versor_{\dilator}) = 2 \inverse{\cosh}\big(\grade{\versor_{\dilator}}{0}\big) \gae{0\infty},
            \end{equation}
            where the bivector $\bivector_{\dilator}$ then relates to the scaling factor $d \in \mathbb{R}^+$ via 
            \begin{equation}
                \bivector_{\dilator} = \log (d) \gae{0\infty}.
            \end{equation}
            Note that, the scaling is always with respect to the origin.
        
        
        \subsection{Rigid Transformation Group}
        \label{sub:rigid_transformation_group}
            The group of rigid transformations in Euclidean space is the most commonly used group in robotics. Traditionally, it is represented by the matrix Lie group $\bm{SE}(3)$ called the special Euclidean group. Alternative representations, such as dual quaternions, are representations of $Spin(3) \ltimes \mathbb{R}^3$, which is the double-cover of $\bm{SE}(3)$. Here, we denote this group as $\group{\rigid}$ and usually call its elements motors $\versor_{\rigid}$. The group is found as $\group{\rigid} = \group{\rotor} \ltimes \group{\translator}$, and we define the canonical decomposition of an element $\versor_{\rigid} \in \group{\rigid}$ as 
            \begin{equation}
                \versor_{\rigid} = \versor_{\translator}\versor_{\rotor}.    
            \end{equation}
            The Lie algebra of $\group{\rigid}$ is the bivector algebra $\algebra{\bivector}_{\rigid} = \text{span}\left\{ \gae{12}, \gae{13}, \gae{23}, \gae{1\infty}, \gae{2\infty}, \gae{3\infty} \right\} $ and an element $\bivector_{\rigid}\in\algebra{\bivector}_{\rigid}$ is decomposed as 
            \begin{equation}
                \bivector_{\rigid} = \bivector_{\translator} + \bivector_{\rotor}.
            \end{equation}
            
            Consequently, given the elements $\versor_{\rigid}\in\group{\rigid}$ and $\bivector_{\rigid} \in \algebra{\bivector}_{\rigid}$, the exponential map $\exp_{\group{\rigid}}: \algebra{\bivector}_{\rigid} \to \group{\rigid}$ and its inverse the logarithmic map $\log_{\group{\rigid}}: \group{\rigid} \to \algebra{\bivector}_{\rigid}$ are
            \begin{equation}
                \versor_{\rigid} 
                = \exp \left( \bivector_{\rigid} \right) 
                = \exp \left( \bivector_{\translator} \right) \exp \left( \bivector_{\rotor} \right),
            \end{equation}
            and
            \begin{equation}
                \bivector_{\rigid} = \log (\versor_{\rigid}) = \log \left( \versor_{\translator} \right) + \log \left( \versor_{\rotor} \right). 
            \end{equation}

        
        \subsection{Similarity Group}
        \label{sub:similarity_group}
            The Lie group of direct similarity transformations combines translations, rotations and uniform scaling. In terms of the matrix Lie groups it can be found as the semi-direct product of the special Euclidean group and positive scalars, i.e. $\bm{SIM}(3) = \bm{SE}(3) \rtimes \mathbb{R}^+$. Its double-cover representation using the spin groups found in geometric algebras therefore is $\big(Spin(3) \ltimes \mathbb{R}^3\big) \rtimes \mathbb{R}^+$. Here, we denote the versor representation of the similarity group in CGA as $\group{\similarity} = \group{\rigid} \rtimes \group{\dilator} = (\group{\rotor} \ltimes \group{\translator}) \rtimes \group{\dilator}$ and we define the canonical decomposition of an element $\versor_{\similarity} \in \group{\similarity}$ as 
            \begin{equation}
                \versor_{\similarity} = \versor_{\translator}\versor_{\rotor}\versor_{\dilator}.
            \end{equation}
            The Lie algebra of $\group{\similarity}$ is the bivector algebra $\algebra{\bivector}_{\similarity}  = \text{span}\left\{ \gae{12}, \gae{13}, \gae{23}, \gae{0\infty}, \gae{1\infty}, \gae{2\infty}, \gae{3\infty} \right\}$ and an element $\bivector_{\similarity}\in\algebra{\bivector}_{\similarity}$ is decomposed as 
            \begin{equation}
                \bivector_{\similarity} = \bivector_{\translator} + \bivector_{\rotor} + \bivector_{\dilator}.
            \end{equation}

            Consequently, given the elements $\versor_{\similarity}\in\group{\similarity}$ and $\bivector_{\similarity} \in \algebra{\bivector}_{\similarity}$, the exponential map $\exp_{\group{\similarity}}: \algebra{\bivector}_{\similarity} \to \group{\similarity}$ and its inverse the logarithmic map $\log_{\group{\similarity}}: \group{\similarity} \to \algebra{\bivector}_{\similarity}$ are
            \begin{equation}
                \versor_{\similarity} 
                = \exp \left( \bivector_{\similarity} \right) 
                = \exp \left( \bivector_{\translator} \right) \exp \left( \bivector_{\rotor} \right) \exp \left( \bivector_{\dilator} \right),
            \end{equation}
            and
            \begin{equation}
                \bivector_{\similarity} = \log (\versor_{\similarity}) = \log \left( \versor_{\translator} \right) + \log \left( \versor_{\rotor} \right) + \log \left( \versor_{\dilator} \right). 
            \end{equation}
            


    
    \section{Jacobians}
    \label{sec:jacobians}

        
        \subsection{Normalizing a Multivector}
        \label{sub:jacobian_of_normalized_multivector}
            Given a multivector $X$, where $X\reverse{X} \in \mathbb{R}$, it can be normalized as
            \begin{equation}
                \bar{X} = X \abs{X\reverse{X}}^{-\frac{1}{2}}.
            \end{equation}

            For a multivector valued function $X = F(\bm{x})$, we can then find its derivative as 
            %
            \begin{equation}
                \begin{split}
                \frac{\partial }{\partial \bm{x}} \bar{F}(\bm{x}) 
                = & \abs{X\reverse{X}}^{-\frac{1}{2}} \jacobian_F 
                \\
                & - \frac{1}{2} \abs{X\reverse{X}}^{-\frac{3}{2}} \left( X (\jacobian_F\reverse{X} + X\reverse{\jacobian}_F) \right) .
                \end{split}
            \end{equation}


        
        \subsection{Inverting a Multivector}
        \label{sub:jacobian_for_inverting_a_multivector}
            Given a multivector $X$, where $X\reverse{X} \in \mathbb{R}$, its inverse can be found as 
            \begin{equation}
                \inverse{X} = \reverse{X} \left(X\reverse{X}\right)^{-1}.
            \end{equation}

            For a multivector valued function $X = F(\bm{x})$, we can then find its derivative as 

            \begin{equation}
                \begin{split}
                \frac{\partial }{\partial \bm{x}} \inverse{F}(\bm{x}) 
                = & \left(X\reverse{X}\right)^{-1} \reverse{\jacobian}_F 
                \\
                & - 2 \left(X\reverse{X}\right)^{-2} \reverse{X} \grade{X \reverse{\jacobian}_F}{0}.
                \end{split}
            \end{equation}
            %
            
        
        
        \subsection{Analytic Circle Similarity Jacobian}
        \label{sub:analytic_similarity_jacobian}
            Here, we derive the analytic similarity Jacobian $\jacobian^A_{S,c}(\posjoint)$, given the cooperative geometric primitive $X_c(\posjoint)$ and the corresponding Jacobian $\jacobian^A_c(\posjoint)$ from Equation~\eqref{eq:cooperative_geometric_primitive_jacobian}. As shown in Equation~\eqref{eq:analytic_similarity_jacobian}, we can decompose the analytic similarity into the Jacobians for the translator $\jacobian^A_{\translator}(\posjoint)$, rotor $\jacobian^A_{\rotor}(\posjoint)$, and dilator $\jacobian^A_{\dilator}(\posjoint)$. 
            
            The analytic dilator Jacobian can be found as
            \begin{equation}
                \jacobian^A_{\dilator}(\posjoint) = -\frac{1}{4} (P_{\infty}\reverse{P}_{\infty})^{-1}  \grade{\jacobian_{\infty} \reverse{P}_{\infty} + P_{\infty} \reverse{\jacobian}_{\infty}}{0} \versor_{\dilator},
            \end{equation}
            where $P_{\infty}$ is the projection of infinity from Equation~\eqref{eq:circle_radius}
            \begin{equation}
                P_{\infty} = \left( X \inner \gae{\infty}\right) \inverse{X},
            \end{equation}
            and $\jacobian_{\infty}$ its Jacobian
            \begin{equation}
                \jacobian_\infty = (\gae{\infty} \inner \jacobian^A_c) \inverse{\circle} + (\gae{\infty} \inner \circle) \hat{\jacobian^A_c}.
            \end{equation}

            The analytic rotor Jacobian can be found as
            \begin{equation}
                \begin{split}
                \jacobian^A_{\rotor} (\posjoint)
                = & \abs{R\reverse{R}}^{-\frac{1}{2}} \jacobian_\rotor
                \\
                & - \frac{1}{2} \abs{R\reverse{R}}^{-\frac{3}{2}} \rotor (\rotor \reverse{\jacobian}_\rotor + \jacobian_\rotor\reverse{\rotor}),
                \end{split}
            \end{equation}
            where $R$ is the rotor $\versor_\rotor(\posjoint)$ before normalization according to Equation~\eqref{eq:circle_plane_rotor} and $\jacobian_\rotor$ the corresponding Jacobian
            \begin{equation}
                \jacobian_\rotor = \gae{3} \inner \bar{\jacobian}_N - \gae{3} \outer \bar{\jacobian}_N, 
            \end{equation}
            where $\bar{\jacobian}_N$ is the normalized Jacobian of the plane normal found in Equation~\eqref{eq:plane_normal}
            \begin{equation}
                \begin{split}
                    \bar{\jacobian}_{N}
                    = & \abs{N\reverse{N}}^{-\frac{1}{2}} \jacobian_N
                    \\
                    & - \frac{1}{2} \abs{N\reverse{N}}^{-\frac{3}{2}} N (N \reverse{\jacobian}_N + \jacobian_N\reverse{N}),
                \end{split}
            \end{equation}
            where
            \begin{equation}
                \jacobian_{N} = \dual{\jacobian}_\plane - \frac{1}{2} (\gae{0} \inner \dual{\jacobian}_\plane) \gae{\infty},
            \end{equation}
            and
            \begin{equation}
                \jacobian_\plane = \jacobian^A_c (\posjoint) \outer \gae{\infty}.
            \end{equation}

            The analytic translator Jacobian can be found as
            \begin{equation}
                \begin{split}
                    \jacobian^A_{\translator} =  
                    -& \frac{1}{2} \grade{P_c}{0}^{-1} \jacobian_P 
                    \\
                    -& \grade{P_c}{0}^{-1} P_c (-\gae{\infty} \inner \grade{\jacobian_P}{0}) \outer \gae{\infty},
                \end{split}
            \end{equation}
            where $P_c$ is the center point and $\jacobian_P$ its Jacobian
            \begin{equation}
                \jacobian_P = \jacobian^A_c(\posjoint) \gae{\infty} X_c(\posjoint) + X_c(\posjoint) \gae{\infty} \jacobian^A_c(\posjoint).
            \end{equation}

\end{document}